\documentclass[letterpaper,conference,compsoc,10pt]{IEEEtran}
\IEEEoverridecommandlockouts

\usepackage{blindtext}
\usepackage{fancyhdr}

\usepackage[colorlinks,
            linkcolor=red,
            anchorcolor=red,
            citecolor=green]{hyperref} 

\usepackage{graphicx}
\usepackage{amsmath}
\usepackage{amssymb}
\usepackage{booktabs}

\usepackage{soul}
\usepackage{times}
\usepackage{epsfig}
\usepackage{bbm}
\usepackage{mathrsfs}
\usepackage{multirow}
\usepackage{paralist}
\usepackage{mathtools}
\usepackage{url}            
\usepackage{booktabs}       
\usepackage{nicefrac}       
\usepackage{threeparttable}
\usepackage{float}
\usepackage{listings}
\usepackage{wrapfig}

\usepackage{pifont}
\usepackage[utf8]{inputenc}

\usepackage{comment}
\usepackage{amsmath,amssymb,amsfonts}

\usepackage{tikz}
\usepackage{amsmath}
\usepackage{amssymb}
\usepackage{graphicx}
\usepackage{multirow}
\usepackage{makecell}
\usepackage{wrapfig}
\usepackage{amsfonts}
\usepackage{amsthm}
\usepackage{bm}
\usepackage{bbm}

\usepackage{caption}
\usepackage{subcaption}
\usepackage{algorithmic}
\usepackage{graphicx}
\usepackage{textcomp}
\usepackage{tabularx} 
\usepackage{collcell}
\usepackage{array}
\usepackage{caption}
\usepackage{hhline}
\usepackage[linesnumbered, ruled, vlined]{algorithm2e}
\usepackage{diagbox}
\let\emptyset\varnothing

\DeclareMathAlphabet\mathbfcal{OMS}{cmsy}{b}{n}

\newcolumntype{P}[1]{>{\centering\arraybackslash}p{#1}}

\usepackage{array}

\usepackage{tikz}
\usepackage{amsmath}
\usepackage{enumitem}
\usepackage[section]{placeins}
\usepackage{ulem}
\usepackage{booktabs}

\usepackage[export]{adjustbox}
\usepackage[square, comma, sort&compress, numbers]{natbib}
\usepackage{color, colortbl}
\definecolor{LightGray}{gray}{0.9}
\definecolor{White}{gray}{1.0}
\definecolor{Celadon}{RGB}{175, 225, 175}
\definecolor{lightyellow}{RGB}{255, 255, 200}
\definecolor{D3D3D3}{HTML}{D3D3D3}
\definecolor{FFCCCC}{HTML}{FFCCCC}
\definecolor{D9EAD3}{HTML}{D9EAD3}

\def\ie{\textit{i.e.}}
\def\eg{\textit{e.g.}}

\usepackage{enumitem}
\usepackage[capitalize]{cleveref}
\Crefname{section}{Sec.}{Secs.}
\Crefname{table}{Table}{Tables}
\Crefname{equation}{Eq.}{Eqs.}
\Crefname{figure}{Fig.}{Figs.}
\Crefname{lemma}{Lemma}{Lemmas}
\Crefname{theorem}{Theorem}{Theorems}
\Crefname{definition}{Definition}{Definitions}
\Crefname{hypothesis}{Hypothesis}{Hypothesises}
\Crefname{Line}{line}{lines} 
\Crefname{line}{line}{lines} 

\usepackage[most]{tcolorbox}
\newtcolorbox[auto counter]{mtbox}[1]{
    left=0.25mm,
    right=0.25mm,
    top=0.25mm,
    bottom=0.25mm,
    sharp corners,
    colframe=blue!50!black,
    boxrule=0.5pt,
    title={#1~\thetcbcounter},
    fonttitle=\bfseries,
    coltitle=blue!50!black,
    attach title to upper={\ --\ }
}

\usepackage{siunitx}
\usepackage{listings}
\usepackage{dsfont}
\usepackage{bm}

\newtheorem{remark-star}{Remark}
\newtheorem{remark-star-1}{Remark}

\newtheorem*{proof-sketch}{Proof Sketch}
\usepackage{booktabs}

\DeclareMathOperator*{\argmax}{\arg\!\max}

\newcommand{\R}{\mathbb{R}}

\newcommand{\mE}{\mathcal{E}}

\newcommand{\mX}{\mathcal{X}}

\newcommand{\red}[1]{\textcolor{red}{#1}}

\newcommand{\clean}{\mathcal{P}}

\newcommand{\trigger}{\mathcal{T}}

\newcommand{\downdata}{\mathcal{D}}
\newcommand{\cleandata}{\mathcal{D}_{\mathrm{clean}}}
\newcommand{\predata}{\mathcal{D}_{\mathrm{pre}}}
\newcommand{\poisondata}{\mathcal{D}_{\mathrm{poison}}}

\newcommand{\seeddata}{\mathcal{S}}
\newcommand{\subdata}{\mathcal{D}_{\mathrm{sub}} }
\newcommand{\prepara}{\theta_{\mathrm{pre}}}
\newcommand{\downpara}{\phi_\mathrm{down}}

\newcommand{\smalltcc}[1]{\footnotesize\textnormal{\tcc{#1}}}

\newcommand{\loss}{\ell}

\newcommand{\threatone}{\textbf{\underline{Threat-1}}}
\newcommand{\threattwo}{\textbf{\underline{Threat-2}}}
\newcommand{\threatthree}{\textbf{\underline{Threat-3}}}

\newcommand{\review}[2]{{\color{red} \underline {[Review \##1, #2]}}}
\SetKwInput{KwParam}{Parameters}

\begin{document}

\title{
Secure Transfer Learning: Training Clean Model Against Backdoor in \\ Pre-trained
Encoder and Downstream Dataset}

\author{
  \IEEEauthorblockN{
    Yechao Zhang\textsuperscript{1},
    Yuxuan Zhou\textsuperscript{1},
    Tianyu Li\textsuperscript{1},
    Minghui Li\textsuperscript{1},
    Shengshan Hu\textsuperscript{1},
    Wei Luo\textsuperscript{2},
    Leo Yu Zhang\textsuperscript{3}
  }
  \IEEEauthorblockA{
    \textsuperscript{1}Huazhong University of Science and Technology 
        \textsuperscript{2}Deakin University
            \textsuperscript{3}Griffith University
  }
\footnotesize{\tt \{ycz, yuxuanchou, tianyuli, minghuili, hushengshan\}@hust.edu.cn}\\
\footnotesize{\tt wei.luo@deakin.edu.au} \  \ \
\footnotesize{\tt leo.zhang@griffith.edu.au}
}

\maketitle


\begin{abstract}
Transfer learning from pre-trained encoders has become essential in modern machine learning, enabling efficient model adaptation across diverse tasks.
However, this combination of pre-training and downstream adaptation creates an expanded attack surface, exposing models to sophisticated backdoor embedding at both the encoder and dataset levels—an area often overlooked in prior research. Additionally, the limited computational resources typically available to users of pre-trained encoders constrain the effectiveness of generic backdoor defenses compared to end-to-end training from scratch.
In this work, we investigate how to mitigate potential backdoor risks in resource-constrained transfer learning scenarios. Specifically, we first conduct an exhaustive analysis of existing defense strategies, revealing that many follow a reactive workflow based on assumptions that do not scale to unknown threats, novel attack types, or different training paradigms. In response, we introduce a proactive mindset focused on identifying clean elements and propose the Trusted Core (T-Core) Bootstrapping framework, which emphasizes the importance of pinpointing trustworthy data and neurons to enhance model security.
Our empirical evaluations demonstrate the effectiveness and superiority of T-Core, specifically assessing 5 encoder poisoning attacks, 7 dataset poisoning attacks, and 14 baseline defenses across 5 benchmark datasets, addressing 4 scenarios of 3 potential backdoor threats.
\end{abstract}

\section{Introduction}

\vspace{-1mm}
Transfer learning (TL) has become an essential tool in machine learning applications, allowing developers to create sophisticated models by modifying existing ones to suit their specific tasks. 
This training paradigm is especially advantageous for users with limited computational and training resources, as it only requires collecting a small amount of training data and minor adaptation of a pre-trained encoder from third-party or open-source repositories.
However, relying on pre-trained encoders or collecting data from external sources opens up significant security risks, exposing transfer learning models to malicious actors with access to the pre-trained model or the data to launch backdoor attacks.
In backdoor attacks, an attacker may manipulate a few training samples of the victim by embedding a backdoor trigger and (mis)labeling them as a target class \cite{gu2019badnets,chen2017targeted} or directly train a backdoor model  \cite{liang2024badclip,BadEncoder,shuowang} and deliver it to the victim.
Either way, during inference time, they allow an attacker to stealthily control the victim model's behavior to produce target outcomes given specific conditions.

\textbf{A Challenging Defense Context.} Transfer learning poses potential backdoor risks from both sources, creating a complex situation where the pre-trained encoder, the dataset, or both may be compromised, leading to three types of backdoor threats. However, prior research has primarily focused on only one vector of poisoning.
Moreover, the transfer learning setup poses underappreciated challenges for general edge users with computational constraints when addressing backdoor threats.
For these users, their limited computational resources typically can only support them in fine-tuning only part of the model parameters in transfer learning, \eg, the last few layers.
Nevertheless, almost all existing defenses are designed for scenarios where the entire model can be trained, usually requiring the support to train a model from scratch in an end-to-end learning manner.

\textbf{A Exhaustive Defense Analysis.}  To further understand the combat against backdoor threats in this general yet challenging resource-limited transfer learning scenario, we exhaustively discuss and analyze existing defense strategies that presumably can help to mitigate the backdoor threats. 
However, we found that defenses effective in eliminating backdoors during end-to-end training from scratch fail to produce a clean model with high accuracy when adapted to the new defense context.
Many existing defenses primarily rely on a reactive workflow to identify and eliminate poison elements associated with specific known threats, heavily depending on assumptions about these elements. Unfortunately, these assumptions do not scale to unknown threats, novel attack types, or different training paradigms, as summarized in \cref{tab:defense_types}.

\textbf{A Proactive Mindset.}
To overcome the limitations of reactive workflows, we advocate for a proactive mindset. Instead of trying to identify all poisoned elements in a mixture of clean and poisoned samples from the post-attack model, we emphasize the importance of identifying high-credibility clean elements and proactively training with these trusted samples to address \textit{unknown} backdoor threats.
The rationale is that pinpointing a limited number of clean elements is significantly easier and more accurate than identifying all poisoned elements, particularly in complex scenarios where poison can originate from different sources and backdoor triggers can take various forms. Defenders can then gradually expand this foundation by evaluating additional trustworthy elements while keeping untrusted elements separate.


\textbf{A Bootstrapping Defense Framework.}
We propose a Trusted Core (T-Core) Bootstrapping framework that utilizes a proactive approach to learning a clean model through gradually bootstrapping verified data and model neurons.
Our process begins by identifying a limited number of high-credibility samples from each class. This selection is based on assessing topological invariance across different layers of a well-trained model using the entire dataset. 
Next, we create a clean subset by treating these samples as seed data. We apply a method of unlearning the seed data while simultaneously learning from the remaining data and selecting the sample with maximal prediction loss to expand the dataset. 
Subsequently, we introduce a selectively imbalanced unlearn-recover process to filter clean channels within the pre-trained encoder, utilizing the clean subset we've established. 
Finally, we leverage both the trusted dataset and the trusted model neurons to bootstrap further the learning of a clean model, which involves gradually expanding the pool of clean samples and refining the model.

\textbf{A Comprehensive Empirical Evaluation.} We extensively evaluate a diverse set of 5 encoder poisoning attacks \cite{BadEncoder,CTRL,DRUPE,SSL-backdoor,zhang2022corruptencoder}, 7 dataset poisoning attacks \cite{gu2019badnets,barni2019new,chen2017targeted,tang2021demon,qi2023revisiting,qi2023revisiting,nguyen2021wanet}, and 5 benchmark datasets of image classification (CIFAR-10, STL-10, GTSRB, SVHN, and ImageNet), covering 4 possible scenarios of all 3 types of backdoor threats in transfer learning.
Throughout the paper, we evaluate 14 baseline defenses \cite{CBD,ABL,hayase21a,CLP,chen2018detecting,tran2018spectral,pan2023asset,qi_sec_2023,gao2019strip,hou2024ibdpsc,guo2023scaleup,ibau,sam_backdoor,han2024effectiveness} within the defense context of transfer learning, and none of them match the effectiveness of our Trusted Core Bootstrapping framework in any module or in its entirety.
This illustrates the potential of the proactive mindset we advocate in defending against unknown backdoor threats.

Finally, we summarize our contributions as follows:
\begin{itemize}
    \item We identify a complex and challenging yet general backdoor threat model within the transfer learning scenario that previous research has overlooked.
    \item We conduct an exhaustive analysis of the existing backdoor defense in the defense context and reveal their limitations under the transfer learning scenario.
    \item We propose a proactive mindset as an alternative and introduce a Trusted Core Bootstrapping defense framework as an instantiation, providing concrete designs that are more robust and generalizable.
    \item We conduct extensive experiments to demonstrate the effectiveness of our Trusted Core Bootstrapping framework, as well as its superiority over existing designs in both module-by-module and end-to-end evaluation.
\end{itemize}



\section{Preliminaries}
In this section, we deﬁne our setup (\cref{subsec:notations}), deliver the threat model of backdoor attacks under transfer learning, and the defense context for a general edge user (\cref{sec:threatmodel}).

\subsection{Training Procedure, Models and Data}\label{subsec:notations}
We consider a popular transfer learning (TL) paradigm in image classification, \ie,``\textit{self-supervised pretraining} followed by \textit{supervised fine-tuning}".
The self-supervised stage involves using a large set of unlabeled data, referred to as the \textit{pre-training dataset} $\predata$, to develop an image encoder serving as a representation function $g: \mX \mapsto \mE$, where $\mX = \R^d$ represents the input space, and $\mE$ is the embedding space.  
In the fine-tuning stage, a task-specific classification head $f(\cdot; \phi)$ is concatenated on top of $g(\cdot; \theta)$, creating a combined network $h(\cdot; \theta, \phi)=f(\phi)\circ g(\cdot; \theta): \mX \mapsto \Delta^{C}$, which is tailored for the \textit{downstream task} of $C$-classes image classification and $\Delta^{C}$ representing the probability output space over the $C$ classes.
Overall, $h$ comprises multiple layers, $\{h^{(l)}: l \in [1, N]\}$. Given an input $x$, the output of the neural network $h$ is computed as $h(x) = f (g(x)) = \left(h^{(N)} \circ \cdots \circ h^{(1)} \right)(x)$, where {$f$} only represents the last few layers. 
Taking advantage of the feature extraction capability of the encoder, the fine-tuning process primarily updates $\phi$ on the $C$-class labeled \textit{downstream training dataset} $\downdata$ with training loss $\sum_{(x_i, y_i) \in \mathcal{D}} \ell(h(x_i), y_i)$ between the probability of output $h(x)$ and $y$, while generally freezing or making minor adjustments to $\theta$. 
During inference, for any input feature $x \in \mathcal{X}$,  the resulting classifier $F(\cdot) := \argmax_c h_c(\cdot)$ takes $F(x)$ as the predicted label. We use $\clean$ to denote the clean data distribution of the downstream classification task and $\mathbb{E}_{(x,y) \sim \clean} \mathbb{I}(F(x) = y)$ as \emph{clean accuracy~(ACC)}.


\subsection{Threat Model and Defense Context} \label{sec:threatmodel}
Our work considers the scenario in which an edge user aims to securely develop an image classifier based on an untrusted pre-trained encoder $g$  and an untrusted training dataset $\predata$.
This is relevant for a typical user who cannot guarantee the authenticity of the data it collects and does not have the resources (\eg, computational power, and memory) to train a model from scratch, thus requiring a pre-trained model to facilitate its training. 
Given the growing trend of users accessing open-source datasets and pre-trained models from AI platforms such as HuggingFace, where anyone can freely upload resources, we believe this considered scenario is increasingly practical.
In particular, our work focuses on the backdoor threats within this scenario.





\noindent \textbf{Attack Goals.} In general, regardless of how and where the trigger is injected, a desired backdoor attack should satisfy:
\begin{gather}
\mathbb{E}_{(x,y)\sim \clean} \mathbb{I}(F(x)  = y) \ge \tau_{ACC} , \label{eq:acc}      \\
\mathbb{E}_{(x,y)\sim \clean|_{y \ne t}}\mathbb{I}(F(\trigger(x)) = t ) \ge  \tau_{ASR}, \label{eq:asr}
\end{gather}
where $\trigger: \mX \mapsto \mX$ represents the adversary's trigger function that transforms a benign input into a trigger-injected one, and $t$ denotes the \textit{target class} chosen by the adversary.
\cref{eq:acc} specifies that the attack shall not affect the standard functionality, that is, 
 a high clean accuracy (above some $\tau_{ACC}$) for the classification model $F$.
 \cref{eq:asr}, on the other hand, requires the backdoored classifier to classify any trigger-injected input $\trigger(x)$ as the target class $t$ with a high probability, thus a high \underline{attack success rate (ASR)}.

\noindent \textbf{Attack Vectors.} In the context of transfer learning, we identify three potential backdoor threats to downstream tasks based on the adversary's capabilities and attack approaches. These threats are illustrated as follows:
\begin{itemize}[leftmargin=10pt]
\item \textbf{(\threatone) Encoder Poisoning:} The attacker of this type aims to inject the backdoor by producing a poisoned image encoder $g$ so that the downstream classifier $F$ built based on $g$ satisfies \cref{eq:acc,eq:asr}.
In this context, an attacker may operate as follows:
1) As a dishonest model provider, the attacker injects a backdoor into a pre-trained encoder and distributes it to users.
2) As a malicious third party, the attacker fine-tunes a clean encoder to include a backdoor and uploads it to platforms like HuggingFace.
3) As a disingenuous data provider, the attacker creates poisoned data for the pre-training dataset \(\predata\) of others, poisoning their encoder.
Either way, the attacker targets a specific concept associated with a downstream task.





\item \textbf{(\threattwo)  Dataset Poisoning:} Another type of attacker can directly or indirectly poison the training dataset $\downdata$ of the downstream user with a limited amount of trigger-injected data. This can occur when the downstream user collects data in an untrusted environment or from a disingenuous data provider. 
We assume the attacker of this type has no knowledge of the pre-trained encoder.
Even if the attacker knows the user is resource-limited and likely to use transfer learning, the architecture and parameters of the pre-trained encoder remain unknown.

\item \textbf{(\threatthree) Adaptive Poisoning:}  In an extreme and yet feasible case, an adaptive attacker could potentially compromise both the pre-trained encoder $g$ and the downstream dataset $\downdata$ using the same backdoor trigger.
This could happen if the user obtains both the model and training data from the same dishonest service provider.
The attacker maximizes effectiveness by applying the same trigger to both  $g$ and $\downdata$. Consequently, even if the model or data is purified, the backdoor may remain effective unless completely eliminated.

 \end{itemize}
We assume that in all these types of threats, the attacker does not know the training details of the downstream task. Once the poisoned data or encoder is delivered to the downstream user, how the user proceeds with the training or any potential defense is unknown to the attackers.
It is important to note that multiple independent attackers can operate simultaneously. For instance, \(\threatone\) and \(\threattwo\) may both occur: one attacker compromises the encoder \(g\) with trigger \(\trigger_1\) targeting class \(t_1\), while another contaminates the downstream training data \(\downdata\) with a different trigger \(\trigger_2\) targeting class \(t_2\). Each attacker remains unaware of the other's presence.


\noindent\textbf{Defender's Goals.} The defender we considered can be downstream users themselves or a cybersecurity service company like Darktrace\footnote{\url{https://darktrace.com}} that provides automatic \textbf{\textit{on-premise}} defense for users since the users may not be willing to adopt a cloud service and update their collected data, sometimes personal data.
The defense framework takes the untrusted encoder $g$, and the untrusted training set $\downdata$ as inputs and intends to build an accurate and safe classifier at the endpoint. The goals of the defense are three-fold as follows: 
\textbf{1)} \textbf{Utility}: the resulting classifier $F$ has a high ACC on the downstream task;
\textbf{2)} \textbf{Security}:  for any injected backdoor in $g$ and $\downdata$, its effect will be eliminated,  the resulting classifier itself $F$ can correctly classify according to the actual semantic of the input, \ie, exhibiting a low ASR without additional effect at the inference time;
\textbf{3)} \textbf{Generalizability}: The defense should effectively counter all backdoor threats using various trigger embedding methods \(\trigger\). Regardless of how the backdoor is injected, the defense methodology must maintain the effectiveness and security of the resulting \(F\) across different datasets, encoders, attack vectors, and hyperparameters (e.g., poison rate).

\noindent\textbf{Defender's Capabilities and Constraints.}  We illustrate the defender's knowledge and constraints below:
\begin{enumerate}[label=\arabic{*}), leftmargin=12pt]
    \item \underline{Access limited to $g$ and $\downdata$}: We consider the defender has no access to additional data other than $\downdata$, such as the pre-training dataset $\predata$ or a hold-out dataset containing a sufficient amount of clean samples.
    This is because model providers typically safeguard their pre-training dataset for proprietary or privacy reasons. Likewise, obtaining a completely clean hold-out dataset may not always be feasible.
    Nevertheless, we assume the defender possesses full autonomy over $\downdata$ and  $g$, which includes the ability to access, analyze, and modify their elements as needed.
    \item  \underline{Ignorance of threat model}: The defender is unaware of the specific kind of backdoor threat
    it is dealing with.
    In other words, whether the encoder $g$ or the dataset $\downdata$ has been poisoned remains uncertain. To the defender, it is possible that neither, either, or both the encoder $g$ and the dataset $\downdata$ have been poisoned.
    As such, the defender has to treat both $g$ and $\downdata$ as untrustworthy.
    \item \underline{Computational constraints}: 
    We assume the defense's computational power is limited, reflecting the reality of defense deployments on edge users' devices.
    Thus, computing gradients for the entire network is often infeasible due to the large size of pre-trained encoders and limited memory.
    This also agrees with the user's intention of adopting the transfer learning pipeline, where the fine-tuning process usually mainly focuses on a few layers that are appended to the encoder. Thus, designing a memory-efficient defense is in the defender's interest. Nevertheless, we assume the defense process can span a relatively long period to ensure the successful removal of potential backdoors from the fine-tuned model.

\end{enumerate}

\section{Methodological Analysis}\label{sec:methodology_analysis}

\begin{table*}[t]
\centering
\caption{An overview of existing backdoor defense methods, their assumptions or requirements, and how they fail or are not well-suited under the defense context.}
\label{tab:defense_types}
\begin{tabular}{p{2cm} p{3.8cm} p{5cm} p{5cm}}
\toprule
\textbf{Defense Type} & \textbf{Approach} & \textbf{Assumption or Requirement} & \textbf{Limitation under the Defense Context} \\
\midrule
\multirow{2}{*}{\textbf{Poison Detection}} &
  Latent separation-based detection &
  Poisoned samples can be distinguished as outliers in the latent space. &
  Poisoned samples blend into the clean samples in the latent space when only tuning the classification head. \\
& Confusion training-based detection &
  Only backdoor correlations are well-preserved after confusion training. & A huge proportion of clean samples still have minimal losses after confusion training when only tuning the classification head. \\
\cmidrule(l){1-4}
\multirow{2}{*}{\textbf{Poison Suppression}} &
  Training restricted suppression & Demanding the computational resource to sufficiently train an entire encoder or DNN. & Edge users' devices typically do not support computing the entire network's gradients. \\
& Spurious correlation-based suppression & Backdoor features are easier to learn compared to clean features. & The learning pace advantage between backdoor and clean data differs across various backdoor attacks and threat types. \\
\cmidrule(l){1-4}
\multirow{2}{*}{\textbf{Poison Removal}} &
  Fine-tuning-based removal or Trigger synthesis-based removal & Requiring a hold-out clean dataset to fine-tune the model or reverse the trigger. & An additional completely hold-out clean dataset may not be feasible. \\
& Lipschitz-based removal & The channels with larger upper bounds of Lipschitz constant are more likely to be backdoor-related. & There is no absolute correlation between the Triggered-Activation Change and the Lipschitz constant's upper bound. \\
\bottomrule
\end{tabular}
\vspace{-4mm}
\end{table*}

\subsection{Assessing Existing Defense Methodology}
In this section, we evaluate the current defense methodologies within the defense context of TL to assess their effectiveness. 
We aim to identify their limitations and uncover the necessities required for a successful defense in the given context.
Note that we do not elaborate the inference-time defenses, such as preprocessing that disrupt trigger-injected input before forwarding into the model \cite{doan2020februus,li2021backdoor,qiu2021deepsweep} or input-based detection that assesses and rejects malicious inputs \cite{liu2023detecting,guo2023scaleup,ted,hou2024ibdpsc}. 
This exclusion stems from our commitment to the model's inherent security and creating a backdoor-free classifier for reliable deployment. 
Also, inference-time defenses can drastically raise processing costs by up to two orders of magnitude \cite{liu2023detecting}, and they often fail to provide a satisfactory security-utility trade-off (see \cref{tab:inference-time-defense}).


Thus, we exhaustively explore the other types of defenses that can presumably mitigate the backdoor threats and can help to fulfill the defender's goals in the defense context:
(1) poison detection to identify backdoor poison data, (2) poison suppression to obstruct backdoor learning during training, and (3) poison removal to eliminate backdoor within a model after the attack. 
It is important to note that most of these defenses are proposed to address a specific backdoor threat, which only applies to part of the cases in the defense context. 
Therefore, we evaluate each defense fairly, focusing on its intended purpose or presumably solvable backdoor threat. 
From a defender's perspective in the defense context, we assess whether existing methods can achieve their intended goals and discuss their limitations.

In this section, we primarily utilize two standard backdoor dataset poisoning methods, BadNet, and Blended attacks for \threattwo. 
For \threatone~and \threatthree, we leverage two encoder poisoning methods, BadEncoder \cite{BadEncoder}, and DRUPE \cite{DRUPE}. 
In \threatthree, we use the trigger of backdoor encoders for downstream poisoning.
For both \threattwo~and \threatthree, 20\% of the samples from the target class are poisoned in the downstream dataset. More configuration details are included in Appendix-\ref{subsec:EncoderExplanation} and Appendix-\ref{subsec:DatasetPoisoningExplanation}.

\begin{table}[]
    \centering
    \caption{Poison detection and subsequent evaluation of fine-tuned model under \threattwo~with CIFAR-10 as the downstream dataset and STL-10 as pre-training dataset: 20\% samples of the target class are poisoned ones. Both CT and ASSET are granted 1000 clean samples. 
    } 
    \adjustbox{center}{
    \resizebox{0.45\textwidth}{!}{\large
    \setlength{\tabcolsep}{0.5pt}
    \begin{tabular}{@{}c|cccc|cccc@{}}
    \toprule
    \multirow{2}{*}{\textbf{Methods}} & \multicolumn{4}{c|}{\textbf{BadNets}} & \multicolumn{4}{c}{\textbf{Blended}} \\ \cmidrule{2-9} 
    & \textbf{TPR$\uparrow$} & \textbf{FPR$\downarrow$} & \textbf{ACC$\uparrow$} & \textbf{ASR$\downarrow$} & \textbf{TPR$\uparrow$} & \textbf{FPR$\downarrow$} & \textbf{ACC$\uparrow$} & \textbf{ASR$\downarrow$} \\ \midrule
    \textbf{No Defense} & - & - & 85.04 & 92.21 & - & - & 84.84 & 89.12 \\ \hline
    \textbf{STRIP} & 6.08 & 6.18 & 82.80 & 87.16 & 6.00 & 5.15 & 83.24 & 83.01 \\ \hline
    \textbf{AC} & 71.92 & 5.61 & 82.88 & 76.02 & 0.00 & 6.86 & 83.74 & 82.48 \\ \hline
    \textbf{Spectral} & 49.76 & 37.19 & 78.33 & 59.24 & 64.48 & 36.81 & 78.44 & 39.21 \\ \hline
    \textbf{SPECTRE} & 22.16 & 37.89 & 76.54 & 74.80 & 29.44 & 37.71 & 76.40 & 68.60 \\ \hline
    \textbf{CT} & 0.00 & - & 85.04 & 92.21 & 0.00 & - & 84.84 & 89.12 \\ \hline
    \textbf{ASSET} & 32.16 & 1.23 & 84.28 & 64.68 & 22.88 & 1.46 & 83.76 & 87.10 \\ \bottomrule
    \end{tabular}
    }
}
\vspace{-5mm}
\label{tab:poison_detection}
\end{table}

\subsection{Defense Type I: Poison Detection}
\label{sec:poison_detection}
Poison detection involves identifying and removing abnormal samples from a backdoor dataset to ensure the creation of a purified dataset, which is then used to train a clean model \cite{tran2018spectral,chen2018detecting,tang2021demon,qi_sec_2023,gao2019strip,pan2023asset}. 
Thus, poison detection techniques are meant to address dataset poisoning of \threattwo.
However, one would face a dilemma when adopting prior techniques for this defense context.

\textit{First}, almost all prior literature on poison detection relies on a preset end-to-end supervised learning (SL) approach, assuming the training of the entire network is feasible\footnote{ASSET, the only method claiming effectiveness on TL, still requires training an additional full DNN.}. These approaches initialize and train an end-to-end DNN \textit{from scratch} on the entire poisoned dataset, identifying and removing backdoor samples based on their distinctive characteristics in the post-attacked model. Most focus on feature characteristics, claiming poisoned samples can be identified as outliers in the latent space, as demonstrated in the BadNets detection of STRIP \cite{gao2019strip}, AC \cite{chen2018detecting}, Spectral \cite{tran2018spectral}, and SPECTRE \cite{hayase21a}. 
However, this assumption is invalidated by \textit{latent space adaptive attacks} \cite{tang2021demon,qi2023revisiting,qi2023revisiting}.
\textit{Second}, to effectively filter out poisoned samples, many methods \cite{ted,qi_sec_2023,ma2022beatrix,gao2019strip,pan2023asset} require a certain number of clean samples to build their outlier detectors, which can conflict with the defender's constraints in our context. This includes state-of-the-art poison detection methods, CT \cite{qi_sec_2023} and ASSET \cite{pan2023asset}, which do not rely on latent separability. Both adopt a confusion training approach, optimizing over a hold-out clean set and the poisoned training dataset in the reverse direction. 
For instance, CT aims to reduce the model's accuracy on clean samples by unlearning on clean data while learning from the training set, leading it to focus primarily on the backdoor data. 
Then, samples predicted correctly or with the smallest loss are considered poisoned \cite{qi_sec_2023, pan2023asset}.


\begin{figure}[pt]
    \centering
    \subfloat[BadNets]{
        \includegraphics[width=0.22\textwidth]{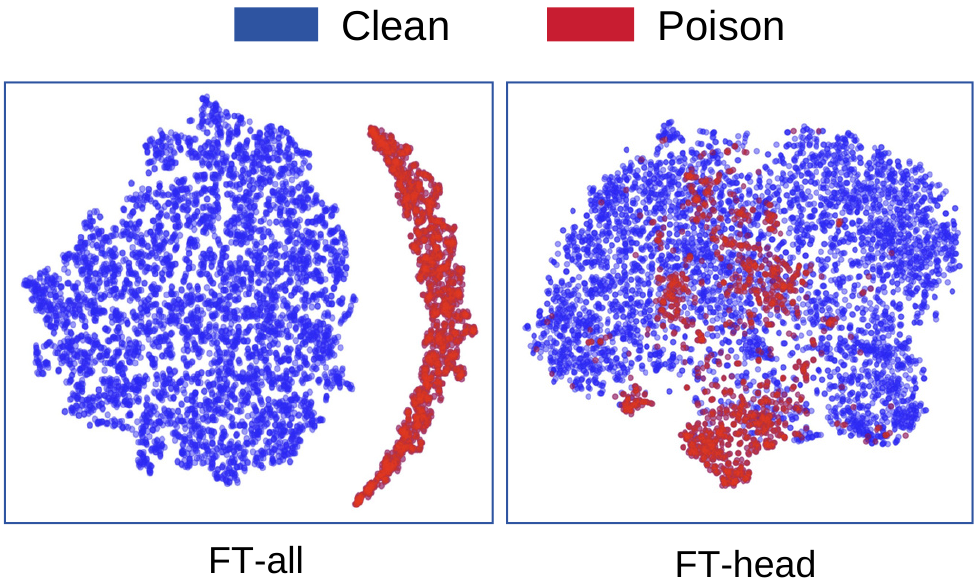}
        \label{fig:poision_dection_sub1}
    }
    \subfloat[Blended]{
        \includegraphics[width=0.22\textwidth]{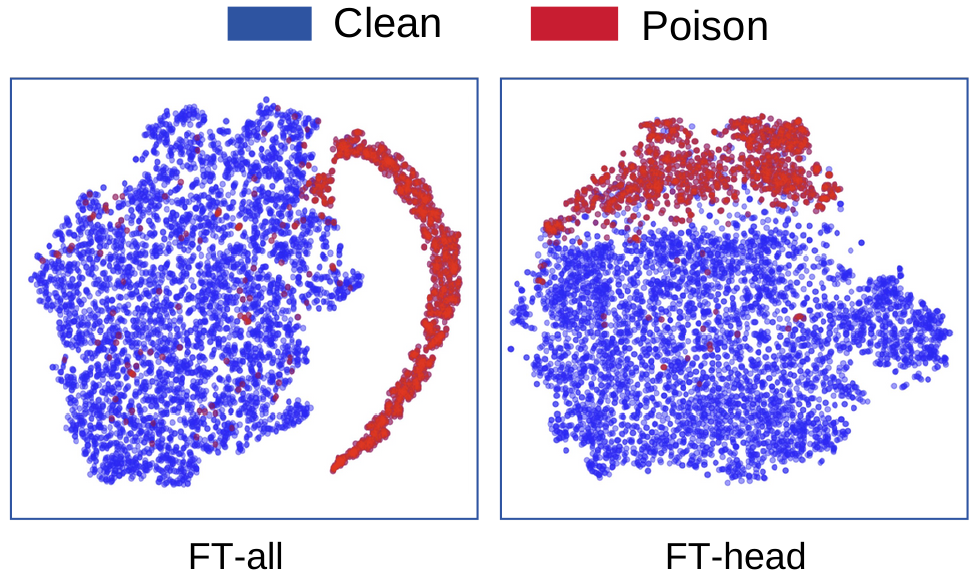}
        \label{fig:poison_dection_sub2}
    }
\caption{t-SNE comparison of feature space from a model trained on poisoned CIFAR-10: contrasting fine-tuning the entire network (FT-all) with fine-tuning only the 3-layer classification head (FT-head) under \threattwo.}
\label{fig:SL_and_fthead}
\vspace{-7mm}
\end{figure}

Despite assuming a clean hold-out set, we still evaluate the effectiveness of CT, ASSET, and others in detecting poisoned samples, measuring True Positive Rate (TPR) and False Positive Rate (FPR) in the context of \threattwo~under the vanilla BadNets \cite{gu2019badnets} and Blended \cite{chen2017targeted} attacks. We further fine-tune models on each purified dataset for subsequent assessment of ASR and ACC,
with model training adjusted for the classification head to fit the defense context. 
However, as shown in \cref{tab:poison_detection}, none of these backdoor detection methods achieve a sufficient TPR or produce a clean classifier free from the backdoor of \threattwo. 

\begin{figure}[t]
    \centering
    \subfloat[BadNets]{
        \includegraphics[width=0.22\textwidth]{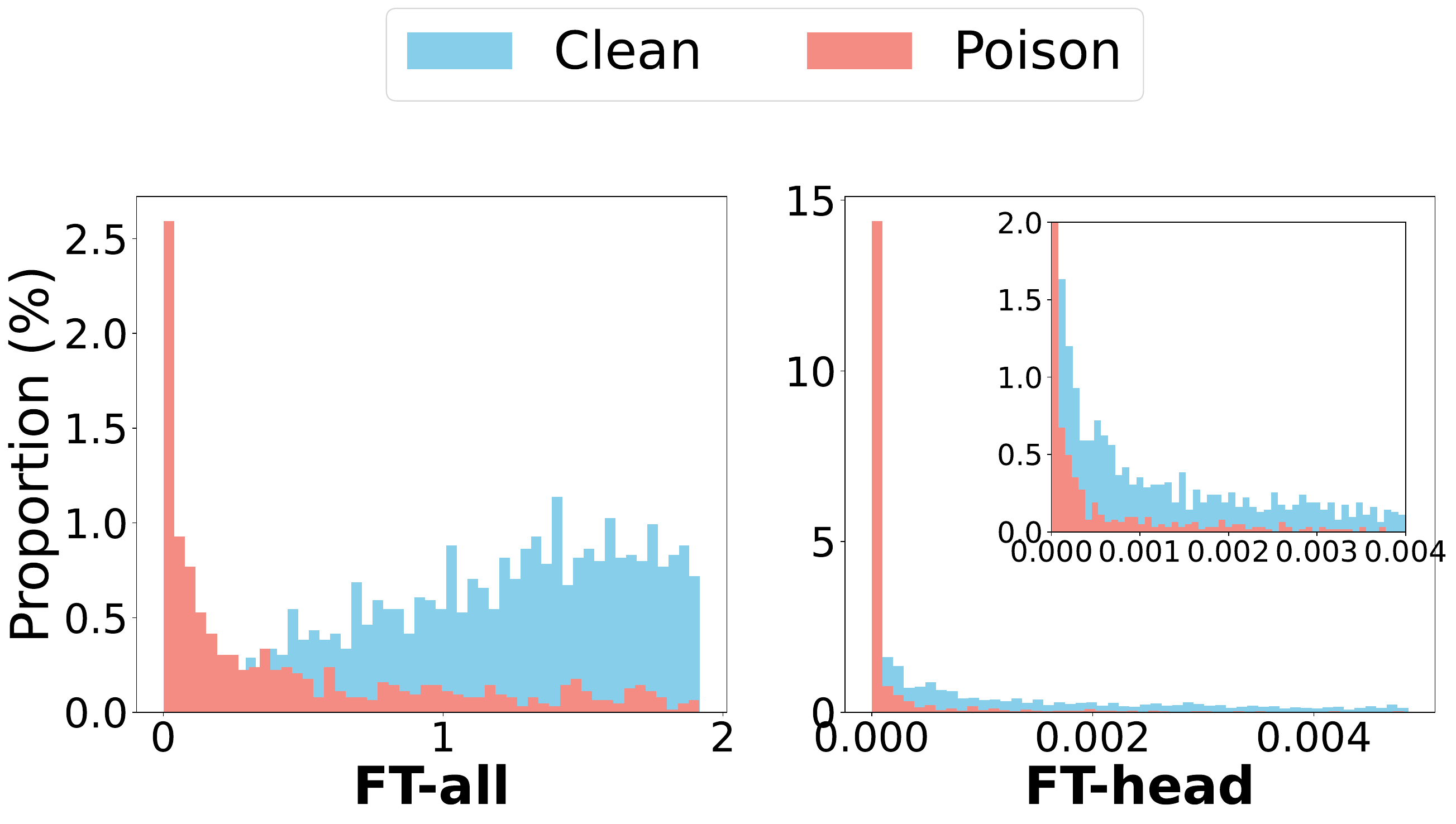}
    }
    \subfloat[Blended]{
        \includegraphics[width=0.22\textwidth]{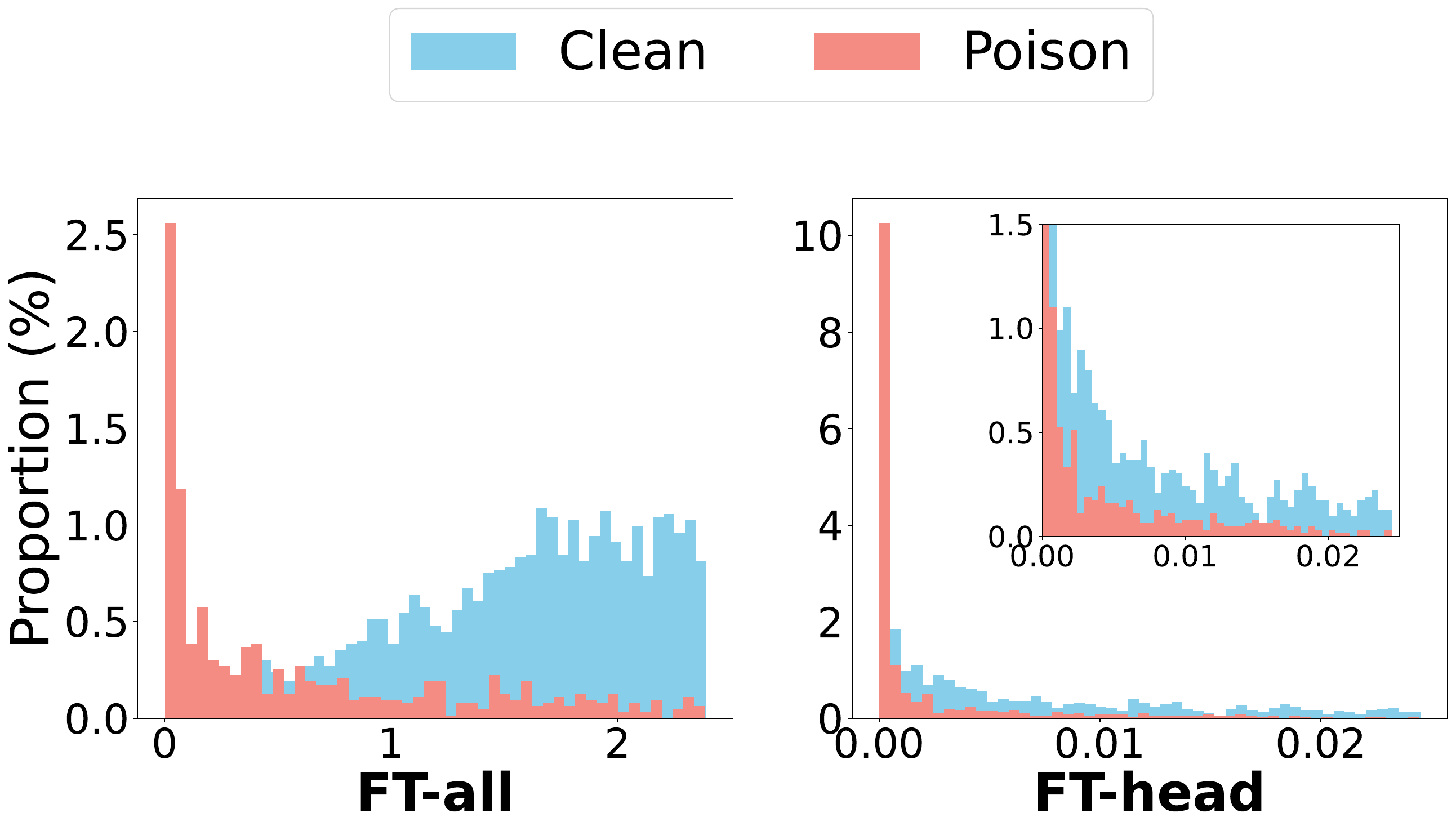}
    }
\caption{Distribution of poisoned and clean samples in the low-loss region (\textbf{lowest 40\% loss} of the training set) after Confusion Training (CT), 
contrasting results from fine-tuning the entire network $h$ (FT-all) and just the 3-layer classification head $f$ (FT-head) under \threattwo.}
    \label{fig:ct_low_loss}

\vspace{-4mm}
\end{figure}

We now examine why these detection methodologies are inadequate in this TL setting. In this context, the encoder is inherited with knowledge from pre-training, and fine-tuning is limited to the linear layers of the classification head.
As a result, features from the penultimate layer, which the latent separation-based detector depends on, are merely a linear transformation of the fixed encoder's output, leading to a condensed feature space that complicates the distinction between clean and poisoned samples.
In \cref{fig:SL_and_fthead}, we show that if we fine-tune the entire network on the poisoned dataset, a clear boundary between poisoned and clean samples would still exist. However, when only the classification head is tuned, this distinction blurs or disappears. 
Furthermore, the narrow optimization space limits the ability to offset clean features in the poisoned set under CT and ASSET. As illustrated in \cref{fig:ct_low_loss}, the low-loss region in the fine-tuned head scenario contains many clean samples, making them indistinguishable from poisoned samples.
These results highlight the challenges of eliminating poisoned samples from a poisoned dataset in this defense context.
\begin{mtbox}{Remark}
The constraint on training the entire network also constrains the ability to detect backdoor-poisoned samples, as their supposedly distinguishable characteristics become indistinct. 
Detecting poisoned samples based on some assumption about them is unreliable because it may not scale across different training paradigms.
\end{mtbox}

\begin{table*}[t]
    \centering
    \begin{minipage}[t]{0.60\textwidth}
        \centering
        \caption{ABL's defense performance (ACC and ASR) under \threattwo~and \threatthree~using GTSRB as the downstream dataset and CIFAR-10 as the pre-training dataset while varying the isolation ratio. We also report the \textit{number of isolation data} (NID) and the \textit{number of poison data} (NPD) within the isolation data for each ratio.}
        \adjustbox{center}{
        \resizebox{0.96\textwidth}{!}{\large
        \setlength{\tabcolsep}{0.5pt}
        \begin{tabular}{@{}c|cccccccc|cccccccc@{}}
        \toprule
        \multicolumn{1}{l|}{Threat Type} & \multicolumn{8}{c|}{\threattwo} & \multicolumn{8}{c}{\threatthree} \\ \midrule
        \multirow{2}{*}{Isolation Ratio} & \multicolumn{4}{c|}{\textbf{BadNets}} & \multicolumn{4}{c|}{\textbf{Blended}} & \multicolumn{4}{c|}{\textbf{BadEncoder}} & \multicolumn{4}{c}{\textbf{DRUPE}} \\ \cmidrule{2-17} 
        & \textbf{ACC$\uparrow$} & \textbf{ASR$\downarrow$} & \textbf{NID} & \multicolumn{1}{c|}{\textbf{NPD}} & \textbf{ACC$\uparrow$} & \textbf{ASR$\downarrow$} & \textbf{NID} & \textbf{NPD} & \textbf{ACC$\uparrow$} & \textbf{ASR$\downarrow$} & \textbf{NID} & \multicolumn{1}{c|}{\textbf{NPD}} & \textbf{ACC$\uparrow$} & \textbf{ASR$\downarrow$} & \textbf{NID} & \textbf{NPD} \\ \midrule
        No Defense & 77.51 & 91.12 & - & \multicolumn{1}{c|}{-} & 77.57 & 83.14 & - & - & 78.10 & 99.97 & - & \multicolumn{1}{c|}{-} & 74.42 & 98.58 & - & - \\ \hline
        0.01 & 46.70 & 64.92 & 392 & \multicolumn{1}{c|}{0} & 46.63 & 49.83 & 392 & 0 & 44.74 & 98.29 & 392 & \multicolumn{1}{c|}{0} & 20.44 & 96.20 & 392 & 1 \\ \hline
        0.05 & 38.30 & 61.78 & 1960 & \multicolumn{1}{c|}{0} & 39.03 & 57.69 & 1960 & 0 & 42.05 & 97.96 & 1960 & \multicolumn{1}{c|}{401} & 17.22 & 1.26 & 1960 & 458 \\ \hline
        0.10 & 27.90 & 72.45 & 3920 & \multicolumn{1}{c|}{36} & 25.66 & 66.45 & 3920 & 0 & 30.14 & 95.86 & 3920 & \multicolumn{1}{c|}{519} & 13.51 & 0.37 & 3920 & 464 \\ \hline
        0.15 & 17.50 & 83.70 & 5881 & \multicolumn{1}{c|}{147} & 16.48 & 30.40 & 5881 & 0 & 17.51 & 0.00 & 5881 & \multicolumn{1}{c|}{520} & 11.03 & 0.40 & 5881 & 480 \\ \hline
        0.20 & 14.26 & 68.12 & 7841 & \multicolumn{1}{c|}{146} & 10.37 & 38.70 & 7841 & 122 & 13.62 & 0.03 & 7841 & \multicolumn{1}{c|}{520} & 7.07 & 0.02 & 7841 & 482 \\ \bottomrule
        \end{tabular}
        \label{tab:abl_results}
        }
        }
    \end{minipage}
    \hspace{0.07cm}
    \begin{minipage}[t]{0.38\textwidth}
        \centering
\caption{CBD's defense performance (ACC and ASR) under \threattwo~and \threatthree~using GTSRB as the downstream dataset and CIFAR-10 as the pre-training dataset while varying the training epochs of the initial backdoor model.}
        \adjustbox{center}{
        \resizebox{1\textwidth}{!}{\large
        \setlength{\tabcolsep}{0.5pt}
        \begin{tabular}{@{}c|cccc|cccc@{}}
        \toprule
        Threat Type & \multicolumn{4}{c|}{\threattwo} & \multicolumn{4}{c}{\threatthree} \\ \midrule
        \multirow{2}{*}{\makecell{Epoch Number of \\Initial Model}} & \multicolumn{2}{c|}{\textbf{BadNets}} & \multicolumn{2}{c|}{\textbf{Blended}} & \multicolumn{2}{c|}{\textbf{BadEncoder}} & \multicolumn{2}{c}{\textbf{DRUPE}} \\ \cmidrule{2-9} 
        & \textbf{ACC$\uparrow$} & \multicolumn{1}{c|}{\textbf{ASR$\downarrow$}} & \textbf{ACC$\uparrow$} & \textbf{ASR$\downarrow$} & \textbf{ACC$\uparrow$} & \multicolumn{1}{c|}{\textbf{ASR$\downarrow$}} & \textbf{ACC$\uparrow$} & \textbf{ASR$\downarrow$} \\ \midrule
        No Defense & 79.15 & \multicolumn{1}{c|}{96.36} & 77.08 & 92.10 & 79.41 & \multicolumn{1}{c|}{99.71} & 74.71 & 99.30 \\ \hline
        1 & 75.14 & \multicolumn{1}{c|}{93.75} & 74.58 & 89.78 & 75.80 & \multicolumn{1}{c|}{99.48} & 45.60 & 96.97 \\ \hline
        3 & 72.90 & \multicolumn{1}{c|}{93.53} & 73.64 & 89.53 & 72.06 & \multicolumn{1}{c|}{98.47} & 44.37 & 95.91 \\ \hline
        5 & 68.91 & \multicolumn{1}{c|}{92.49} & 64.73 & 88.18 & 61.47 & \multicolumn{1}{c|}{87.65} & 42.87 & 94.86 \\ \hline
        7 & 63.19 & \multicolumn{1}{c|}{78.64} & 57.51 & 88.64 & 54.04 & \multicolumn{1}{c|}{0.19} & 39.46 & 93.28 \\ \hline
        9 & 61.46 & \multicolumn{1}{c|}{77.48} & 51.93 & 54.88 & 53.38 & \multicolumn{1}{c|}{0.75} & 38.78 & 92.78 \\ \bottomrule
        \end{tabular}
        \label{tab:cbd_results}
        }
        }
    \end{minipage}
    \label{tab:performance_comparison}
    \vspace{-4mm}
\end{table*}

\subsection{Defense Type II: Poison Suppression} \label{sec:poison_suppression}
Poison suppression reduces the impact of poisoned samples during training, allowing for direct learning of a clean model under dataset poisoning as \threattwo~without needing access to clean datasets.
However, we found that many defenses of this type, by design, require training the entire network, limiting their applicability in the defense context.
For instance, DBD \cite{DBD} explicitly demands a self-supervised learning process over the entire encoder for semantic clustering.
Other defenses \cite{NONE,PIPD} observe that backdoors increase the linearity through activation functions, thus filtering out potentially affected neurons with more linearity during training.
As a result, they depend on monitoring specific non-linear layers and activation functions during training, but fine-tuning only the classification head does not affect them.

The two poison suppression methods suitable for the defense context are \textit{Anti-Backdoor Learning} (ABL) \cite{ABL} and \textit{Causality-inspired Backdoor Defense} (CBD) \cite{CBD}, both aim to suppress the model's \textit{spurious correlations} to mitigate backdoor attacks. Since DNNs often confuse causal relationships with statistical associations, favoring easier correlations, ABL and CBD target features that are easier to learn.
ABL identifies samples with rapidly decreasing training losses and applies unlearning to these easy-to-learn samples. In contrast, CBD first trains an initial backdoor model on the poisoned dataset for a few epochs, allowing backdoor features to be learned while clean features remain underdeveloped. Then, it trains a clean model from the poisoned set while minimizing the mutual information between its representations and those of the initial backdoor model.

Since both ABL and CBD assume that backdoor features are learned faster, we infer that they may also be suitable for the case when both the encoder and downstream dataset contain the same trigger. Thus, we evaluate them under both \threattwo~and \threatthree. We adjust their key hyperparameters to assess the security-utility trade-off.
For ABL, the isolation ratio indicates the percentage of lowest-loss training samples selected for unlearning. For CBD, the training epoch number indicates how well backdoor's spurious correlation is captured.
According to the results in \cref{tab:abl_results,tab:cbd_results}, neither ABL nor CBD can achieve a high ACC and low ASR at any point. 
ABL shows much lower ACC compared to when the defense is inactive, with inadequate ASR reduction, as it isolates clean samples instead of poisoned ones and fails to isolate poisoned samples at small isolation ratios (0.01 and 0.05) under \threattwo.
CBD consistently decreases both ACC and ASR as the training epochs increase, indicating entanglement between poison and clean features in the representation, and minimizing mutual information could harm both.



    

\begin{figure}[hb]
    \vspace{0pt}
    \centering
    \begin{minipage}[htbp]{0.6\textwidth}

        \subfloat[\scalebox{1}{BadNets}]{%
            \includegraphics[width=0.45\textwidth]{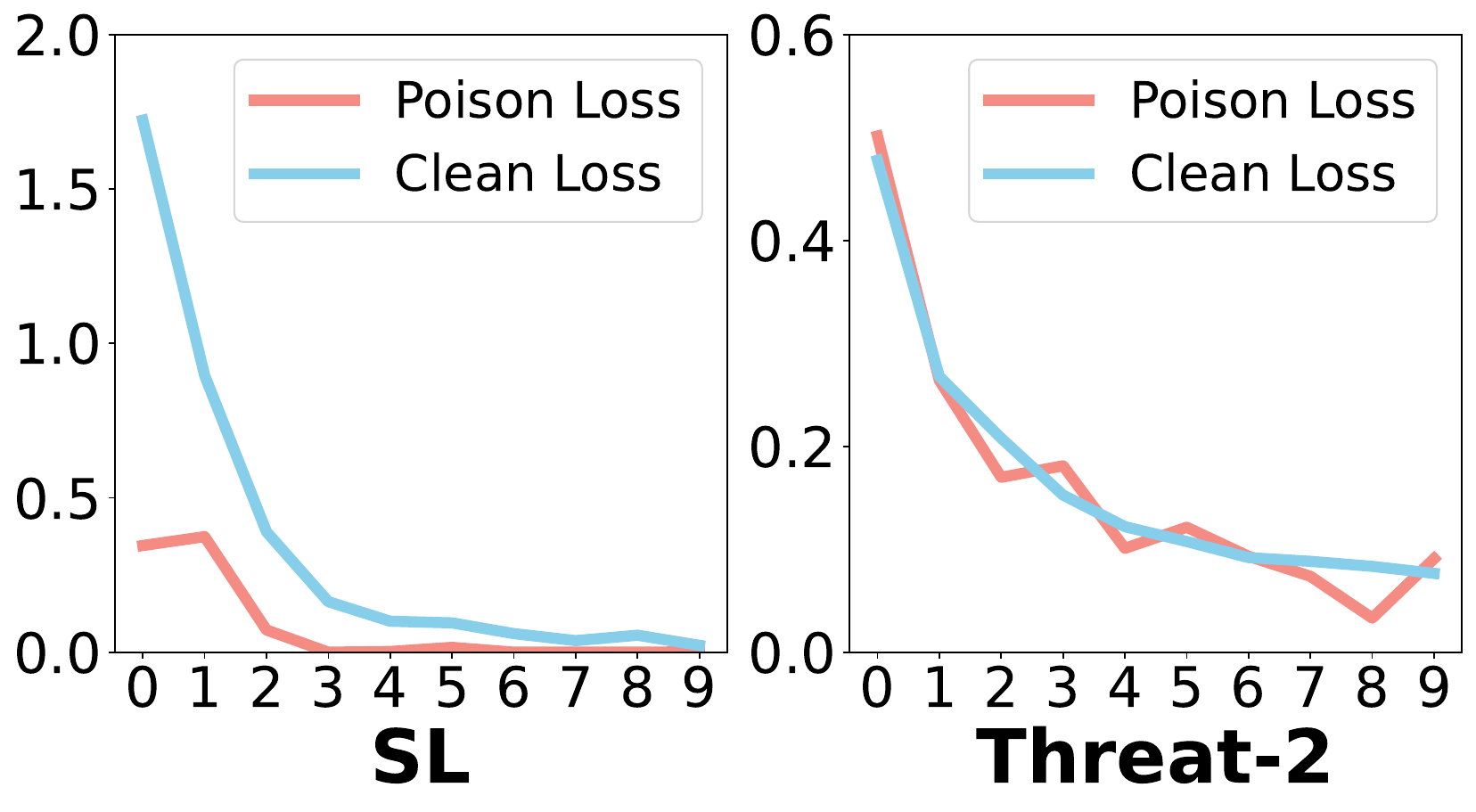} 
        }
        \hspace{0.3cm}
        \subfloat[\scalebox{1}{BadEncoder}]{
            \includegraphics[width=0.245\textwidth]{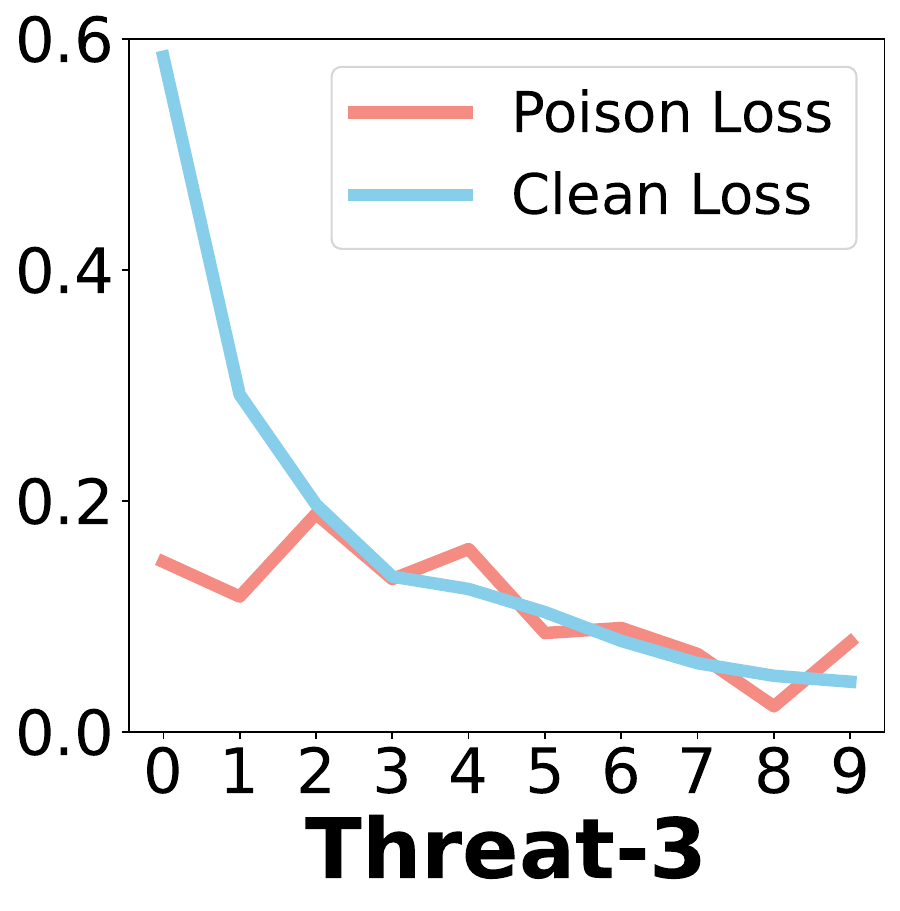}
        }
    \end{minipage}

    \begin{minipage}[t]{0.6\textwidth}
    
        \subfloat[\scalebox{1}{Blended}]{%
            \includegraphics[width=0.45\textwidth]{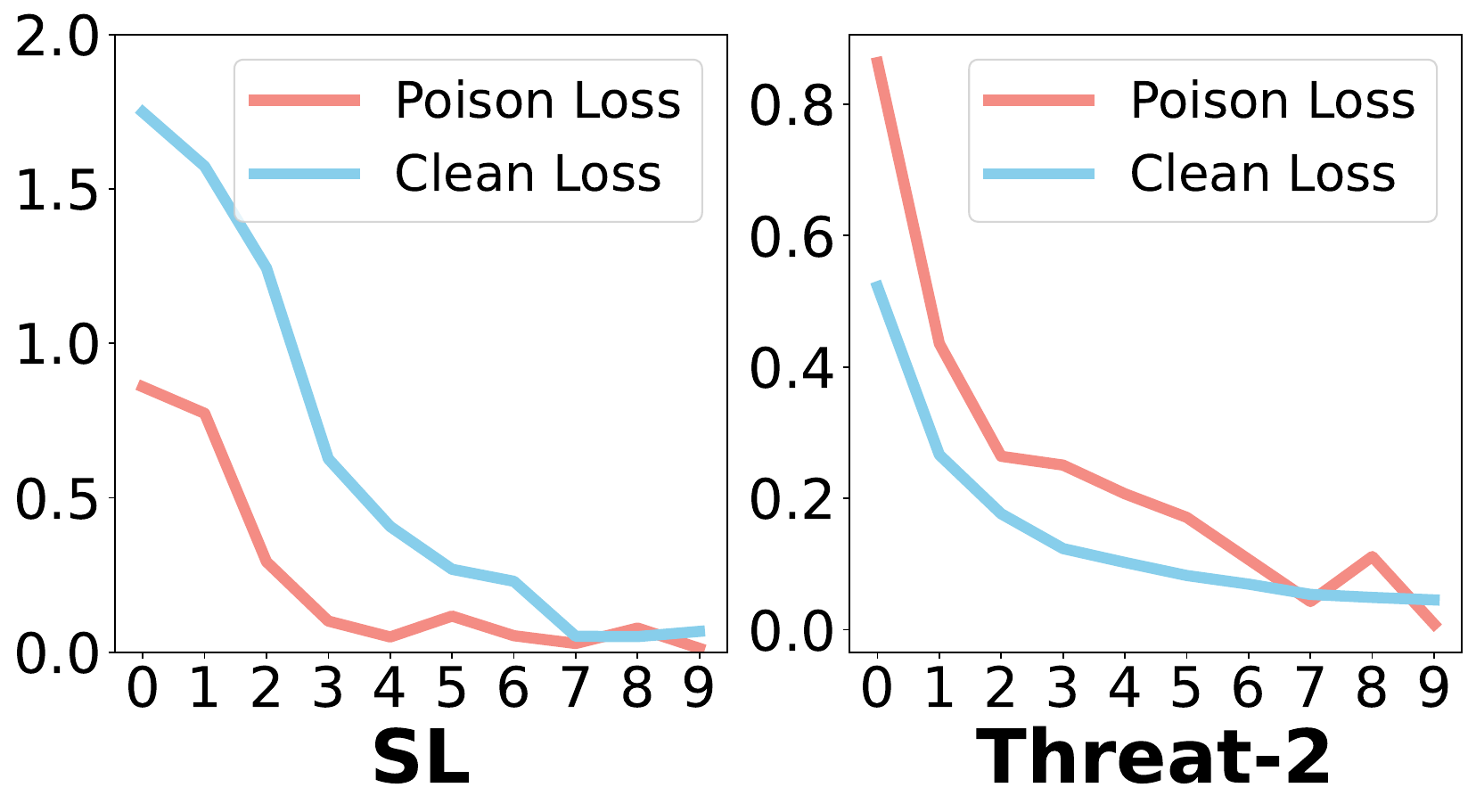} 
        }
        \hspace{0.3cm}
        \subfloat[\scalebox{1}{DRUPE}]{
            \includegraphics[width=0.245\textwidth]{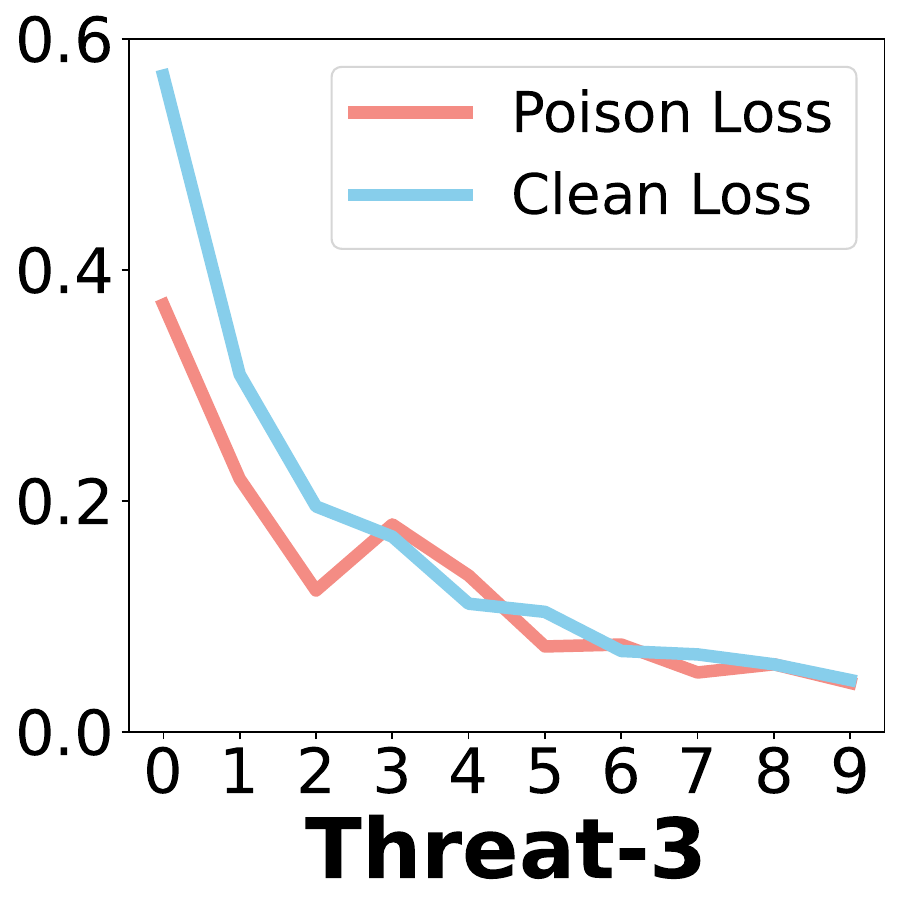}
        }
    \end{minipage}
    \caption{Comparison of average training losses for poisoned and clean samples in early epochs: \textbf{SL} trains from scratch on a poisoned dataset (BadNets or Blended). \threattwo~fine-tunes a classifier head after a clean encoder on a poisoned dataset. \threatthree~fine-tunes a classifier head after a poisoned encoder (BadEncoder or DRURE) on a poisoned dataset with the same trigger.
    }
    \label{fig:loss_trajectory}
    \vspace{-5mm}
\end{figure}

To analyze their failure, we plot the loss trajectories under end-to-end supervised learning (SL), \threattwo, and \threatthree. 
\cref{fig:loss_trajectory} shows that the model learns backdoor data faster than clean data in the SL. Overall, in the TL, the model learns both types of data faster. Under \threatthree, poisoned data is learned faster than clean data only in the initial epochs. Under \threattwo, the model learns poisoned and clean data at similar paces for BadNets but learns clean data faster for Blended.
Thus, there is no consistent evidence to support that backdoor features are easier to learn than clean features, indicating that both ABL and CBD target inaccurate objects for suppression, whether sample-wise or representation-wise.

\begin{mtbox}{Remark}
The shift of the training paradigm also dramatically shifts the learning dynamic of both backdoor features and clean features.
Locating the accurate ``poison'' for poison suppression based on some assumptions about how they are learned differently is also unreliable, especially in such a complex scenario where diverse triggers and backdoor threats are possible.
\end{mtbox}

\subsection{Defense Type III: Poison Removal}  \label{sec:poison_removal}
Poison removal works during the post-attack period, aiming to reconstruct a clean model by directly modifying the backdoor model, regardless of how the backdoor was injected, which makes it suitable to address \threatone, \threattwo, and \threatthree~all. However, most backdoor removal methods also rely on a clean dataset, including fine-tuning methods that use clean data to adjust the model \cite{sam_backdoor,ibau,fine_pruning} and trigger synthesis methods that generate potential backdoor patterns from clean data \cite{guo2020towards,shen2021backdoor,neuroncleanse,feng2023detecting}. The only method that operates independently of clean data is Channel-wise Lipschitz Pruning (CLP) \cite{CLP}.

CLP is based on an observation that channels associated with backdoors exhibit greater changes in activation when presented with backdoor samples. 
Specifically, the \textit{Trigger-Activated Change} (TAC) \( TAC_k^{(l)}(x) = \| F_k^{(l)}(x) - F_k^{(l)}(\trigger(x)) \|_2 \) should be larger for backdoor-related channels than for normal ones, where \( F_k^{(l)}(\cdot) \) is the output of the \( k \)-th channel at the \( l \)-th layer for input \( x \).
Additionally, the Lipschitz-continuous function \( h_k^{(l)}(\cdot) \) has an \textit{upper bound} of \textit{Channel Lipschitz Constant} (UCLC) \( \sigma(\theta_k^{(l)}) \), which limits output differences.
For any inputs \( z_1 \) and \( z_2 \) at layer \( l \), \( \| h_k^{(l)}(z_1) - h_k^{(l)}(z_2) \|_2 \leq \textrm{UCLC} \), and \(\sigma(\theta_k^{(l)}) \) is the spectral norm of model parameters \( \theta_k^{(l)} \) at \( k \)-th channel of \( l \)-th layer.
Since TAC is untraceable due to the unknown backdoor trigger \( \trigger \), CLP identifies channels with high \( \sigma(\theta_k^{(l)}) \) values as backdoor-related. A larger \( \sigma(\theta_k^{(l)}) \) indicates more room for activation changes when a backdoor trigger is present, suggesting a higher likelihood of being backdoor-related.

\begin{figure*}[t]
    \centering
    \begin{subfigure}{0.32\textwidth}
        \centering
        \includegraphics[width=\linewidth]{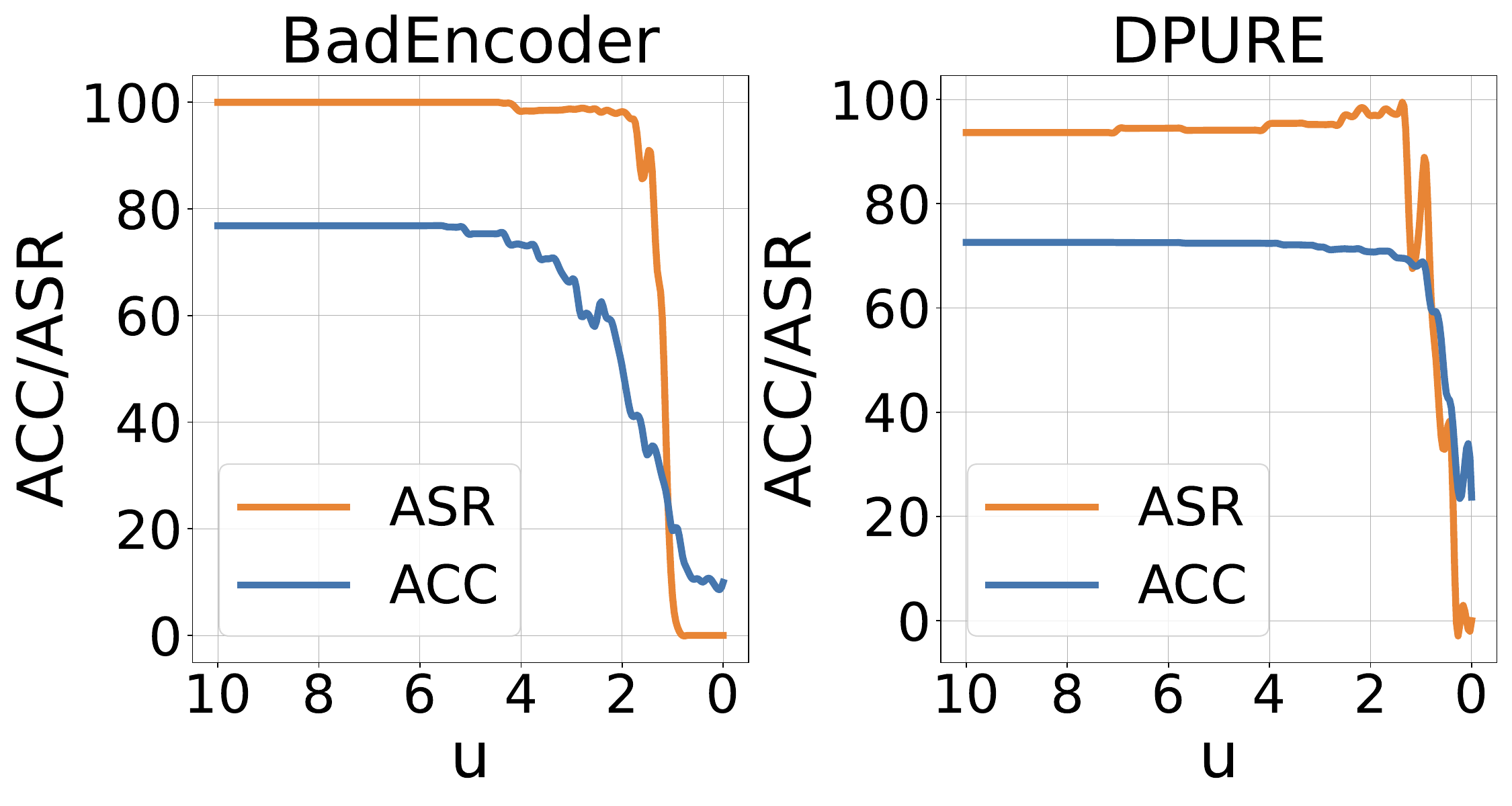}
        \caption{
        \threatone
        }
    \end{subfigure}%
    \hfill
    \begin{subfigure}{0.32\textwidth}
        \centering
        \includegraphics[width=\linewidth]{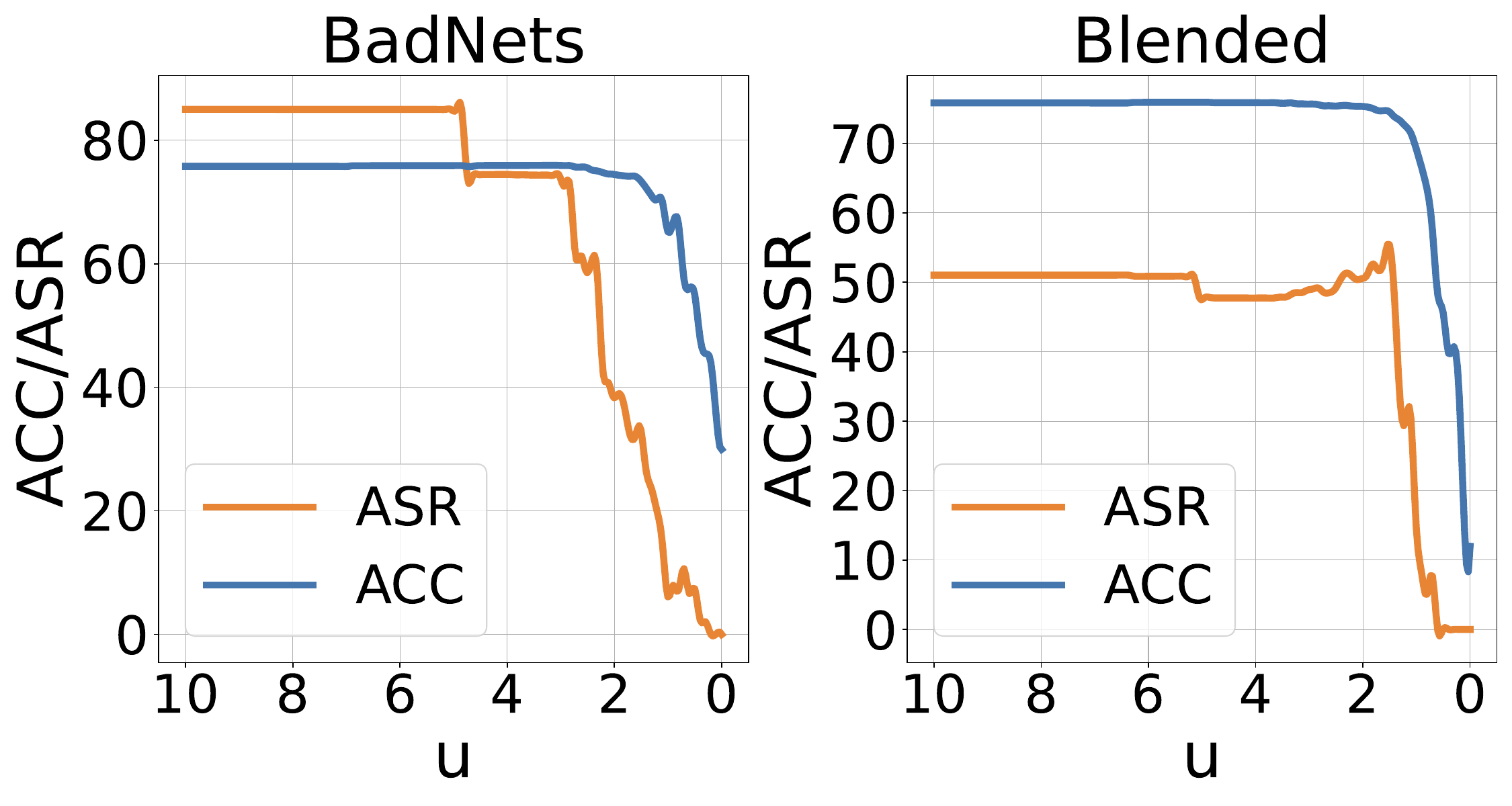}
        \caption{\threattwo}
    \end{subfigure}%
    \hfill
    \begin{subfigure}{0.32\textwidth}
        \centering
        \includegraphics[width=\linewidth]{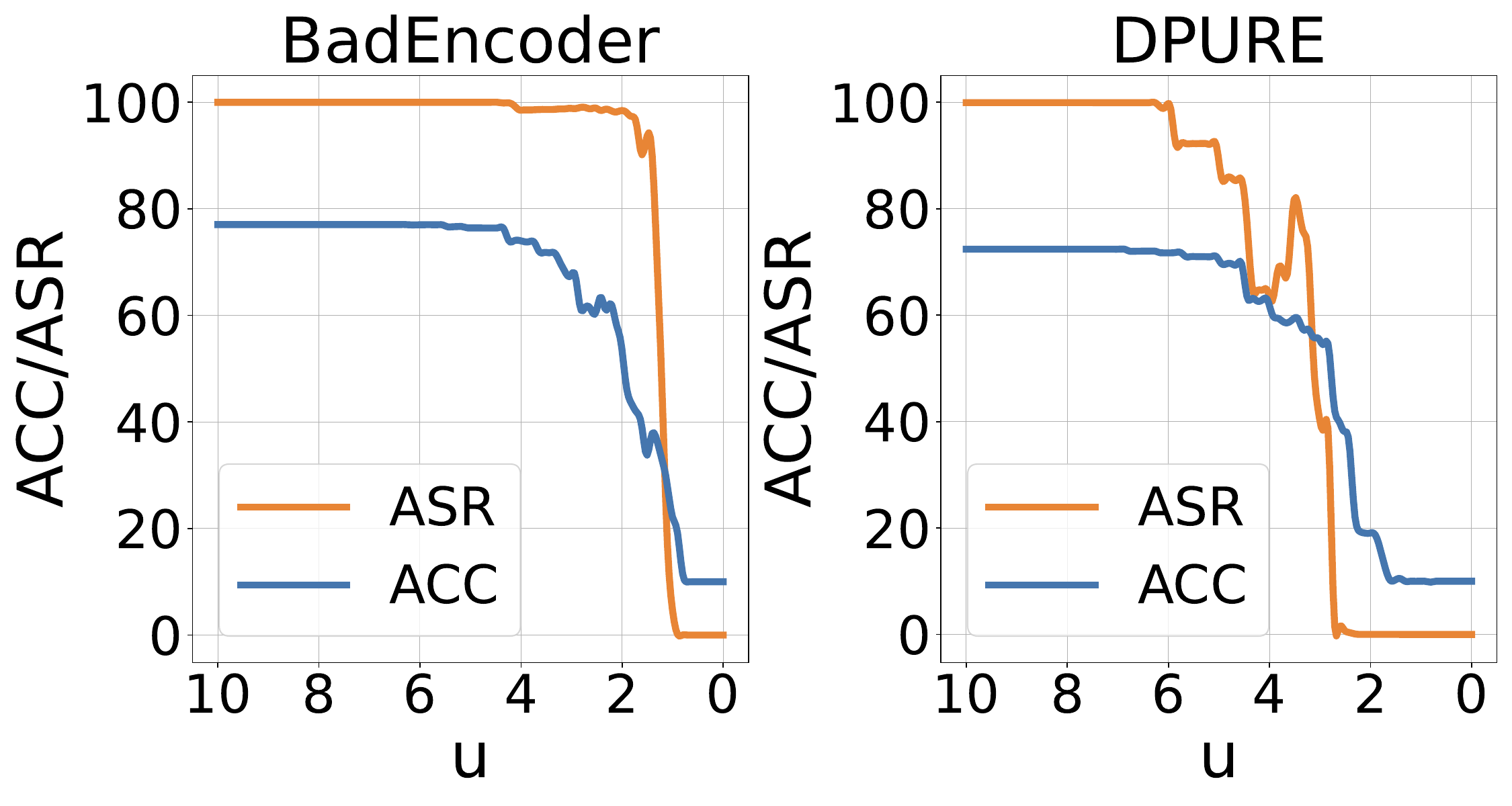}
        \caption{\threatthree}
    \end{subfigure}
    \caption{CLP performance on different types of threat from an \textbf{omniscient} defender's perspective: (a) CLP is applied to the encoder $g$ only because the pre-trained encoder is poisoned and the downstream dataset $\downdata$ is clean under $\threatone$; (b) CLP is applied to the linear layers of the classification head $f$ because the encoder $g$ is clean and only $f$ is fine-tuned over a poisoned  $\downdata$  under $\threattwo$; (c) CLP is applied to $f$ and $g$ since both encoder and dataset are poisoned under $\threatthree$.
    }
    \label{fig:CLP_allthreat}
    \vspace{-4mm}
\end{figure*}


For each layer, CLP sets the threshold for the \( l \)-th layer with \( K \) channels as \( \mu^{(l)} + u \cdot s^{(l)} \), where \( \mu^{(l)} \) and \( s^{(l)} \) are the mean and standard deviation of the layer's \( \{ \sigma(\theta_k^{(l)}): k=1, 2,\dots, K \} \). 
A smaller \( u \) results in more channels being pruned. 
We evaluate CLP's effectiveness by varying \( u \) from 0 to 10 in increments of 0.1 across all three backdoor threat types.
Note that the original CLP is only employed on the convolution operators.
In this experiment, we apply CLP to $g$, $f$, or both, depending on the specific threat model from an omniscient defense perspective.
As shown in \cref{fig:CLP_allthreat}, CLP fails to achieve low ASR and high ACC for any \( u \) across all threat types. 
The ACC and ASR descend almost together with the decline of $u$, suggesting that wherever the backdoor is injected, the channels of the backdoor do not depend exclusively on the ones with the larger upper bound on activation changes (UCLC).

\begin{figure}[ht]
    \centering
    \subfloat[BadNets]{\includegraphics[width=0.24\textwidth]{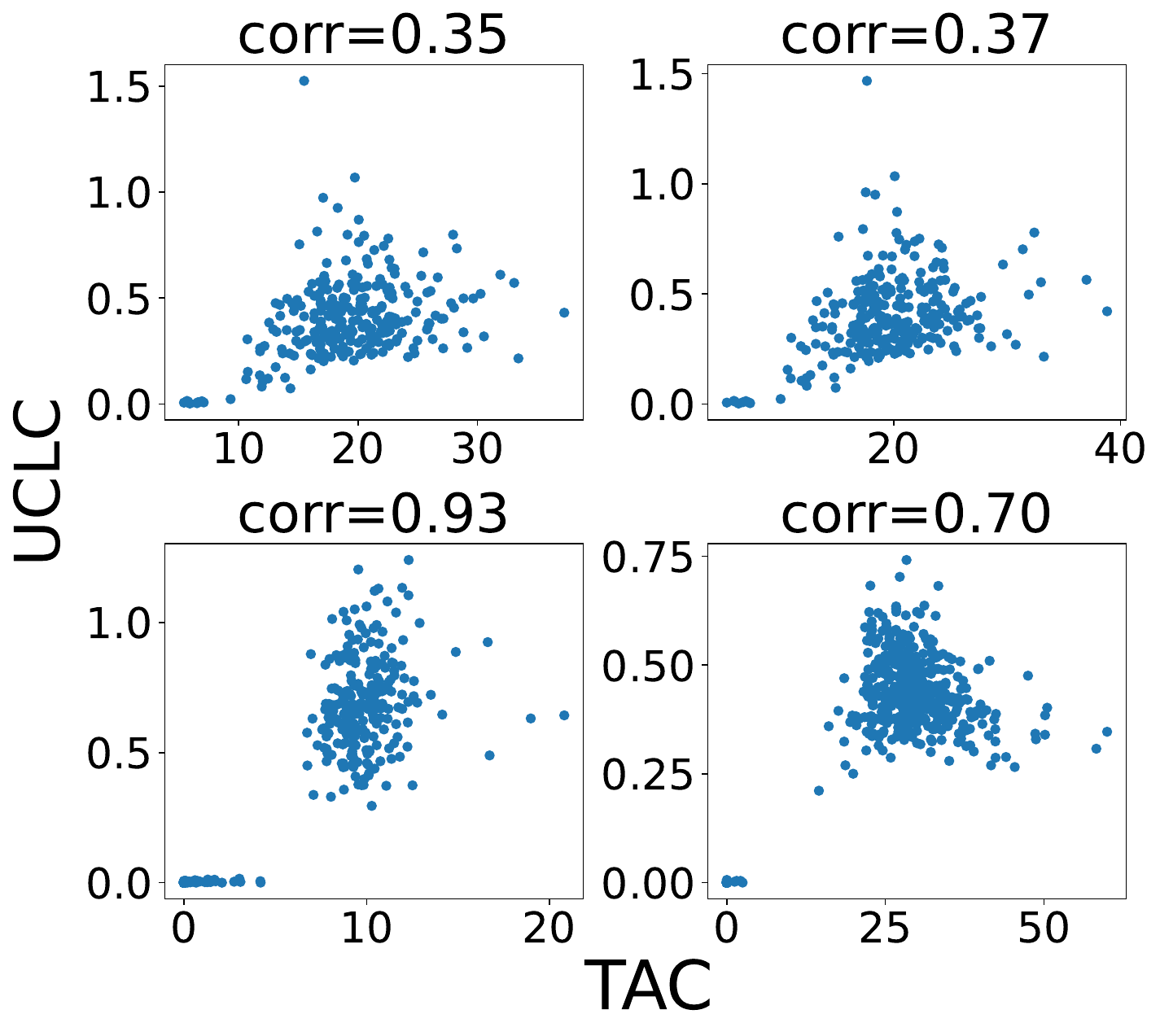}}
    \hfill
    \subfloat[DRUPE]{\includegraphics[width=0.24\textwidth]{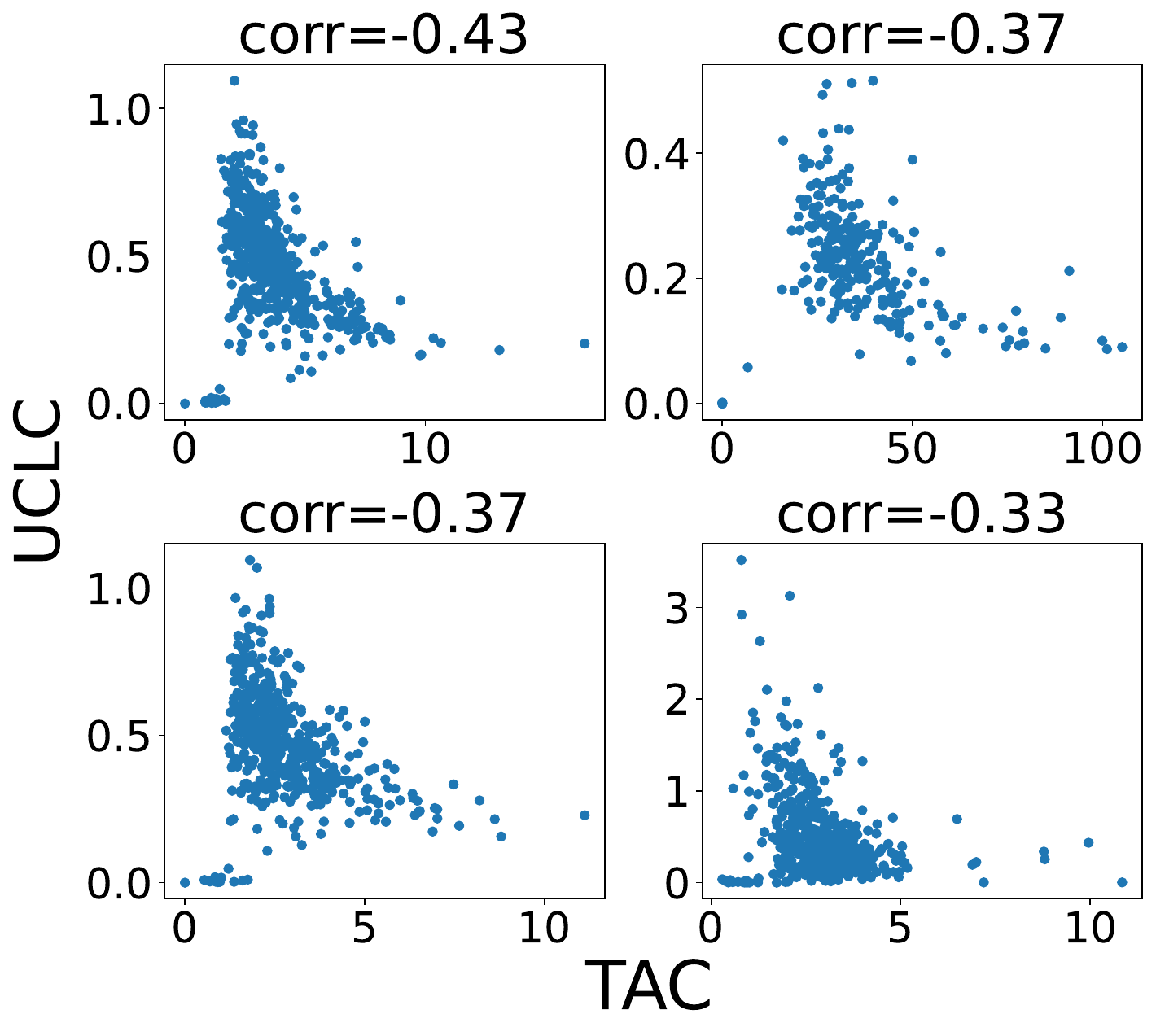}}
    \caption{The scatter plot of the upper bound of activation changes (UCLC) versus actual triggered-activation changes (TAC) for all channels in the last four convolution layers. $corr$ presents the Pearson Correlation Coefficient. (a) depicts an end-to-end SL-trained ResNet18 classifier with BadNets dataset poisoning. (b) illustrates a ResNet18 encoder injected by DRUPE through encoder poisoning.}
    \label{fig:clp_correlation}
    \vspace{-6mm}
\end{figure}

To understand the gap in CLP's performance between the end-to-end SL and the FT-head setting, we visualize the correlation between UCLC and TAC in \cref{fig:clp_correlation}. 
Under SL, it does show a positive correlation ($corr$) between UCLC and TAC.
However, under the transfer learning context with novel backdoor injection, there is no absolute correlation between UCLC and TAC, indicating a lack of strong evidence linking real backdoor-related channels (high TAC) with the presumptive ones (high UCLC). 
In other words, the backdoor trigger in SL inductively chooses the channels that can help induce TAC more easily.
Instead, the backdoor trigger under TL is more covert as it depends on 
channels that are, in a way, more dispersed.
\begin{mtbox}{Remark}
The change in how the backdoor is injected also changes where the backdoor activation relies on.
Blindly making assumptions on what kind of neurons are more likely to be responsible for backdoor, only based on the model itself, is also unreliable, as there is no guarantee of the distribution of where the backdoor activates.
\end{mtbox}

\subsection{Towards A Proactive Mindset}
\label{sec:constructive_mindset}
To mitigate backdoor risks and achieve a clean classifier with high ACC and low ASR, poison detection, poison suppression, and poison removal are the typical solutions for a defender.
At a high level, each of them requires identifying what constitutes poisoned features or characteristics—whether on a sample-wise, representation-wise, or neuron-wise basis—followed by eliminating these poison elements.
In a sense, these solutions can be seen as \textit{reactive}, as they primarily focus on defending against somewhat \textit{known} threats with a specific threat model (targeting what has or may have been compromised), then taking action to locate the poison elements and minimize their malicious influences by directly removing them or suppressing them. 
In general, these reactive strategies rely heavily on certain assumptions about the characteristics of the poison elements.
However, these assumptions may not hold across \textit{unknown} threats, novel types of attacks, or different training paradigms. 
As a result, their effectiveness is significantly compromised, as they fail to accurately identify poison elements.

In this study, rather than responding to specific backdoor threats by searching for poisoned elements, we advocate for a \textit{proactive} mindset that focuses on identifying and amplifying clean elements to defend against \textit{unknown} backdoor threats.
On the one hand, poison elements can arise from multiple sources (such as the dataset and encoder) and manifest in various forms (including different types of backdoor triggers), making their accurate characterization challenging. In contrast, clean elements are generally more stable and uniform, making it easier to locate at least a small number of them.
On the other hand, in a complex environment where many available resources are untrusted, it makes sense for defenders to utilize their limited trusted resources to bootstrap and obtain more trustworthy elements.


\section{Trusted Core Bootstrapping}\label{sec:T-Core}
In this section, we present the pipeline of our Trusted Core Bootstrapping framework, which consists of three stages that align with the proactive mindset we promote in \cref{sec:constructive_mindset} to develop an accurate and secure classifier within the defense context.
We start with sifting a limited high-credible seed data from the poisoned dataset (\cref{sec:seedsifting}), followed by a data expansion that expands the seed data into a small clean subset (\cref{sec:data_expansion}).
Then an encoder neuron filtering is employed to filter the trusted model neurons for the clean downstream task (\cref{sec:encoder_filtering}).
Finally, we bootstrap the training process using the filtered encoder and clean data pool, progressively improving the model by gradually incorporating more trusted data (\cref{sec:bootstrapping}).

\subsection{Sifting A Clean Set} \label{sec:datasifting}
In this section, we introduce how to extract a trusted subset, referred to as  $\subdata$, from the untrusted downstream dataset $\downdata$, that can support the successful training of a clean model for the downstream task.
Intuitively, directly locating a clean subset will be non-trivial. 
Thus, we propose to first screen a limited number of representative data points from each class by setting a very high standard, then using these samples as seed data to bootstrap a clean subset.

\subsubsection{Selecting Seed Data} \label{sec:seedsifting}

As illustrated in \cref{sec:poison_detection}, poison samples blend with benign samples, making it challenging to sift out clean samples through simple representation analysis in feature space. 
To address this, we sidestep the full dependence on latent separability in a single feature space and instead resort to examining the \textbf{\textit{topological invariance }}of each sample across different DNN layers. 
The rationale is that the most representative samples should maintain consistency in their position within the data distribution as they propagate through the network.
Specifically, we propose two rules to identify high-credible samples:
\begin{itemize}[leftmargin=10pt]
\item \textbf{Majority Rule}: \uline{A high-credible sample should belong to the majority group of samples in a DNN layer.} This rule assumes that the number of poisoned samples in a given class is smaller than that of clean samples, a condition generally met in backdoor poisoning.
This rule imposes a global topological invariance, as it requires selection to avoid small groups in the whole data distribution.
By enforcing this rule, we expect selected samples to be embedded within the core of the distribution across layers.

\item \textbf{Consistency Rule}: \uline{A high-credible sample should have consistent nearest neighbors from its class across different DNN layers.}
This rule imposes local topological invariance as it requires selection to favor a consistent local neighborhood.
By enforcing this rule, we expect that a selected sample is truly representative of its class, not because of the clustering effect of DNN.
\end{itemize}
We present the detail of the \textbf{topological invariance sifting} (TIS), which are based on these two rules, in \cref{algo:sifting}.

Concretely, given the untrusted encoder $g(\cdot;\prepara)$ and the classification head $f(\cdot; {\phi_\mathrm{down})}$ fine-tuned over $\downdata$, for each class $k$, we record the activations of its all samples $\downdata_k$ of the $L$ layers before the final layer of the entire network $h=f({\phi_\mathrm{down}}) \circ g(\cdot;\prepara)$ (\cref{line:record_activation}).
Then, we proceed with the data sifting in each $\downdata_k$.

\begin{algorithm}[!t] 
\footnotesize
    \caption{Topological Invariance Sifting}
    \label{algo:sifting}
    \KwIn{
    $\prepara$ (Untrusted encoder parameters);
    \\\quad \quad \quad
    $\downdata$ (Untrusted downstream dataset);
    }
    \KwOut{ $\seeddata$ (A small set of seed data);}
    \KwParam{
    $L$ (Number of considered layers);
    \\  \quad \quad \quad  \quad \quad \quad 
    $m$ (Number of nearest neighbors);
    \\  \quad \quad \quad  \quad \quad \quad 
    $\alpha$ (Seed data proportion selected);
    }
    \BlankLine
    \smalltcc{Record samples activations of $L$ layers}
    $\mathcal{A} \leftarrow \textsc{RecordActivations}(h, L, [\downdata_1, \dots, \downdata_K])$\; \label{line:record_activation}
    \For{$k \leftarrow 1$ \KwTo $K$}{
        \smalltcc{Cluster activations in each layer}
    
        \For{$l \leftarrow 1$ \KwTo $L$}{
            $C^l \leftarrow \textsc{Clustering}(\mathcal{A}[l,k])$\;\label{line:clustering}
           $C^l_{max} \leftarrow \textsc{max} (C^l)$\;\label{line:maxcluster}
           
        }
         $\mathcal{B}_k \leftarrow C^1_{max} \cap C^2_{max} \dots \cap C^L_{max} $\;\label{line:cluster_intersection}

        \smalltcc{Obtain consistent neighbors of $x$}
        \For{$x \in \mathcal{B}_k$}{
            $\mathtt{Numbers} \leftarrow \emptyset$\;
            \For{$l \leftarrow 1$ \KwTo $L$}{\label{line:L-layers_neighbors}

                ${N}^l_x \leftarrow \textsc{NearestNeighbors}(x, \mathcal{A}[l], m)$\;
                \label{line:nearestneighbors}
            }

              $\mathtt{Numbers}.\textsc{append}(|{N}^1_x \cap {N}^2_x \dots \cap  {N}^L_x \cap \downdata_k |)$\;\label{line:neighbors_number}
        }
    \smalltcc{Sort by consistent neighbors number}
    $\mathcal{B}_k^* \leftarrow \textsc{SortDescending}(\mathcal{B}_k, \textsc{key}= \mathtt{Numbers})$\;\label{line:num_ranking}
    $\seeddata_k \leftarrow \mathcal{B}_k^*[0:  \alpha \times  |
    \downdata_k|]$\;\label{line:num_picking}
    }
    $\seeddata \leftarrow \seeddata_1 \cup \seeddata_2 \dots  \cup \seeddata_K$\;
    \Return{$\seeddata$}

\end{algorithm}

We first apply the majority rule using a clustering procedure on the output activations of each layer.
We notice a poison detection \cite{chen2018detecting} uses clustering to identify poisoned samples, relying on the last hidden layer's representation with K-means for two-class clusters.
However, in our approach, the majority rule is applied before the consistency rule, aiming to remove samples from any smaller clusters that may contain poisoned samples across all layers. 
Thus, we utilize density-based clustering to partition \(\downdata_k\) and obtain the intersection \(\mathcal{B}_k\) of the largest clusters in each layer (\cref{line:clustering,line:maxcluster,line:cluster_intersection}). Unlike fixed-number clustering methods, density-based clustering dynamically determines the number of clusters, effectively detecting irregularly shaped clusters and managing both dense and sparse regions.

After applying the majority rule, we use the consistency rule to refine the output \(\mathcal{B}_k\). For each sample \(x\) in \(\mathcal{B}_k\), we identify its nearest \(m\) neighbors \(N^l_x\) in each layer \(l\) (\cref{line:L-layers_neighbors,line:nearestneighbors}). We then filter these neighbors by class and assess how many are consistent across all layers. The number of consistent neighbors from the same class across \(L\) layers can be denoted as \(|{N}^1_x \cap {N}^2_x \cap \dots \cap {N}^L_x \cap \downdata_k|\) (\cref{line:neighbors_number}). Finally, we select samples with the most consistent neighbors from each class, maintaining a proportion of \(\alpha\) (\cref{line:num_ranking,line:num_picking}) to ensure an equal number of seed data per class.


\subsubsection{Boostrapping the Clean Set}\label{sec:data_expansion}

\begin{figure*}[t]
    \centering
    \includegraphics[width=0.8\textwidth]{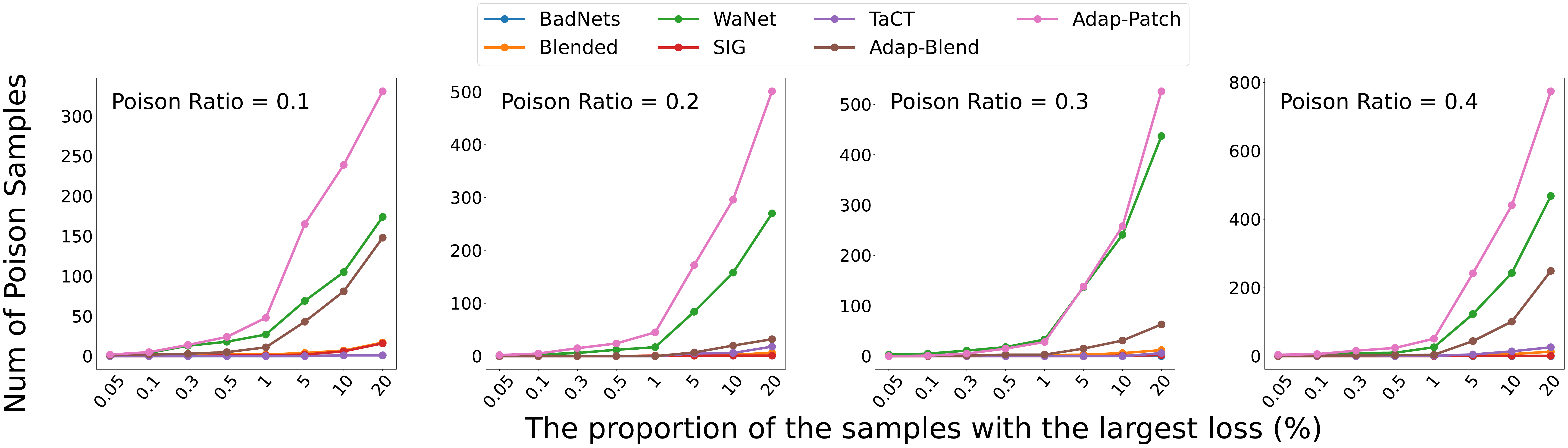} 
    \caption{Number of poison samples in the largest loss region of the target class after seed expansion.
    Experiments are conducted on 7 dataset poisoning attacks at poison ratios of 0.1, 0.2, 0.3, and 0.4 for the target class.}
    \label{fig:ct_loss_distribution}
\end{figure*}

Once we obtain a set of high-credible samples from each class, we can use them as seed data to bootstrap a clean subset. In \cref{sec:poison_detection}, we introduced a state-of-the-art poison detection method called Confusion Training (CT) \cite{qi_sec_2023}, which uses a hold-out clean dataset to offset clean features, resulting in lower loss values for poisoned samples.
However, this approach is ineffective in identifying poisoned samples in the defense context, as the lowest loss area is mixed with both poisoned and clean samples, due to the optimization limited to the classification head, as shown in \cref{fig:ct_low_loss}. 
Nevertheless, the largest loss area is exclusively filled with clean samples regardless of the dataset poisoning type (see \cref{fig:ct_loss_distribution}). Therefore, we tailor CT as an expansion tool to bootstrap a clean subset using the seed data.

We perform two concurrent minimizations: one over poisoned dataset and another over mislabeled clean samples. The joint loss is defined as:
\begin{align}
    \lambda \cdot \loss(h(\mathbf{X}_i), \mathbf{Y}_i) + (1-\lambda) \cdot \loss(h({\mathbf X}'_i), \mathbf{Y}^*),
    \label{eq:confusion_training}
\end{align}
where \((\mathbf{X}_i, \mathbf{Y}_i)\) is a mini-batch from the poisoned dataset, and \(({\mathbf X}'_i, \mathbf{Y}^*)\) is created by randomly mislabeling a clean batch from a clean base set.
The overall procedure is outlined in \cref{algo:seedexpansion}. We start with a small set of clean seed data \(\seeddata\) and the entire untrusted dataset \(\downdata\). We randomly initialize the classification head parameters \(\phi\) and fine-tune it on \(\downdata \setminus \subdata\), then apply confusion training (CT) using both \(\downdata \setminus \subdata\) and seed data \(\subdata\).
After CT, we add the \(r_\mathrm{expand}\) samples with the highest loss from \(\downdata \setminus \subdata\) to \(\downdata\). We then sort the remaining samples by loss in descending order and add the top \(|\downdata \setminus \subdata| \times r_\mathrm{expand}\) samples to the clean subset \(\subdata\). This process is repeated until we achieve a sufficient subset, stopping when \(\subdata\) reaches 10\% of \(\downdata\).

\begin{algorithm}[t]
\caption{\mbox{Seed Expansion}\label{algo:seedexpansion}}
\footnotesize
    \KwIn{$\seeddata$ (A small set of clean seed data);
    \\\quad \quad \quad
    $\downdata$ (Downstream dataset);
    }
    \KwOut{
    $\subdata$(A clean subset)\;
    }
    \KwParam{
        $r_\textrm{expand}$ (Select ratio for expansion)\;
    }
    
    \BlankLine
    $\subdata \gets \seeddata$\;
      \Repeat{$i \leftarrow 1 $ \KwTo $I$ }{
         Randomly initialize $\phi$\;
        
        Fine-tune $\phi$ with  $\loss\left( f(\cdot;\phi) \circ g(\prepara), \cdot\right)$ over $\downdata$\;

        \smalltcc{Confusion training with $\subdata$}        
        $\phi \gets \textbf{CT}  \Big( \phi, \downdata \setminus \subdata,  \subdata \Big)$ with \cref{eq:confusion_training}\;

         $\mathtt{Loss} \gets \left\{ \loss\left(h\left(x\right), y\right) | (x, y) \in \downdata \setminus \subdata \right\}$\;
            \smalltcc{Sort by loss in descending order}
    $\mathcal{R} \gets \textsc{SortDescending}(\downdata \setminus \subdata,\textsc{key}=\mathtt{Loss})$;\\
            \smalltcc{Add samples with the largest loss}
            $\subdata \gets \subdata \cup \mathcal{R}\left[1: |\downdata \setminus \subdata| \times r_\mathrm{expand}\right]$\;\label{line:sample_select}
    }

\Return{$\subdata$}
\end{algorithm}

\subsection{Filtering the Encoder Channel} 
\label{sec:encoder_filtering}
After obtaining a trusted subset of downstream data \(\subdata\), the next stage is to locate clean neurons within the encoder.
To facilitate defense operations on user devices with limited computational resources, which can only support the gradient computation of a limited number of model parameters, we introduce a selectively imbalanced unlearn-recover process to identify clean neurons in the pre-trained encoder.
Typically, an encoder DNN is constructed in a modular fashion, with each module containing a transformation layer for feature extraction and a normalization layer for scaling and shifting features. For instance, in a convolutional network, different convolution channels extract both backdoor-related and normal features, with normalization parameters adjusting them for the final classification representation.
As shown in \cref{sec:poison_removal}, assumptions about which neurons may be responsible for the backdoor are unreliable in data-free analysis using CLP. Therefore, we propose utilizing both the untrusted downstream dataset \(\downdata\) and the trusted \(\subdata\) to directly locate clean channels in the encoder.
The procedure of our encoder channel filtering is described in \cref{algo:pruning}.

\noindent\textbf{Selective Unlearning.} We start by an unlearning process on the downstream data \(\downdata\) as follows:
\begin{align}
    \max _{\theta_{\text{norm}}} \mathbb{E}_{(x, y) \in \downdata} \left[ \ell\left( f(\downpara) \circ g(x; \prepara), y \right) \right]
    \label{eq:selective_unlearning}
\end{align}
Here, \(f(\downpara)\) represents the previously fully trained classification head on \(\downdata\), with the encoder \(g(\cdot; \prepara)\) fixed. \(\theta_{\text{norm}}\) denotes the parameters within the normalization layers.
This approach allows the model to intentionally ``lose'' some performance over the entire downstream data while keeping other layers and classification heads frozen during the unlearning phase.
Additionally, this process is memory-efficient, selectively adjusting only the normalization parameters, which typically constitute less than 1\% of the encoder parameters. 
Consequently, the encoder's ability to extract clean or backdoor features, along with the functionality of the classification head, remains largely intact.

\noindent\textbf{Filter Recovering.} The recovering process aims to restore the model's ability to predict clean samples, which the previous unlearning process has impaired. This recovery is crucial for discerning clean channels. 
By enforcing recovery of the downstream clean task, the network enhances the association between clean samples and clean channels over backdoor-related ones.
To achieve this, we introduce a filter mask \(\mathbf{m}^{\kappa}\) to indicate which operators to select for enhancing associations. Formally, the recovery works by solving the following minimization:
\begin{align}
    \min _{\mathbf{m}^{\kappa}} \mathbb{E}_{(x, y) \in \subdata} \left[ \ell\left( f(\downpara) \circ g(x; \mathbf{m}^{\kappa} \odot \hat{\theta}_{\textrm{pre}}), y \right) \right],
    \label{eq:filter_recovery}
\end{align}
where \(\hat{\theta}_{\textrm{pre}}\) denotes the modified encoder parameters updated by the unlearning process, which only affects normalization layers, and \(\mathbf{m}^{\kappa}\) denotes the masks applied to the channels in transformation layers. 
In this minimization, \(\mathbf{m}^{\kappa}\), which constitutes less than 1\% of the encoder parameters, acts as a soft filter to identify the channels most beneficial for recovering clean task performance, reinforcing correct associations between inputs and outputs. We initialize \(\mathbf{m}^{\kappa}\) as all ones and apply a clipping operation during updates (\cref{line:clip}).


\begin{algorithm}[t]
    \footnotesize 
    \caption{Encoder Channel Filtering}
    \label{algo:pruning}    
    \KwIn{
    $\prepara$ (Untrusted encoder parameters);
    \\ \quad \quad \quad
    $\subdata$ (A clean subset);
    \\ \quad \quad \quad
    $\theta_{\textrm{norm}}$ (Unlearning parameters)
    }
    \KwOut{$\mathbf{m}^{\kappa}$ (A suspected filter binary mask);}
    \KwParam{
    $ACC_\textrm{min}$ (Clean accuracy threshold);
    \\  \quad \quad \quad  \quad \quad \quad 
    $\sigma$ (Filter threshold);
    }
    \BlankLine
    \smalltcc{Selective Unlearning}
    \Repeat{training accuracy of classifier F reaches $ACC_\textrm{min}$  }{
    Optimize $\theta_\textrm{norm}$ with \cref{eq:selective_unlearning} using downstream data $\downdata$\;  \label{line:max_loss} 
    
       
    }
    \smalltcc{Filter Recovering}
    $\mathbf{m}^{\kappa} = [1]^n$\;
    \Repeat{training converged}{
        Optimize $\mathbf{m}^{\kappa}$ with \cref{eq:filter_recovery} using trusted subset $\subdata$\;
        $\mathbf{m}^{\kappa} = clip_{[0,1]}(\mathbf{m}^{\kappa})$\;\label{line:clip}
    }
    \smalltcc{Channel Filtering}
    Untrusted channels parameters $ \psi \gets \theta[\mathbb{I}\left(\mathbf{m}^{\kappa} \le \sigma \right)]$\;  \label{line:untrusted_operators}
    Trusted channels parameters $ \chi \gets \theta[\mathbb{I}\left(\mathbf{m}^{\kappa} > \sigma \right)]$\; \label{line:Trusted_operators}
    \Return{$\psi$ and $\chi$}
\vspace{-1mm}
\end{algorithm}

\noindent\textbf{Channel Filtering.} Once recovery is complete, the mask values indicate each channel's contribution to the clean task. A high value suggests that the channel (and its corresponding neurons) has a greater correlation with the downstream clean task, identifying these channels as trusted.
In practice, filtering is applied to the original encoder with parameters \(\prepara\), based on the mask learned from \(\hat{\theta}_{\textrm{pre}}\). The threshold \(\sigma\) is determined by the percentage of channels to preserve in each layer. After thresholding, we obtain both the untrusted channels \(\psi\) and trusted channels \(\chi\) from the encoder.




\subsection{Bootstrapping Learning}\label{sec:bootstrapping}
After completing the previous stages, we have established two trustworthy elements: the clean subset \(\subdata\) and the trusted channels \(\chi\) of the pre-trained encoder. We now explain how to leverage these trusted elements to bootstrap the learning of a clean model.
The bootstrapping process involves gradually training the model with the clean pool \(\cleandata\) and expanding \(\cleandata\) until it reaches a sufficient proportion of the entire dataset. 
{\color{black}The bootstrapping procedure is detailed in the overall T-Core framework in \cref{algo:bootstrapping_learning}.}

\noindent\textbf{Optimization of Untrusted Channels.} 
The untrusted channels demonstrate a lower correlation with the downstream clean task. 
As a result, they misalign with the input-label mapping of the clean subset and harbor backdoor-related channels.
Thus, we reinitialize the parameters of these untrusted channels and optimize them along with the classification head through gradient computation\footnote{In PyTorch, we define a list of parameters as leaf tensors, each tensor having the same shape as the untrusted channels of a respective layer, and assign the values of these leaf tensors to the untrusted channels accordingly.} as follows:
\begin{equation}
\begin{gathered}
 \min_{\phi, \psi}   \mathbb{E}_{(x, y) \in \cleandata } \left[ \loss\left(f(\phi)\circ  g({x} ; {\psi \cup\chi}), y\right)  \right].
\end{gathered}
\label{eq:meta_phi}
\end{equation}
This not only eliminates the backdoor from the encoder but turns them into the channels of the clean downstream task.
Empirically, we identify 90\% of the channels in transformation layers as clean in Encoder Channel Filtering, thus we merely optimize less than 10\% of the encoder's parameters in subsequent training.

\begin{algorithm}[t]
\footnotesize
    \caption{ Trusted Core Bootstrapping}
    \label{algo:bootstrapping_learning}
    \KwIn{
    $g(\cdot; \prepara)$ (Untrusted encoder);
    \\\quad \quad \quad
    $\downdata$ (Untrusted downstream dataset);
    }
    \KwOut{
    Trusted encoder 
    }
    \KwParam{
    $WarmupEpoch$ (Epoch number)
    \\  \quad \quad \quad  \quad \quad \quad 
    $\rho$ (filtered out data size);
    }
    \BlankLine

Randomly initialize $\phi$\;
Train $\phi$ with  $\loss\Big( f_\phi \circ g(\cdot| \prepara), \cdot \Big)$ over $\cleandata$\;
$\seeddata \gets \textbf{Topological Invariance Sifting}\left( \prepara, D\right)$\;
$\subdata \gets \textbf{Seed Expansion} \left( \seeddata, \downdata \right)$\;
$\psi, \chi \gets \textbf{Encoder Channel Filtering}\left(\prepara, \downdata\right)$\;
Randomly initialize classifier head $\phi$ and untrusted neurons $\psi$\; 
\label{line:init_para}
$\cleandata \gets \subdata$\; \label{line:init_data}

    \For{$ i= 1$ to $Iter1$}{
    Train $\phi$ and $\psi$ using $\cleandata$ for $T$ epochs \;
        Select $\gamma_1 \% $ samples with  the smallest loss from \textit{each class} of $\downdata \setminus \cleandata$ and add into $\cleandata$\; \label{line:add1}
    }

 \For{$ i= 1$ to $Iter2$}{
    Train $\phi$ and $\psi$ using $\cleandata$ for $T$ epochs \;
        Select $\gamma_2 \% $ samples with  the smallest loss from \textit{entire dataset} of $\downdata \setminus \cleandata$ and add into $\cleandata$\;  \label{line:add2}
    }

    \Repeat{ $|\cleandata| / |\downdata| \ge \rho $}{
        Train $\phi$ and $\psi$ with $\cleandata$ \;
        $\phi' \leftarrow \phi, \psi' \gets \psi $\;
         Train $\phi', \psi'$ using $\downdata\setminus\cleandata$ for one epoch \;
         $\mathtt{Loss}_1 \gets \{\loss(f(\phi) \circ g(x; \phi \cup \chi), y) \mid (x, y) \in \downdata \setminus \cleandata\}$\;
         $\mathtt{Loss}_2 \gets \{\loss(f(\phi') \circ g(x; \phi' \cup \chi), y) \mid (x, y) \in \downdata \setminus \cleandata\}$\;
         Select $\gamma_3\%$ samples with the lowest values in $\mathtt{Loss}_1 - \mathtt{Loss}_2$ and add them into $\cleandata$\; \label{line:add3}
    }
    

    \Return{$\phi, \psi, \chi$}

\label{algo:entire}
\end{algorithm}

\noindent\textbf{Clean Pool Expansion with Loss Guidance.} This stage involves two processes to expand \(\cleandata\) and train the model. We first initialize the clean pool \(\cleandata\) with \(\subdata\) (\cref{line:init_data}).
In the first process, we gradually expand \(\cleandata\) based on the model's prediction loss for each class in \(\downdata \setminus \cleandata\). After training the model with \(\cleandata\) for \(T\) epochs, we add samples with the lowest \(\gamma_1\) loss \textit{in each class} to maintain class balance (\cref{line:add1}). This process is repeated for \(Iter1\) iterations.
After this, the model achieves a certain accuracy on clean samples. In the second process, we incorporate samples with the lowest \(\gamma_2\) loss from the entire set \(\downdata \setminus \cleandata\) (\cref{line:add2}) for \(Iter2\) iterations. This strategy helps avoid selecting poisoned samples from the target class, enabling the selection of more clean samples from non-target classes.

\noindent \textbf{Clean Pool Expansion with Meta Guidance.} After expanding with loss guidance, continually adding samples with the smallest prediction loss may include poisoned samples.
To differentiate clean complex samples from poisoned ones, we adopt a meta-learning approach \cite{gao2023backdoor,finn2017model} to augment \(\cleandata\). We first train a temporary model on \(\downdata \setminus \cleandata\), then select samples with the smallest loss reduction between the original and temporary models to add to \(\cleandata\) (\cref{line:add3}). 
The rationale is that clean hard examples are more challenging to learn compared to easily inserted backdoor-poisoned examples. Even if some hard-to-learn backdoor samples are mistakenly selected, they require significantly more data and training to become effective. We halt this process once the ratio \(|\cleandata| / |\downdata|\) reaches 0.9 (90\% of the dataset).




\section{Experiments} \label{sec:experiments}

In this section, we first evaluate the effectiveness of our Trusted Core Bootstrapping framework by analyzing both its individual modules and the end-to-end framework as a whole.
Since no prior work has considered the complex defense context we explore, and existing methodologies cannot serve as standalone solutions—even when adjusted as shown in \cref{sec:methodology_analysis}—we compare each of our modules with existing approaches that aim to achieve the same goals to demonstrate the superiority and irreplaceability of ours.
Finally, we assess the end-to-end defense performance of our T-Core framework.
In summary, our main evaluation aims to answer the following research questions:

\begin{itemize}[leftmargin=5pt]
    \item{\textbf{RQ1}: How effective is our Clean Data Sifting in obtaining a  clean subset from various dataset poisoning types across downstream datasets under threats \threattwo~and \threatthree?}
    \item{\textbf{RQ2}: How effective is our Encoder Channel Filtering in producing a purified encoder for transfer learning over a clean downstream dataset against \threatone?}
    \item{\textbf{RQ3}: How effective is our Bootstrapping Learning in developing the clean classification head from a clean subset and clean encoder under dataset poisoning of \threattwo?}
    \item{\textbf{RQ4}: How effective is our end-to-end Trusted Core Bootstrapping framework in defending against any \textit{unknown} backdoor threats \threatone, \threattwo, \threatthree, or the scenario where both \threatone~and \threattwo~exist?}
\end{itemize}

{\color{black}
In addition, we then evaluate the scalability of our T-Core bootstrapping framework across different dimensions to provide a holistic understanding of its practical deployment potential. Specifically, we extend our analysis beyond static threat scenarios and basic performance metrics to explore T-Core’s resilience against adaptive attacks, its sensitivity to configuration choices, adaptability to emerging transformer architecture, and efficiency in resource-constrained settings. 
These aspects evaluate whether a theoretically sound defense of T-Core is a viable solution for real-world applications.
In general, our scalability evaluation aims to answer the following research question:
\begin{itemize}[leftmargin=5pt] \item \textbf{RQ5}: How scalable is T-Core in terms of defense effectiveness against adaptive adversaries, sensitivity to hyperparameter variations, adaptability to pre-trained Vision Transformer (ViT) models, and computational efficiency compared to other defense mechanisms? \end{itemize}
}

\begin{table*}[ht]
    \centering
    \caption{
Number of poisoned samples in the sifted-out samples (deemed clean) from poisoned datasets with varying poison ratios (0.1, 0.15, 0.2, 0.23, 0.3) in the target class, covering 5 datasets and 7 dataset poisoning attacks in \threattwo~and 4 adaptive poisoning attacks in \threatthree. NPD denotes the \uline{\textit{number of poisoned data} in the target class}, and NFD denotes the \uline{\textit{number of filtered data} in the target class deemed clean}. Cells with \uline{zero NPD in filtered data} are marked \colorbox{D9EAD3}{green}, while cells with \uline{NPD exceeding NFD $\times$ Poison Ratio $\times$ 0.5} are marked \colorbox{FFCCCC}{red}.
    }

    \adjustbox{center}{
    \resizebox{1\textwidth}{!}{\large
    \setlength{\tabcolsep}{0.1pt}
\begin{tabular}{cc|ccccccccccccccccccccccccccccccccccc|cccccccccc}
        \toprule
        \multicolumn{2}{c|}{\textbf{Threat Type}} & \multicolumn{35}{c|}{\threattwo} & \multicolumn{10}{c}{\threatthree} \\ \midrule
        \multicolumn{1}{c|}{} &  & \multicolumn{5}{c|}{\textbf{BadNets}} & \multicolumn{5}{c|}{\textbf{Blended}} & \multicolumn{5}{c|}{\textbf{SIG}} & \multicolumn{5}{c|}{\textbf{WaNet}} & \multicolumn{5}{c|}{\textbf{TaCT}} & \multicolumn{5}{c|}{\textbf{Adap-Blend}} & \multicolumn{5}{c|}{\textbf{Adap-Patch}} & \multicolumn{5}{c|}{\textbf{BadEncoder}} & \multicolumn{5}{c}{\textbf{DRUPE}} \\ \cline{3-47} 
        \multicolumn{1}{c|}{\multirow{-2}{*}{\textbf{Dataset}}} & \multirow{-2}{*}{\textbf{Methods}} & 0.1 & 0.15 & 0.2 & 0.25 & \multicolumn{1}{c|}{0.3} & 0.1 & 0.15 & 0.2 & 0.25 & \multicolumn{1}{c|}{0.3} & 0.1 & 0.15 & 0.2 & 0.25 & \multicolumn{1}{c|}{0.3} & 0.1 & 0.15 & 0.2 & 0.25 & \multicolumn{1}{c|}{0.3} & 0.1 & 0.15 & 0.2 & 0.25 & \multicolumn{1}{c|}{0.3} & 0.1 & 0.15 & 0.2 & 0.25 & \multicolumn{1}{c|}{0.3} & 0.1 & 0.15 & 0.2 & 0.25 & 0.3 & 0.1 & 0.15 & 0.2 & 0.25 & \multicolumn{1}{c|}{0.3} & 0.1 & 0.15 & 0.2 & 0.25 & 0.3 \\
\bottomrule
        \end{tabular}
    }
    }
\label{tab:clean_data_sifting}
\end{table*}

\begin{table*}[t]
    \centering
    \caption{Number of poisoned samples obtained by seed expansion when expanding till different expansion ratios under \threattwo.
    In each cell, the first number is the number of poisoned samples in the seed data, and the second number is the number of poisoned samples in the added data after seed expansion.
    NTD: \textit{number} of \textit{total} \textit{data}  after seed expansion.
    }
    \adjustbox{center}{
    \resizebox{1\textwidth}{!}{\large
    \setlength{\tabcolsep}{1pt}
    %
} \\ \midrule
    \textbf{STL-10} & 0+0 & 0+0 & 0+0 & 0+0 & 0+0 & 0+0 & \multicolumn{1}{c|}{0+0} & 500 & 0+0 & 0+3 & 0+0 & 0+\ 0 & 0+0 & 0+2 & \multicolumn{1}{c|}{0+2} & 1000 & 0+\ 0 & 0+\ 6 & 0+22 & 0+\ 22 & 0+20 & 0+18 & \multicolumn{1}{c|}{0+\ 16} & 1500 & 0+82 & 0+58 & 0+75 & 0+\ 75 & 0+\ 68 & 0+\ 43 & \multicolumn{1}{c|}{0+\ 37} & 2500 \\ \hline
    \textbf{CIFAR-10} & 0+1 & 0+0 & 0+0 & 0+0 & 0+1 & 0+2 & \multicolumn{1}{c|}{0+0} & 5000 & 0+4 & 0+0 & 0+0 & 0+\ 0 & 0+2 & 0+5 & \multicolumn{1}{c|}{0+7} & 10000 & 0+13 & 0+22 & 0+\ 0 & 0+115 & 0+36 & 0+76 & \multicolumn{1}{c|}{0+\ 68} & 15000 & 0+32 & 0+94 & 0+\ 4 & 0+387 & 0+217 & 0+151 & \multicolumn{1}{c|}{0+128} & 25000 \\ \hline
    \textbf{GTSRB} & 0+0 & 0+0 & 0+0 & 0+1 & 0+0 & 0+0 & \multicolumn{1}{c|}{0+0} & 3920 & 0+0 & 0+0 & 0+0 & 0+\ 3 & 0+0 & 0+0 & \multicolumn{1}{c|}{0+1} & 7841 & 0+\ 3 & 0+\ 0 & 0+\ 0 & 0+\ 84 & 0+\ 0 & 0+\ 5 & \multicolumn{1}{c|}{0+\ 17} & 11762 & 0+15 & 0+\ 0 & 0+\ 7 & 0+145 & 0+\ \ 7 & 0+\ \ 5 & \multicolumn{1}{c|}{0+\ 67} & 19604 \\ \hline
    \textbf{SVHN} & 0+0 & 0+0 & 6+1 & 0+5 & 0+1 & 0+3 & \multicolumn{1}{c|}{0+2} & 7325 & 0+1 & 0+3 & 6+5 & 0+22 & 0+5 & 0+5 & \multicolumn{1}{c|}{0+7} & 14651 & 0+\ 6 & 0+22 & 6+23 & 0+\ 67 & 0+20 & 0+22 & \multicolumn{1}{c|}{0+141} & 21977 & 0+14 & 0+89 & 6+77 & 0+245 & 0+\ 32 & 0+\ 65 & \multicolumn{1}{c|}{0+300} & 36628 \\ \hline
    \textbf{ImageNet-10} & 0+0 & 0+0 & 0+0 & 0+2 & 0+0 & 0+0 & \multicolumn{1}{c|}{0+0} & 1006 & 0+0 & 0+0 & 0+0 & 0+\ 7 & 0+0 & 0+0 & \multicolumn{1}{c|}{0+0} & 2012 & 0+\ 0 & 0+\ 0 & 0+\ 7 & 0+\ 13 & 0+\ 0 & 0+\ 4 & \multicolumn{1}{c|}{0+\ 6\ } & 3018 & 0+95 & 0+91 & 0+24 & 0+119 & 0+\ \ 1 & 0+\ 14 & \multicolumn{1}{c|}{0+\ 40} & 5030 \\ \bottomrule
    \end{tabular}
    }
    }
    \label{tab:expand_samples}
    \vspace{-4mm}
\end{table*}

\subsection{Experimental Setups}
\noindent\textbf{Datasets.} We use five image datasets: CIFAR-10~\cite{krizhevsky2009learning}, GTSRB~\cite{stallkamp2012man}, SVHN~\cite{netzer2011reading}, STL-10~\cite{coates2011analysis}, and ImageNet~\cite{deng2009imagenet} to construct clean and poisoned pre-trained encoders, as well as clean or poisoned downstream datasets, to evaluate our defense effectiveness. Additional details, including image dimensions, dataset sizes, and evaluation applicability, are provided in Appendix-\ref{sec:dataset_explanation}.

\noindent\textbf{Models.} We adopt ResNet18 as the default backbone for the pre-trained encoders, following \cite{BadEncoder,DRUPE,SSL-backdoor,CTRL,zhang2022corruptencoder}, which allows us to leverage their clean and poisoned checkpoints for evaluation. 
For transfer learning, we use a simple MLP consisting of three linear layers for the classification head.

\noindent\textbf{Encoder Poisoning Attacks.} We consider five attacks that aim to inject backdoors into the pre-trained encoder: BadEncoder \cite{BadEncoder}, DRUPE \cite{DRUPE}, SSLBackdoor \cite{SSL-backdoor}, CTRL \cite{CTRL}, and CorruptEncoder \cite{zhang2022corruptencoder}.
Details about how these encoder poisoning attacks are conducted and how we process their outcome encoders are provided in Appendix-\ref{subsec:EncoderExplanation}.

\noindent\textbf{Dataset Poisoning Attack.} 
We consider seven data poisoning backdoor attacks on the downstream dataset to evaluate the generalizability of our Clean Bootstrapping approach. These include: 1) the vanilla \textit{dirty label} attacks: BadNets \cite{gu2019badnets} and Blended \cite{chen2017targeted},
2) the \textit{clean label} attack: SIG \cite{barni2019new},
3) the \textit{sample-specific} attack: WaNet \cite{nguyen2021wanet},
4) the \textit{latent space adaptive attacks}: TaCT \cite{tang2021demon}, Adap-Blend \cite{qi2023revisiting}, and Adap-Patch \cite{qi2023revisiting}.
In our primary evaluation, \uline{we default to the poison ratio as 20\% of the target class}. 
We further adjust the poison ratio to showcase the scalability of our seed data filtering approach in \cref{sec:data_sift_exp}. 
We strictly follow the guidelines from each attack's paper for implementation.
Detailed configurations are outlined in the Appendix-\ref{subsec:DatasetPoisoningExplanation}.

\subsection{RQ1: Effectiveness of Clean Data Sifting}\label{sec:data_sift_exp}
We evaluate the effectiveness of our Topological Invariance Sifting (TIS) across various poison ratios (0.1, 0.15, 0.2, 0.23, 0.3) of the target class in dataset poisoning attacks, addressing threats \threattwo~and \threatthree. 
We maintain a sifting ratio \(\alpha\) of 0.01, selecting 1\% of the most credible samples from each class as seed data.
For comparison, we utilize two poison detection methods, SPECTRE \cite{hayase21a} and Spectral \cite{tran2018spectral}, alongside two state-of-the-art test-time backdoor input detection techniques, IBD-PSC \cite{hou2024ibdpsc} and SCALE-UP \cite{guo2023scaleup}, which do not require an additional clean base set. We also evaluate META-SIFT \cite{zeng2023meta}, the only prior method for sifting clean data tested on backdoor poisoning under the SL setting, in our TL setting.
To compete with our TIS, we configure the thresholds of each method to meet 1\% selection, categorizing the remaining 99\% as potentially untrustworthy. Additionally, we apply our Seed Expansion at a default poison ratio of 20\% of the target class, using the corresponding seed data to continually expand and recording results when the expansion ratio \(|\subdata| / |\downdata|\) reaches 10\%, 20\%, 40\%, and 50\% to assess effectiveness.

We report the number of poisoned samples sifted out as clean seed data (False Positive cases) in \cref{tab:clean_data_sifting}, along with the \textit{number of poisoned data} (NPD) and the \textit{number of filtered data} (NFD) in the target class. 
Our TIS effectively identifies clean samples with minimal false positives for most backdoor attacks across all downstream datasets. 
In 212 out of 225 scenarios, covering various poison ratios, backdoor types, and datasets, TIS achieves a 100\% precision, meaning all sifted samples are clean.
For SVHN, false positive cases are slightly higher due to its noisy nature, where extraneous digits appear in images, resulting in fewer high-credibility samples. 

Our TIS framework consistently outperforms existing methods, demonstrating robust reliability across all tested scenarios. While IBD-PSC, SCALE-UP, Spectral, and SPECTRE struggle with sample-specific (WaNet) and adaptive attacks (Adap-Blend, Adap-Patch) due to poisoned sample dispersion in latent space, META-SIFT fails against straightforward backdoors (BadNets, Blended, SIG), particularly under \threatthree~in noisy datasets like SVHN.
In contrast, TIS maintains high precision, achieving 100\% clean sifting in 212 out of 225 cases—spanning varying poison ratios, attack types, and datasets. 
SVHN exhibits marginally higher false positives, yet its large class sizes still allow effective 1\% sifting.

For T-Core’s Seed Expansion, \cref{tab:expand_samples} shows that poisoned sample counts remain low even at high expansion rates. False positives stay below 1\% at 50\% expansion, while 10\% and 20\% rates exhibit negligible contamination—sufficient for most defense mechanisms requiring clean subsets.
Given this balance between security and utility, we adopt 20\% as the default expansion rate.

\subsection{RQ2: Effectiveness of Encoder Filtering}

Our encoder channel filtering is designed to create a backdoor-free encoder to facilitate the clean classifier of the downstream task. 
Employing knowledge distillation techniques, some prior methods \cite{han2024mutual,han2024effectiveness,bie2024mitigating} utilize a clean dataset to distill clean knowledge from the backdoor encoder to also produce a clean student encoder.
We evaluate our approach against SSL-Distillation \cite{han2024effectiveness}, the only publicly available distillation-based method, under \threatone, to compare efficacy in removing backdoors. SSL-Distillation uses a subset of the clean pre-training dataset, referred to as SSL-Distillation$_{\textrm{pre}}$, which we additionally extend to the downstream dataset, denoted as SSL-Distillation$_{\textrm{down}}$. 

For a fair comparison, we use the same 20\% of the downstream dataset for both our method and SSL-Distillation$_{\textrm{down}}$, while SSL-Distillation$_{\textrm{pre}}$ uses an equivalent size subset from the clean \(\predata\). We assess their impact on downstream tasks by training a classification head on each purified or distilled encoder. 
Notably, SSL-Distillation requires complete fine-tuning of the backdoor encoder and building the student encoder from scratch, whereas our Encoder Channel Filtering only necessitates training less than 1\% of the total parameters. During subsequent training, we apply two configurations: one trains only the classification head after pruning untrusted channels (Ours$_{\textrm{head}}$), and the other trains both the classification head and the initialized untrusted channels (Ours$_{\textrm{head+untrusted}}$).


\begin{table}[t]
    \centering
    \caption{Comparison with SSL-Distillation under \threatone. 
    Results are shown for two configurations of our approach (Ours$_{\textrm{head}}$ and Ours$_{\textrm{head+untrusted}}$), as well as for two configurations of SSL-Distillation(SSL-Distillation$_{\textrm{pre}}$ and SSL-Distillation$_{\textrm{down}}$). Note that for CTRL, SSLBackdoor, and CorruptEncoder, the downstream dataset is the same as or a subset of the clean pre-training dataset.
    }
    \adjustbox{center}{
    \resizebox{0.47\textwidth}{!}{\large
    \setlength{\tabcolsep}{0.5pt}
    \begin{tabular}{c|c|c|cc|cc|cc|cc|cc}
    \toprule
    \multirow{2}{*}{\textbf{\makecell{Encoder\\Poisoning}}} & \multirow{2}{*}{\textbf{\makecell{Pre-training \\ Dataset}}} & \multirow{2}{*}{\textbf{\makecell{Downstream \\ Dataset}}} & \multicolumn{2}{c|}{\textbf{No Defense}} & \multicolumn{2}{c|}{\textbf{\makecell{SSL-\\Distillation$_\textrm{pre}$}}} & \multicolumn{2}{c|}{\textbf{\makecell{SSL-\\Distillation$_\textrm{down}$}}} & \multicolumn{2}{c|}{\textbf{Ours$_{\textrm{head}}$}} & \multicolumn{2}{c}{\textbf{Ours$_{\textrm{head+untrusted}}$}} \\ \cline{4-13} 
     &  &  & \textbf{ACC$\uparrow$} & \textbf{ASR$\downarrow$}  & \textbf{ACC$\uparrow$} & \textbf{ASR$\downarrow$}  & \textbf{ACC$\uparrow$} & \textbf{ASR$\downarrow$}  & \textbf{ACC$\uparrow$} & \textbf{ASR$\downarrow$}  & \textbf{ACC$\uparrow$} & \textbf{ASR$\downarrow$} \\ \midrule
    \multirow{6}{*}{\textbf{BadEncoder}} & \multirow{3}{*}{CIFAR-10} & STL-10 & 76.58 & 98.51 & 64.41 & 6.21 & 54.49 & 3.50 & \underline{64.46} & \textbf{0.01} & \textbf{76.26} & \underline{0.90} \\
     &  & GTSRB & 80.77 & 99.63 & 68.09 & 1.51 & \underline{69.29} & 1.57 & 63.70 & \underline{1.12} & \textbf{90.07} & \textbf{0.43} \\
     &  & SVHN & 65.35 & 97.56 & 66.68 & 10.38 & \underline{71.75} & 32.32 & 64.64 & \underline{3.49} & \textbf{92.20} & \textbf{2.82} \\ \cline{2-13} 
     & \multirow{3}{*}{STL-10} & CIFAR-10 & 70.57 & 98.93 & 50.27 & 13.09 & 51.77 & 12.27 & \underline{61.09} & \textbf{1.36} & \textbf{66.59} & \underline{2.92} \\
     &  & GTSRB & 70.83 & 98.99 & 50.89 & 2.08 & 49.98 & \underline{1.45} & \underline{53.14} & \textbf{0.12} & \textbf{88.22} & 1.98 \\
     &  & SVHN & 64.89 & 98.98 & 53.16 & 10.02 & 53.75 & 9.61 & \underline{62.07} & \textbf{3.15} & \textbf{86.99} & \underline{3.26} \\ \midrule
    \multirow{6}{*}{\textbf{DRUPE}} & \multirow{3}{*}{CIFAR-10} & STL-10 & 71.85 & 97.72 & 60.91 & 5.36 & 53.59 & 3.54 & \underline{68.38} & \textbf{1.89} & \textbf{72.06} & \underline{2.04} \\
     &  & GTSRB & 76.39 & 98.10 & 60.85 & \underline{0.17} & 61.76 & 0.48 & \underline{63.49} & 2.26 & \textbf{90.81} & \textbf{0.02} \\
     &  & SVHN & 72.99 & 92.71 & 65.72 & 57.69 & 72.76 & 50.08 & \underline{74.09} & \underline{5.23} & \textbf{92.20} & \textbf{2.82} \\ \cline{2-13} 
     & \multirow{3}{*}{STL-10} & CIFAR-10 & 71.14 & 80.49 & 55.70 & 8.74 & 55.14 & 8.34 & \underline{61.15} & \underline{3.64} & \textbf{73.42} & \textbf{0.91} \\
     &  & GTSRB & 65.11 & 85.03 & 51.00 & 1.64 & \underline{53.70} & \textbf{0.74} & 52.63 & 4.03 & \textbf{85.86} & \underline{1.00} \\
     &  & SVHN & 58.43 & 96.28 & 46.22 & 23.19 & 50.44 & \underline{4.99} & \underline{60.10} & 5.11 & \textbf{91.20} & \textbf{3.99} \\ \midrule
    \multirow{3}{*}{\textbf{CTRL}} & STL-10 & STL-10 & 52.15 & 9.88 & - & - & \underline{50.47} & \underline{2.49} & 48.13 & 2.86 & \textbf{50.99} & \textbf{0.32} \\ \cline{2-13} 
     & CIFAR-10 & CIFAR-10 & 75.31 & 44.90 & - & - & \underline{62.35} & \underline{1.65} & 50.83 & 2.54 & \textbf{62.42} & \textbf{0.74} \\ \cline{2-13} 
     & GTSRB & GTSRB & 66.78 & 6.54 & - & - & \underline{55.75} & 0.70 & 55.06 & \underline{0.61} & \textbf{84.45} & \textbf{0.22} \\ \midrule
    \textbf{SSLBackdoor} & ImageNet & ImageNet-10 & 82.85 & 36.48 & - & - & \underline{70.47} & 3.12 & 65.88 & \underline{2.94} & \textbf{85.76} & \textbf{2.73} \\ \midrule
    \textbf{CorruptEncoder} & ImageNet & ImageNet-10 & 82.35 & 58.46 & - & - & \underline{69.55} & 4.67 & 63.59 & \underline{2.61} & \textbf{84.53} & \textbf{0.36} \\ \bottomrule
    \end{tabular}
        }
}
     \label{tab:ssl-distillation}
     \vspace{-2mm}
\end{table}

The results from \cref{tab:ssl-distillation} show that our method generally trumps SSL-Distillation across all downstream tasks and achieves a very low ASR. 
Notably, SSL-Distillation performs poorly on the noisy downstream dataset of SVHN, supporting findings in \cite{han2024effectiveness} that SVHN poses increased security risks due to its exploitable noisy features under backdoor attacks. 
In contrast, our encoder filtering method performs significantly better on SVHN. We believe this is because our approach progressively selects channels beneficial for the clean classification task that aims to distinguish features apart. 
On the other hand, SSL-Distillation indiscriminately distills feature embeddings, especially from the already noisy samples in SSL-Distillation$_\textrm{down}$, which allows the backdoor to traverse through distillation.
Besides, our method yields an increased ACC for many cases due to the additional optimization of the original untrusted channels.

\subsection{RQ3: Effectiveness of Bootstrapping Learning}
We compare our Bootstrapping Learning with backdoor-removal methods that utilize a clean subset under \threattwo. 
Specifically, we consider two state-of-the-art methods: the trigger synthesis-based I-BAU \cite{ibau} and the fine-tuning-based FT-SAM \cite{sam_backdoor}. We apply them to the post-attack classification head. 
We also limit the training of our Bootstrapping Learning to the classification head without the untrusted channels.
In this experiment, we vary the clean data ratio to assess its impact on subset size dependence.

\begin{table}[t]
    \centering
    \caption{Comparison of our method with FT-SAM and I-BAU under \threattwo, using poisoned GTSRB as downstream dataset and a CIFAR-10 encoder, with a clean subset at varying ratios of the original clean GTSRB for defense.}
    \tabcolsep=0.1cm
    \adjustbox{center}{
    \tabcolsep=0.04cm
    \renewcommand{\arraystretch}{1}
    \resizebox{0.45\textwidth}{!}{\large
    \begin{tabular}{cc|cc|cc|cc|cc|cc|cc|cc}
    \toprule
    \multicolumn{1}{c|}{\multirow{2}{*}{Clean Ratio}} & \multicolumn{1}{c|}{\multirow{2}{*}{Methods}} & \multicolumn{2}{c|}{BadNets} & \multicolumn{2}{c|}{Blended} & \multicolumn{2}{c|}{SIG} & \multicolumn{2}{c|}{WaNet} & \multicolumn{2}{c|}{TaCT} & \multicolumn{2}{c|}{Adap-Blend} & \multicolumn{2}{c}{Adap-Patch} \\ \cline{3-16} 
    \multicolumn{1}{c|}{} & \multicolumn{1}{c|}{} & \multicolumn{1}{c}{ACC$\uparrow$} & \multicolumn{1}{c|}{ASR$\downarrow$} & \multicolumn{1}{c}{ACC$\uparrow$} & \multicolumn{1}{c|}{ASR$\downarrow$} & \multicolumn{1}{c}{ACC$\uparrow$} & \multicolumn{1}{c|}{ASR$\downarrow$} & \multicolumn{1}{c}{ACC$\uparrow$} & \multicolumn{1}{c|}{ASR$\downarrow$} & \multicolumn{1}{c}{ACC$\uparrow$} & \multicolumn{1}{c|}{ASR$\downarrow$} & \multicolumn{1}{c}{ACC$\uparrow$} & \multicolumn{1}{c|}{ASR$\downarrow$} & \multicolumn{1}{c}{ACC$\uparrow$} & \multicolumn{1}{c}{ASR$\downarrow$} \\ \midrule   
    \multicolumn{2}{c|}{No Defense} & 81.79 & 95.02 & 81.30 & 90.39 & 81.90 & 74.37 & 80.74 & 8.81 & 81.95 & 89.20 & 80.85 & 69.73 & 78.54 & 28.20 \\ \midrule
    \multicolumn{1}{c|}{\multirow{3}{*}{0.1}} & FT-SAM & 21.88 & 20.61 & 22.24 & 4.50 & 22.00 & 51.68 & 19.03 & 0.69 & 22.82 & 15.20 & 22.02 & 2.33 & 21.54 & 1.13 \\ \cline{2-16} 
    \multicolumn{1}{c|}{} & I-BAU & 16.98 & 26.36 & 17.36 & 0.74 & 17.03 & 46.20 & 13.86 & 1.40 & 18.39 & 28.80 & 17.22 & 0.35 & 15.38 & 0.59 \\ \cline{2-16} 
    \multicolumn{1}{c|}{} & Ours & 78.65 & 4.13 & 78.47 & 4.15 & 77.91 & 3.70 & 79.73 & 3.59 & 78.99 & 7.07 & 78.24 & 3.11 & 78.08 & 3.12 \\ \midrule
    \multicolumn{1}{c|}{\multirow{3}{*}{0.3}} & FT-SAM & 23.82 & 19.41 & 24.70 & 10.87 & 23.91 & 39.52 & 22.27 & 0.97 & 24.96 & 2.13 & 23.35 & 3.40 & 23.89 & 0.92 \\ \cline{2-16} 
    \multicolumn{1}{c|}{} & I-BAU & 16.60 & 21.17 & 17.82 & 0.75 & 17.66 & 42.94 & 13.15 & 1.41 & 16.36 & 18.67 & 16.68 & 0.13 & 14.73 & 0.08 \\ \cline{2-16} 
    \multicolumn{1}{c|}{} & Ours & 80.78 & 2.07 & 80.24 & 1.04 & 79.82 & 0.06 & 80.55 & 1.40 & 80.31 & 2.00 & 79.78 & 1.93 & 79.70 & 1.55 \\ \midrule
    \multicolumn{1}{c|}{\multirow{3}{*}{0.5}} & FT-SAM & 25.69 & 19.31 & 27.58 & 10.70 & 26.56 & 31.04 & 26.19 & 0.94 & 27.71 & 3.73 & 26.07 & 3.51 & 26.07 & 0.97 \\ \cline{2-16} 
    \multicolumn{1}{c|}{} & I-BAU & 17.50 & 29.51 & 15.60 & 0.48 & 13.98 & 37.63 & 12.25 & 1.78 & 17.51 & 30.67 & 16.23 & 0.13 & 15.04 & 0.05 \\ \cline{2-16} 
    \multicolumn{1}{c|}{} & Ours & 82.12 & 0.04 & 80.95 & 0.13 & 80.68 & 0.00 & 81.08 & 0.79 & 80.99 & 0.53 & 80.34 & 0.64 & 80.82 & 0.47 \\ \bottomrule
    \end{tabular}
        }
    }
    \label{tab:bootstrapping_learning}
    \vspace{-3mm}
\end{table}

\cref{tab:bootstrapping_learning}  demonstrate that these methods cannot produce a satisfying result under any clean ratio, while our Bootstrapping Learning works exceptionally well even when the subset is small.
These results again reveal that the backdoor and clean features are tangled in the hidden representation space when the training is restricted to the classification head, as previously illustrated in \cref{sec:poison_suppression}. In this setting, the clean subset also cannot help to remove the backdoor without sacrificing the clean accuracy.
In that sense, it also indicates the necessity of the proactive mindset behind Bootstrapping Learning, whose design aims to explicitly train on clean elements while keeping the clean and poison elements separated in the whole training process.

\begin{table*}[t]
    \centering
    
    \caption{
    {\color{black}
    Performance of our end-to-end defense framework under scenarios where both \threatone~and \threattwo~exist. In this setting, transfer learning is subjected to both encoder poisoning and downstream poisoning attacks concurrently, each with a different backdoor trigger. We report the accuracy, the \textit{attack success rate} of the \textit{encoder} poisoning attack (ASR-E), and the \textit{attack success rate} of the \textit{dataset} poisoning attack (ASR-D) for the classifier, both with and without our defense.}
    }
    \adjustbox{center}{
    \resizebox{1\textwidth}{!}{
    \setlength{\tabcolsep}{0.5pt}
    \begin{tabular}{c|c|c|c|ccc|ccc|ccc|ccc|ccc|ccc|ccc}
    \toprule
    \rowcolor[HTML]{CCCCCC} 
    \cellcolor[HTML]{CCCCCC}{\color[HTML]{08090C} } & \cellcolor[HTML]{CCCCCC}{\color[HTML]{08090C} } & \cellcolor[HTML]{CCCCCC}{\color[HTML]{08090C} } & \cellcolor[HTML]{CCCCCC}{\color[HTML]{08090C} } & \multicolumn{3}{c|}{\cellcolor[HTML]{CCCCCC}{\color[HTML]{08090C} \textbf{BadNets}}} & \multicolumn{3}{c|}{\cellcolor[HTML]{CCCCCC}{\color[HTML]{08090C} \textbf{Blended}}} & \multicolumn{3}{c|}{\cellcolor[HTML]{CCCCCC}{\color[HTML]{08090C} \textbf{SIG}}} & \multicolumn{3}{c|}{\cellcolor[HTML]{CCCCCC}{\color[HTML]{08090C} \textbf{WaNet}}} & \multicolumn{3}{c|}{\cellcolor[HTML]{CCCCCC}{\color[HTML]{08090C} \textbf{TaCT}}} & \multicolumn{3}{c|}{\cellcolor[HTML]{CCCCCC}{\color[HTML]{08090C} \textbf{Adap-Blend}}} & \multicolumn{3}{c}{\cellcolor[HTML]{CCCCCC}{\color[HTML]{08090C} \textbf{Adap-Patch}}} \\ \cline{5-25} 
    \rowcolor[HTML]{CCCCCC} 
    \multirow{-2}{*}{\cellcolor[HTML]{CCCCCC}{\color[HTML]{08090C} \textbf{\makecell{Encoder\\Poisoning}}}} & \multirow{-2}{*}{\textbf{\makecell{\color[HTML]{08090C} Pre-training \\ Dataset}}} & \multirow{-2}{*}{\textbf{\makecell{\color[HTML]{08090C} Downstream \\ Dataset}}} & \multirow{-2}{*}{\cellcolor[HTML]{CCCCCC}{\color[HTML]{08090C} \textbf{\makecell{Dataset\\Poisoning}}}} & {\color[HTML]{08090C} \textbf{ACC}} & {\color[HTML]{08090C} \textbf{ASR-E}} & {\color[HTML]{08090C} \textbf{ASR-D}} & {\color[HTML]{08090C} \textbf{ACC}} & {\color[HTML]{08090C} \textbf{ASR-E}} & {\color[HTML]{08090C} \textbf{ASR-D}} & {\color[HTML]{08090C} \textbf{ACC}} & {\color[HTML]{08090C} \textbf{ASR-E}} & {\color[HTML]{08090C} \textbf{ASR-D}} & {\color[HTML]{08090C} \textbf{ACC}} & {\color[HTML]{08090C} \textbf{ASR-E}} & {\color[HTML]{08090C} \textbf{ASR-D}} & {\color[HTML]{08090C} \textbf{ACC}} & {\color[HTML]{08090C} \textbf{ASR-E}} & {\color[HTML]{08090C} \textbf{ASR-D}} & {\color[HTML]{08090C} \textbf{ACC}} & {\color[HTML]{08090C} \textbf{ASR-E}} & {\color[HTML]{08090C} \textbf{ASR-D}} & {\color[HTML]{08090C} \textbf{ACC}} & {\color[HTML]{08090C} \textbf{ASR-E}} & {\color[HTML]{08090C} \textbf{ASR-D}} \\ \midrule
     &  &  & No Defense & 76.30 & 99.51 & 91.50 & 76.28 & 99.96 & 60.10 & 76.51 & 99.99 & 59.36 & 76.43 & 99.56 & 4.51 & 75.71 & 99.90 & 62.75 & 76.19 & 96.54 & 10.14 & 76.93 & 99.99 & 1.57 \\
     &  & \multirow{-2}{*}{STL-10} & \cellcolor[HTML]{EFEFEF}Ours & \cellcolor[HTML]{EFEFEF}67.75 & \cellcolor[HTML]{EFEFEF}4.67 & \cellcolor[HTML]{EFEFEF}1.00 & \cellcolor[HTML]{EFEFEF}67.04 & \cellcolor[HTML]{EFEFEF}6.85 & \cellcolor[HTML]{EFEFEF}6.68 & \cellcolor[HTML]{EFEFEF}53.10 & \cellcolor[HTML]{EFEFEF}3.88 & \cellcolor[HTML]{EFEFEF}2.53 & \cellcolor[HTML]{EFEFEF}67.54 & \cellcolor[HTML]{EFEFEF}5.11 & \cellcolor[HTML]{EFEFEF}1.82 & \cellcolor[HTML]{EFEFEF}67.46 & \cellcolor[HTML]{EFEFEF}5.72 & \cellcolor[HTML]{EFEFEF}4.25 & \cellcolor[HTML]{EFEFEF}68.75 & \cellcolor[HTML]{EFEFEF}6.65 & \cellcolor[HTML]{EFEFEF}1.40 & \cellcolor[HTML]{EFEFEF}68.28 & \cellcolor[HTML]{EFEFEF}6.03 & \cellcolor[HTML]{EFEFEF}6.22 \\ \cline{3-25} 
     &  &  & No Defense & 72.60 & 99.24 & 93.75 & 73.22 & 99.77 & 86.36 & 73.16 & 99.15 & 74.81 & 78.17 & 99.94 & 6.09 & 73.86 & 99.20 & 91.73 & 72.98 & 95.95 & 65.60 & 72.22 & 99.69 & 28.43 \\
     &  & \multirow{-2}{*}{GTSRB} & \cellcolor[HTML]{EFEFEF}Ours & \cellcolor[HTML]{EFEFEF}90.54 & \cellcolor[HTML]{EFEFEF}0.01 & \cellcolor[HTML]{EFEFEF}1.38 & \cellcolor[HTML]{EFEFEF}88.27 & \cellcolor[HTML]{EFEFEF}0.31 & \cellcolor[HTML]{EFEFEF}5.05 & \cellcolor[HTML]{EFEFEF}91.69 & \cellcolor[HTML]{EFEFEF}0.00 & \cellcolor[HTML]{EFEFEF}0.98 & \cellcolor[HTML]{EFEFEF}91.88 & \cellcolor[HTML]{EFEFEF}0.04 & \cellcolor[HTML]{EFEFEF}0.66 & \cellcolor[HTML]{EFEFEF}92.60 & \cellcolor[HTML]{EFEFEF}0.80 & \cellcolor[HTML]{EFEFEF}0.00 & \cellcolor[HTML]{EFEFEF}87.79 & \cellcolor[HTML]{EFEFEF}0.00 & \cellcolor[HTML]{EFEFEF}3.30 & \cellcolor[HTML]{EFEFEF}93.90 & \cellcolor[HTML]{EFEFEF}0.27 & \cellcolor[HTML]{EFEFEF}0.29 \\ \cline{3-25} 
     &  &  & No Defense & 68.47 & 98.80 & 99.27 & 67.98 & 98.95 & 98.11 & 68.19 & 98.70 & 96.63 & 67.99 & 98.78 & 11.86 & 68.19 & 98.80 & 94.12 & 68.07 & 98.81 & 90.81 & 68.26 & 97.90 & 71.75 \\
     & \multirow{-6}{*}{CIFAR-10} & \multirow{-2}{*}{SVHN} & \cellcolor[HTML]{EFEFEF}Ours & \cellcolor[HTML]{EFEFEF}92.19 & \cellcolor[HTML]{EFEFEF}4.29 & \cellcolor[HTML]{EFEFEF}3.79 & \cellcolor[HTML]{EFEFEF}92.19 & \cellcolor[HTML]{EFEFEF}4.29 & \cellcolor[HTML]{EFEFEF}0.10 & \cellcolor[HTML]{EFEFEF}92.80 & \cellcolor[HTML]{EFEFEF}4.80 & \cellcolor[HTML]{EFEFEF}0.65 & \cellcolor[HTML]{EFEFEF}90.20 & \cellcolor[HTML]{EFEFEF}7.94 & \cellcolor[HTML]{EFEFEF}2.76 & \cellcolor[HTML]{EFEFEF}91.51 & \cellcolor[HTML]{EFEFEF}2.49 & \cellcolor[HTML]{EFEFEF}0.75 & \cellcolor[HTML]{EFEFEF}90.30 & \cellcolor[HTML]{EFEFEF}4.23 & \cellcolor[HTML]{EFEFEF}0.14 & \cellcolor[HTML]{EFEFEF}92.72 & \cellcolor[HTML]{EFEFEF}4.86 & \cellcolor[HTML]{EFEFEF}0.07 \\ \cline{2-25} 
     &  &  & No Defense & 69.56 & 97.88 & 78.00 & 70.33 & 98.39 & 71.98 & 69.72 & 99.83 & 77.42 & 69.94 & 99.82 & 9.12 & 69.66 & 99.66 & 70.00 & 69.84 & 99.77 & 16.28 & 70.03 & 99.76 & 5.78 \\
     &  & \multirow{-2}{*}{CIFAR-10} & \cellcolor[HTML]{EFEFEF}Ours & \cellcolor[HTML]{EFEFEF}63.27 & \cellcolor[HTML]{EFEFEF}5.76 & \cellcolor[HTML]{EFEFEF}4.76 & \cellcolor[HTML]{EFEFEF}62.73 & \cellcolor[HTML]{EFEFEF}6.28 & \cellcolor[HTML]{EFEFEF}4.97 & \cellcolor[HTML]{EFEFEF}68.42 & \cellcolor[HTML]{EFEFEF}8.29 & \cellcolor[HTML]{EFEFEF}3.64 & \cellcolor[HTML]{EFEFEF}62.63 & \cellcolor[HTML]{EFEFEF}6.61 & \cellcolor[HTML]{EFEFEF}4.47 & \cellcolor[HTML]{EFEFEF}65.47 & \cellcolor[HTML]{EFEFEF}6.36 & \cellcolor[HTML]{EFEFEF}0.00 & \cellcolor[HTML]{EFEFEF}64.38 & \cellcolor[HTML]{EFEFEF}7.71 & \cellcolor[HTML]{EFEFEF}2.03 & \cellcolor[HTML]{EFEFEF}63.05 & \cellcolor[HTML]{EFEFEF}6.08 & \cellcolor[HTML]{EFEFEF}0.13 \\ \cline{3-25} 
     &  &  & No Defense & 70.67 & 97.52 & 83.43 & 69.59 & 98.77 & 82.33 & 70.86 & 99.19 & 74.56 & 69.63 & 99.80 & 4.33 & 68.33 & 98.05 & 81.07 & 68.56 & 99.10 & 54.45 & 69.58 & 98.95 & 12.30 \\
     &  & \multirow{-2}{*}{GTSRB} & \cellcolor[HTML]{EFEFEF}Ours & \cellcolor[HTML]{EFEFEF}85.65 & \cellcolor[HTML]{EFEFEF}0.11 & \cellcolor[HTML]{EFEFEF}5.45 & \cellcolor[HTML]{EFEFEF}86.03 & \cellcolor[HTML]{EFEFEF}0.70 & \cellcolor[HTML]{EFEFEF}0.87 & \cellcolor[HTML]{EFEFEF}85.18 & \cellcolor[HTML]{EFEFEF}1.73 & \cellcolor[HTML]{EFEFEF}0.24 & \cellcolor[HTML]{EFEFEF}85.27 & \cellcolor[HTML]{EFEFEF}0.22 & \cellcolor[HTML]{EFEFEF}4.39 & \cellcolor[HTML]{EFEFEF}86.03 & \cellcolor[HTML]{EFEFEF}0.05 & \cellcolor[HTML]{EFEFEF}1.06 & \cellcolor[HTML]{EFEFEF}85.58 & \cellcolor[HTML]{EFEFEF}1.10 & \cellcolor[HTML]{EFEFEF}5.13 & \cellcolor[HTML]{EFEFEF}87.05 & \cellcolor[HTML]{EFEFEF}1.80 & \cellcolor[HTML]{EFEFEF}1.52 \\ \cline{3-25} 
     &  &  & No Defense & 67.44 & 85.95 & 98.85 & 66.29 & 85.93 & 98.93 & 67.45 & 88.96 & 93.92 & 64.88 & 84.07 & 11.91 & 67.78 & 87.69 & 94.53 & 67.60 & 81.29 & 89.94 & 66.77 & 80.30 & 26.85 \\
    \multirow{-12}{*}{\textbf{BadEncoder}} & \multirow{-6}{*}{STL-10} & \multirow{-2}{*}{SVHN} & \cellcolor[HTML]{EFEFEF}Ours & \cellcolor[HTML]{EFEFEF}83.90 & \cellcolor[HTML]{EFEFEF}4.30 & \cellcolor[HTML]{EFEFEF}10.10 & \cellcolor[HTML]{EFEFEF}86.63 & \cellcolor[HTML]{EFEFEF}3.72 & \cellcolor[HTML]{EFEFEF}5.32 & \cellcolor[HTML]{EFEFEF}85.96 & \cellcolor[HTML]{EFEFEF}9.18 & \cellcolor[HTML]{EFEFEF}2.55 & \cellcolor[HTML]{EFEFEF}88.96 & \cellcolor[HTML]{EFEFEF}5.10 & \cellcolor[HTML]{EFEFEF}1.01 & \cellcolor[HTML]{EFEFEF}86.34 & \cellcolor[HTML]{EFEFEF}3.15 & \cellcolor[HTML]{EFEFEF}0.31 & \cellcolor[HTML]{EFEFEF}86.40 & \cellcolor[HTML]{EFEFEF}4.87 & \cellcolor[HTML]{EFEFEF}2.09 & \cellcolor[HTML]{EFEFEF}86.92 & \cellcolor[HTML]{EFEFEF}6.15 & \cellcolor[HTML]{EFEFEF}4.50 \\ \midrule
     &  &  & No Defense & 71.94 & 99.43 & 75.22 & 71.09 & 98.00 & 53.97 & 72.49 & 93.63 & 35.50 & 72.08 & 90.18 & 10.14 & 71.78 & 97.54 & 49.75 & 71.34 & 99.39 & 11.42 & 71.63 & 98.35 & 1.89 \\
     &  & \multirow{-2}{*}{STL-10} & \cellcolor[HTML]{EFEFEF}Ours & \cellcolor[HTML]{EFEFEF}63.16 & \cellcolor[HTML]{EFEFEF}14.90 & \cellcolor[HTML]{EFEFEF}10.92 & \cellcolor[HTML]{EFEFEF}68.30 & \cellcolor[HTML]{EFEFEF}10.89 & \cellcolor[HTML]{EFEFEF}5.89 & \cellcolor[HTML]{EFEFEF}64.34 & \cellcolor[HTML]{EFEFEF}7.49 & \cellcolor[HTML]{EFEFEF}0.49 & \cellcolor[HTML]{EFEFEF}64.59 & \cellcolor[HTML]{EFEFEF}6.38 & \cellcolor[HTML]{EFEFEF}4.29 & \cellcolor[HTML]{EFEFEF}63.63 & \cellcolor[HTML]{EFEFEF}11.24 & \cellcolor[HTML]{EFEFEF}13.00 & \cellcolor[HTML]{EFEFEF}64.74 & \cellcolor[HTML]{EFEFEF}7.92 & \cellcolor[HTML]{EFEFEF}2.67 & \cellcolor[HTML]{EFEFEF}65.00 & \cellcolor[HTML]{EFEFEF}7.39 & \cellcolor[HTML]{EFEFEF}2.96 \\ \cline{3-25} 
     &  &  & No Defense & 74.35 & 73.36 & 94.19 & 74.57 & 72.99 & 87.63 & 74.95 & 74.70 & 69.57 & 74.48 & 73.02 & 6.58 & 74.67 & 72.91 & 87.07 & 73.95 & 73.01 & 61.30 & 73.76 & 72.97 & 14.79 \\
     &  & \multirow{-2}{*}{GTSRB} & \cellcolor[HTML]{EFEFEF}Ours & \cellcolor[HTML]{EFEFEF}87.98 & \cellcolor[HTML]{EFEFEF}7.05 & \cellcolor[HTML]{EFEFEF}3.16 & \cellcolor[HTML]{EFEFEF}90.17 & \cellcolor[HTML]{EFEFEF}7.23 & \cellcolor[HTML]{EFEFEF}6.66 & \cellcolor[HTML]{EFEFEF}88.16 & \cellcolor[HTML]{EFEFEF}3.18 & \cellcolor[HTML]{EFEFEF}0.74 & \cellcolor[HTML]{EFEFEF}89.14 & \cellcolor[HTML]{EFEFEF}3.61 & \cellcolor[HTML]{EFEFEF}0.47 & \cellcolor[HTML]{EFEFEF}89.93 & \cellcolor[HTML]{EFEFEF}5.82 & \cellcolor[HTML]{EFEFEF}6.82 & \cellcolor[HTML]{EFEFEF}89.14 & \cellcolor[HTML]{EFEFEF}5.05 & \cellcolor[HTML]{EFEFEF}7.63 & \cellcolor[HTML]{EFEFEF}89.87 & \cellcolor[HTML]{EFEFEF}3.10 & \cellcolor[HTML]{EFEFEF}1.85 \\ \cline{3-25} 
     &  &  & No Defense & 71.35 & 75.53 & 99.45 & 71.37 & 75.74 & 97.60 & 71.21 & 75.81 & 94.45 & 71.04 & 76.95 & 11.60 & 71.31 & 72.91 & 96.35 & 71.26 & 77.03 & 85.17 & 71.09 & 76.30 & 51.23 \\
     & \multirow{-6}{*}{CIFAR-10} & \multirow{-2}{*}{SVHN} & \cellcolor[HTML]{EFEFEF}Ours & \cellcolor[HTML]{EFEFEF}89.54 & \cellcolor[HTML]{EFEFEF}9.64 & \cellcolor[HTML]{EFEFEF}6.78 & \cellcolor[HTML]{EFEFEF}88.73 & \cellcolor[HTML]{EFEFEF}6.92 & \cellcolor[HTML]{EFEFEF}4.90 & \cellcolor[HTML]{EFEFEF}89.02 & \cellcolor[HTML]{EFEFEF}9.88 & \cellcolor[HTML]{EFEFEF}4.32 & \cellcolor[HTML]{EFEFEF}87.19 & \cellcolor[HTML]{EFEFEF}6.66 & \cellcolor[HTML]{EFEFEF}3.66 & \cellcolor[HTML]{EFEFEF}92.34 & \cellcolor[HTML]{EFEFEF}3.60 & \cellcolor[HTML]{EFEFEF}2.77 & \cellcolor[HTML]{EFEFEF}89.20 & \cellcolor[HTML]{EFEFEF}5.10 & \cellcolor[HTML]{EFEFEF}1.01 & \cellcolor[HTML]{EFEFEF}89.70 & \cellcolor[HTML]{EFEFEF}5.04 & \cellcolor[HTML]{EFEFEF}2.97 \\ \cline{2-25} 
     &  &  & No Defense & 70.26 & 78.54 & 74.24 & 70.71 & 77.58 & 74.19 & 70.83 & 79.10 & 69.62 & 70.87 & 78.66 & 9.27 & 70.62 & 78.55 & 69.00 & 70.81 & 78.63 & 14.13 & 71.15 & 78.63 & 4.93 \\
     &  & \multirow{-2}{*}{CIFAR-10} & \cellcolor[HTML]{EFEFEF}Ours & \cellcolor[HTML]{EFEFEF}64.74 & \cellcolor[HTML]{EFEFEF}6.87 & \cellcolor[HTML]{EFEFEF}7.43 & \cellcolor[HTML]{EFEFEF}63.46 & \cellcolor[HTML]{EFEFEF}7.53 & \cellcolor[HTML]{EFEFEF}7.69 & \cellcolor[HTML]{EFEFEF}67.31 & \cellcolor[HTML]{EFEFEF}4.94 & \cellcolor[HTML]{EFEFEF}1.91 & \cellcolor[HTML]{EFEFEF}66.18 & \cellcolor[HTML]{EFEFEF}4.02 & \cellcolor[HTML]{EFEFEF}1.73 & \cellcolor[HTML]{EFEFEF}66.28 & \cellcolor[HTML]{EFEFEF}5.49 & \cellcolor[HTML]{EFEFEF}0.10 & \cellcolor[HTML]{EFEFEF}62.63 & \cellcolor[HTML]{EFEFEF}4.96 & \cellcolor[HTML]{EFEFEF}3.40 & \cellcolor[HTML]{EFEFEF}63.56 & \cellcolor[HTML]{EFEFEF}3.31 & \cellcolor[HTML]{EFEFEF}6.31 \\ \cline{3-25} 
     &  &  & No Defense & 63.40 & 78.25 & 90.50 & 63.71 & 84.92 & 88.70 & 64.29 & 85.40 & 74.55 & 63.99 & 78.12 & 6.09 & 63.47 & 86.80 & 78.54 & 61.18 & 80.32 & 67.40 & 62.00 & 79.83 & 18.46 \\
     &  & \multirow{-2}{*}{GTSRB} & \cellcolor[HTML]{EFEFEF}Ours & \cellcolor[HTML]{EFEFEF}86.10 & \cellcolor[HTML]{EFEFEF}0.21 & \cellcolor[HTML]{EFEFEF}3.94 & \cellcolor[HTML]{EFEFEF}87.08 & \cellcolor[HTML]{EFEFEF}1.42 & \cellcolor[HTML]{EFEFEF}5.85 & \cellcolor[HTML]{EFEFEF}86.44 & \cellcolor[HTML]{EFEFEF}2.82 & \cellcolor[HTML]{EFEFEF}0.03 & \cellcolor[HTML]{EFEFEF}84.47 & \cellcolor[HTML]{EFEFEF}1.00 & \cellcolor[HTML]{EFEFEF}3.18 & \cellcolor[HTML]{EFEFEF}82.18 & \cellcolor[HTML]{EFEFEF}0.25 & \cellcolor[HTML]{EFEFEF}5.45 & \cellcolor[HTML]{EFEFEF}81.90 & \cellcolor[HTML]{EFEFEF}1.61 & \cellcolor[HTML]{EFEFEF}2.95 & \cellcolor[HTML]{EFEFEF}81.32 & \cellcolor[HTML]{EFEFEF}0.62 & \cellcolor[HTML]{EFEFEF}7.58 \\ \cline{3-25} 
     &  &  & No Defense & 59.12 & 94.66 & 96.56 & 59.77 & 97.48 & 97.43 & 58.03 & 92.94 & 91.53 & 59.77 & 95.08 & 15.17 & 59.47 & 97.46 & 92.33 & 60.02 & 98.69 & 84.58 & 59.74 & 96.81 & 16.52 \\
    \multirow{-12}{*}{\textbf{DRUPE}} & \multirow{-6}{*}{STL-10} & \multirow{-2}{*}{SVHN} & \cellcolor[HTML]{EFEFEF}Ours & \cellcolor[HTML]{EFEFEF}82.13 & \cellcolor[HTML]{EFEFEF}5.95 & \cellcolor[HTML]{EFEFEF}6.25 & \cellcolor[HTML]{EFEFEF}83.22 & \cellcolor[HTML]{EFEFEF}4.03 & \cellcolor[HTML]{EFEFEF}4.56 & \cellcolor[HTML]{EFEFEF}83.75 & \cellcolor[HTML]{EFEFEF}9.64 & \cellcolor[HTML]{EFEFEF}2.77 & \cellcolor[HTML]{EFEFEF}82.76 & \cellcolor[HTML]{EFEFEF}2.45 & \cellcolor[HTML]{EFEFEF}3.59 & \cellcolor[HTML]{EFEFEF}83.85 & \cellcolor[HTML]{EFEFEF}2.93 & \cellcolor[HTML]{EFEFEF}0.98 & \cellcolor[HTML]{EFEFEF}81.13 & \cellcolor[HTML]{EFEFEF}9.01 & \cellcolor[HTML]{EFEFEF}5.10 & \cellcolor[HTML]{EFEFEF}83.17 & \cellcolor[HTML]{EFEFEF}3.05 & \cellcolor[HTML]{EFEFEF}1.65 \\ \midrule
     &  &  & No Defense & 51.80 & 9.97 & 56.39 & 53.20 & 10.15 & 17.67 & 53.14 & 8.99 & 18.96 & 53.43 & 10.19 & 5.75 & 52.06 & 10.83 & 40.25 & 53.28 & 8.79 & 4.99 & 52.88 & 8.65 & 4.64 \\
     & \multirow{-2}{*}{STL-10} & \multirow{-2}{*}{STL-10} & \cellcolor[HTML]{EFEFEF}Ours & \cellcolor[HTML]{EFEFEF}48.81 & \cellcolor[HTML]{EFEFEF}1.79 & \cellcolor[HTML]{EFEFEF}1.21 & \cellcolor[HTML]{EFEFEF}49.80 & \cellcolor[HTML]{EFEFEF}2.23 & \cellcolor[HTML]{EFEFEF}2.51 & \cellcolor[HTML]{EFEFEF}45.56 & \cellcolor[HTML]{EFEFEF}3.57 & \cellcolor[HTML]{EFEFEF}0.99 & \cellcolor[HTML]{EFEFEF}49.68 & \cellcolor[HTML]{EFEFEF}0.24 & \cellcolor[HTML]{EFEFEF}2.44 & \cellcolor[HTML]{EFEFEF}50.43 & \cellcolor[HTML]{EFEFEF}2.58 & \cellcolor[HTML]{EFEFEF}3.51 & \cellcolor[HTML]{EFEFEF}49.63 & \cellcolor[HTML]{EFEFEF}2.44 & \cellcolor[HTML]{EFEFEF}5.93 & \cellcolor[HTML]{EFEFEF}48.44 & \cellcolor[HTML]{EFEFEF}1.03 & \cellcolor[HTML]{EFEFEF}1.96 \\ \cline{2-25} 
     &  &  & No Defense & 75.33 & 48.93 & 93.29 & 75.22 & 44.92 & 50.59 & 76.19 & 38.54 & 52.56 & 74.31 & 41.14 & 13.98 & 75.80 & 48.87 & 84.00 & 75.08 & 49.41 & 16.93 & 75.93 & 46.74 & 14.90 \\
     & \multirow{-2}{*}{CIFAR-10} & \multirow{-2}{*}{CIFAR-10} & \cellcolor[HTML]{EFEFEF}Ours & \cellcolor[HTML]{EFEFEF}61.78 & \cellcolor[HTML]{EFEFEF}0.94 & \cellcolor[HTML]{EFEFEF}0.10 & \cellcolor[HTML]{EFEFEF}62.72 & \cellcolor[HTML]{EFEFEF}2.87 & \cellcolor[HTML]{EFEFEF}6.88 & \cellcolor[HTML]{EFEFEF}61.94 & \cellcolor[HTML]{EFEFEF}0.59 & \cellcolor[HTML]{EFEFEF}0.00 & \cellcolor[HTML]{EFEFEF}62.32 & \cellcolor[HTML]{EFEFEF}3.60 & \cellcolor[HTML]{EFEFEF}3.44 & \cellcolor[HTML]{EFEFEF}62.79 & \cellcolor[HTML]{EFEFEF}0.57 & \cellcolor[HTML]{EFEFEF}0.70 & \cellcolor[HTML]{EFEFEF}63.51 & \cellcolor[HTML]{EFEFEF}8.78 & \cellcolor[HTML]{EFEFEF}8.84 & \cellcolor[HTML]{EFEFEF}58.80 & \cellcolor[HTML]{EFEFEF}0.63 & \cellcolor[HTML]{EFEFEF}6.72 \\ \cline{2-25} 
     &  &  & No Defense & 64.68 & 4.60 & 73.81 & 67.16 & 3.05 & 69.37 & 65.83 & 4.96 & 60.30 & 66.00 & 3.76 & 4.67 & 65.86 & 5.82 & 62.40 & 66.94 & 0.94 & 52.71 & 65.13 & 4.84 & 21.36 \\
    \multirow{-6}{*}{\textbf{CTRL}} & \multirow{-2}{*}{GTSRB} & \multirow{-2}{*}{GTSRB} & \cellcolor[HTML]{EFEFEF}Ours & \cellcolor[HTML]{EFEFEF}85.30 & \cellcolor[HTML]{EFEFEF}0.09 & \cellcolor[HTML]{EFEFEF}2.79 & \cellcolor[HTML]{EFEFEF}88.28 & \cellcolor[HTML]{EFEFEF}0.19 & \cellcolor[HTML]{EFEFEF}0.14 & \cellcolor[HTML]{EFEFEF}87.05 & \cellcolor[HTML]{EFEFEF}0.00 & \cellcolor[HTML]{EFEFEF}0.07 & \cellcolor[HTML]{EFEFEF}87.82 & \cellcolor[HTML]{EFEFEF}0.08 & \cellcolor[HTML]{EFEFEF}0.34 & \cellcolor[HTML]{EFEFEF}88.37 & \cellcolor[HTML]{EFEFEF}0.15 & \cellcolor[HTML]{EFEFEF}0.40 & \cellcolor[HTML]{EFEFEF}88.13 & \cellcolor[HTML]{EFEFEF}1.77 & \cellcolor[HTML]{EFEFEF}4.97 & \cellcolor[HTML]{EFEFEF}87.73 & \cellcolor[HTML]{EFEFEF}0.11 & \cellcolor[HTML]{EFEFEF}0.03 \\ \midrule
     &  &  & No Defense & 84.47 & 19.15 & 84.79 & 83.88 & 24.97 & 32.85 & 82.85 & 45.88 & 60.00 & 81.18 & 24.61 & 2.48 & 83.12 & 40.18 & 93.00 & 82.47 & 26.36 & 26.42 & 82.94 & 33.70 & 2.24 \\
    \multirow{-2}{*}{\textbf{SSLBackdoor}} & \multirow{-2}{*}{ImageNet} & \multirow{-2}{*}{ImageNet-10} & \cellcolor[HTML]{EFEFEF}Ours & \cellcolor[HTML]{EFEFEF}80.65 & \cellcolor[HTML]{EFEFEF}2.61 & \cellcolor[HTML]{EFEFEF}3.15 & \cellcolor[HTML]{EFEFEF}80.82 & \cellcolor[HTML]{EFEFEF}1.21 & \cellcolor[HTML]{EFEFEF}6.55 & \cellcolor[HTML]{EFEFEF}79.94 & \cellcolor[HTML]{EFEFEF}3.21 & \cellcolor[HTML]{EFEFEF}4.97 & \cellcolor[HTML]{EFEFEF}78.00 & \cellcolor[HTML]{EFEFEF}3.21 & \cellcolor[HTML]{EFEFEF}2.24 & \cellcolor[HTML]{EFEFEF}83.47 & \cellcolor[HTML]{EFEFEF}1.88 & \cellcolor[HTML]{EFEFEF}3.21 & \cellcolor[HTML]{EFEFEF}83.18 & \cellcolor[HTML]{EFEFEF}0.67 & \cellcolor[HTML]{EFEFEF}3.82 & \cellcolor[HTML]{EFEFEF}83.47 & \cellcolor[HTML]{EFEFEF}1.82 & \cellcolor[HTML]{EFEFEF}2.61 \\ \midrule
     &  &  & No Defense & 84.41 & 34.91 & 82.73 & 83.76 & 57.45 & 26.73 & 82.53 & 54.18 & 61.82 & 81.41 & 44.06 & 3.94 & 83.76 & 60.12 & 91.00 & 80.06 & 53.09 & 15.03 & 83.47 & 57.27 & 3.45 \\
    \multirow{-2}{*}{\textbf{CorruptEncoder}} & \multirow{-2}{*}{ImageNet} & \multirow{-2}{*}{ImageNet-10} & \cellcolor[HTML]{EFEFEF}Ours & \cellcolor[HTML]{EFEFEF}82.35 & \cellcolor[HTML]{EFEFEF}1.33 & \cellcolor[HTML]{EFEFEF}4.55 & \cellcolor[HTML]{EFEFEF}82.35 & \cellcolor[HTML]{EFEFEF}2.00 & \cellcolor[HTML]{EFEFEF}4.61 & \cellcolor[HTML]{EFEFEF}82.06 & \cellcolor[HTML]{EFEFEF}1.33 & \cellcolor[HTML]{EFEFEF}5.88 & \cellcolor[HTML]{EFEFEF}80.00 & \cellcolor[HTML]{EFEFEF}1.58 & \cellcolor[HTML]{EFEFEF}3.15 & \cellcolor[HTML]{EFEFEF}81.65 & \cellcolor[HTML]{EFEFEF}1.21 & \cellcolor[HTML]{EFEFEF}6.50 & \cellcolor[HTML]{EFEFEF}82.53 & \cellcolor[HTML]{EFEFEF}0.97 & \cellcolor[HTML]{EFEFEF}6.36 & \cellcolor[HTML]{EFEFEF}84.06 & \cellcolor[HTML]{EFEFEF}1.82 & \cellcolor[HTML]{EFEFEF}1.76 \\

    \bottomrule
    \end{tabular}
    }
    }
    \label{tab:threat4-end2end}
    \vspace{-2mm}
\end{table*}

\begin{table}[t]
    \centering
    \caption{Peformance of our entire end-to-end defense framework under \threattwo.}
          \adjustbox{width=0.48\textwidth, center}{\large
          \setlength{\tabcolsep}{0.5pt}
          \begin{tabular}{c|c|cc|cc|cc|cc|cc|cc|cc}
          \toprule
          \rowcolor[HTML]{CCCCCC} 
          \cellcolor[HTML]{CCCCCC}{\color[HTML]{08090C} } & \cellcolor[HTML]{CCCCCC}{\color[HTML]{08090C} } & \multicolumn{2}{c|}{\cellcolor[HTML]{CCCCCC}{\color[HTML]{08090C} \textbf{BadNets}}} & \multicolumn{2}{c|}{\cellcolor[HTML]{CCCCCC}{\color[HTML]{08090C} \textbf{Blended}}} & \multicolumn{2}{c|}{\cellcolor[HTML]{CCCCCC}{\color[HTML]{08090C} \textbf{SIG}}} & \multicolumn{2}{c|}{\cellcolor[HTML]{CCCCCC}{\color[HTML]{08090C} \textbf{WaNet}}} & \multicolumn{2}{c|}{\cellcolor[HTML]{CCCCCC}{\color[HTML]{08090C} \textbf{TaCT}}} & \multicolumn{2}{c|}{\cellcolor[HTML]{CCCCCC}{\color[HTML]{08090C} \textbf{Adap-Blend}}} & \multicolumn{2}{c}{\cellcolor[HTML]{CCCCCC}{\color[HTML]{08090C} \textbf{Adap-Patch}}} \\ \cline{3-16} 
          \rowcolor[HTML]{CCCCCC} 
          \multirow{-2}{*}{\cellcolor[HTML]{CCCCCC}{\color[HTML]{08090C} \textbf{Dataset}}} & \multirow{-2}{*}{\cellcolor[HTML]{CCCCCC}{\textbf{\makecell{\color[HTML]{08090C}Dataset\\ Poisoning}}}} & {\color[HTML]{08090C} \textbf{ACC$\uparrow$}} & {\color[HTML]{08090C} \textbf{ASR$\downarrow$}} & {\color[HTML]{08090C} \textbf{ACC$\uparrow$}} & {\color[HTML]{08090C} \textbf{ASR$\downarrow$}} & {\color[HTML]{08090C} \textbf{ACC$\uparrow$}} & {\color[HTML]{08090C} \textbf{ASR$\downarrow$}} & {\color[HTML]{08090C} \textbf{ACC$\uparrow$}} & {\color[HTML]{08090C} \textbf{ASR$\downarrow$}} & {\color[HTML]{08090C} \textbf{ACC$\uparrow$}} & {\color[HTML]{08090C} \textbf{ASR$\downarrow$}} & {\color[HTML]{08090C} \textbf{ACC$\uparrow$}} & {\color[HTML]{08090C} \textbf{ASR$\downarrow$}} & {\color[HTML]{08090C} \textbf{ACC$\uparrow$}} & {\color[HTML]{08090C} \textbf{ASR$\downarrow$}} \\ \midrule
          & No Defense & 75.64 & 90.24 & 75.65 & 50.35 & 76.51 & 59.97 & 76.21 & 4.76 & 75.19 & 64.13 & 75.75 & 9.04 & 76.43 & 1.92 \\ \cline{2-16} 
          \multirow{-2}{*}{STL-10} & \cellcolor[HTML]{EFEFEF}Ours & \cellcolor[HTML]{EFEFEF}64.08 & \cellcolor[HTML]{EFEFEF}2.15 & \cellcolor[HTML]{EFEFEF}65.59 & \cellcolor[HTML]{EFEFEF}1.60 & \cellcolor[HTML]{EFEFEF}62.85 & \cellcolor[HTML]{EFEFEF}6.00 & \cellcolor[HTML]{EFEFEF}64.55 & \cellcolor[HTML]{EFEFEF}1.60 & \cellcolor[HTML]{EFEFEF}66.26 & \cellcolor[HTML]{EFEFEF}1.00 & \cellcolor[HTML]{EFEFEF}65.93 & \cellcolor[HTML]{EFEFEF}3.24 & \cellcolor[HTML]{EFEFEF}62.55 & \cellcolor[HTML]{EFEFEF}1.08 \\ \hline
          & No Defense & 85.04 & 92.21 & 84.84 & 89.12 & 84.72 & 89.10 & 84.40 & 9.11 & 84.28 & 82.60 & 83.39 & 34.34 & 84.16 & 5.66 \\ \cline{2-16} 
          \multirow{-2}{*}{CIFAR-10} & \cellcolor[HTML]{EFEFEF}Ours & \cellcolor[HTML]{EFEFEF}87.38 & \cellcolor[HTML]{EFEFEF}3.48 & \cellcolor[HTML]{EFEFEF}87.35 & \cellcolor[HTML]{EFEFEF}5.90 & \cellcolor[HTML]{EFEFEF}87.31 & \cellcolor[HTML]{EFEFEF}2.54 & \cellcolor[HTML]{EFEFEF}87.58 & \cellcolor[HTML]{EFEFEF}0.23 & \cellcolor[HTML]{EFEFEF}89.04 & \cellcolor[HTML]{EFEFEF}0.10 & \cellcolor[HTML]{EFEFEF}87.31 & \cellcolor[HTML]{EFEFEF}2.54 & \cellcolor[HTML]{EFEFEF}87.38 & \cellcolor[HTML]{EFEFEF}3.48 \\ \hline
          & No Defense & 81.79 & 95.02 & 81.30 & 90.39 & 81.90 & 74.37 & 80.74 & 8.81 & 81.95 & 89.20 & 80.85 & 69.73 & 78.54 & 28.20 \\ \cline{2-16} 
          \multirow{-2}{*}{GTSRB} & \cellcolor[HTML]{EFEFEF}Ours & \cellcolor[HTML]{EFEFEF}92.03 & \cellcolor[HTML]{EFEFEF}1.31 & \cellcolor[HTML]{EFEFEF}91.37 & \cellcolor[HTML]{EFEFEF}3.04 & \cellcolor[HTML]{EFEFEF}94.13 & \cellcolor[HTML]{EFEFEF}0.38 & \cellcolor[HTML]{EFEFEF}91.10 & \cellcolor[HTML]{EFEFEF}1.31 & \cellcolor[HTML]{EFEFEF}91.82 & \cellcolor[HTML]{EFEFEF}1.87 & \cellcolor[HTML]{EFEFEF}90.87 & \cellcolor[HTML]{EFEFEF}0.62 & \cellcolor[HTML]{EFEFEF}92.25 & \cellcolor[HTML]{EFEFEF}1.09 \\ \hline
          & No Defense & 59.80 & 99.42 & 60.11 & 98.30 & 59.83 & 97.58 & 59.65 & 15.77 & 59.91 & 91.90 & 59.84 & 89.90 & 59.87 & 70.86 \\ \cline{2-16} 
          \multirow{-2}{*}{SVHN} & \cellcolor[HTML]{EFEFEF}Ours & \cellcolor[HTML]{EFEFEF}91.19 & \cellcolor[HTML]{EFEFEF}4.14 & \cellcolor[HTML]{EFEFEF}90.88 & \cellcolor[HTML]{EFEFEF}6.82 & \cellcolor[HTML]{EFEFEF}91.09 & \cellcolor[HTML]{EFEFEF}3.22 & \cellcolor[HTML]{EFEFEF}90.11 & \cellcolor[HTML]{EFEFEF}1.45 & \cellcolor[HTML]{EFEFEF}91.25 & \cellcolor[HTML]{EFEFEF}2.92 & \cellcolor[HTML]{EFEFEF}90.22 & \cellcolor[HTML]{EFEFEF}1.31 & \cellcolor[HTML]{EFEFEF}90.95 & \cellcolor[HTML]{EFEFEF}1.23 \\ \hline
          & No Defense & 85.06 & 92.85 & 85.00 & 40.42 & 86.29 & 55.33 & 85.71 & 3.33 & 85.88 & 95.00 & 86.35 & 24.06 & 85.71 & 6.48 \\ \cline{2-16} 
          \multirow{-2}{*}{ImageNet-10} & \cellcolor[HTML]{EFEFEF}Ours & \cellcolor[HTML]{EFEFEF}80.46 & \cellcolor[HTML]{EFEFEF}3.86 & \cellcolor[HTML]{EFEFEF}81.65 & \cellcolor[HTML]{EFEFEF}2.42 & \cellcolor[HTML]{EFEFEF}82.00 & \cellcolor[HTML]{EFEFEF}2.85 & \cellcolor[HTML]{EFEFEF}83.71 & \cellcolor[HTML]{EFEFEF}0.94 & \cellcolor[HTML]{EFEFEF}84.53 & \cellcolor[HTML]{EFEFEF}3.33 & \cellcolor[HTML]{EFEFEF}80.24 & \cellcolor[HTML]{EFEFEF}1.94 & \cellcolor[HTML]{EFEFEF}81.71 & \cellcolor[HTML]{EFEFEF}2.48 \\ \bottomrule
          \end{tabular}
            \label{tab:threat2-end2end}
        }
        \vspace{-2mm}
\end{table}

\begin{table}[t]
    \centering
    \caption{Peformance of our entire end-to-end defense framework under \threatone~and \threatthree.}
    \adjustbox{width=0.45\textwidth, center}{\large
    \begin{tabular}{cccc|cc|cc}
    \toprule
    \rowcolor[HTML]{CCCCCC} 
    \multicolumn{4}{c|}{\cellcolor[HTML]{CCCCCC}{\color[HTML]{08090C} \textbf{Threat Type}}} & \multicolumn{2}{c|}{\cellcolor[HTML]{CCCCCC}{\color[HTML]{08090C} \textbf{\threatone}}} & \multicolumn{2}{c}{\cellcolor[HTML]{CCCCCC}{\color[HTML]{08090C} \textbf{\threatthree}}} \\ \midrule
    \rowcolor[HTML]{CCCCCC} 
    \multicolumn{1}{c|}{\cellcolor[HTML]{CCCCCC}{\color[HTML]{08090C} \textbf{\makecell{\ Encoder\ \\\ \ \ \ \ Poisoning\ \ \ \ \ }}}} & \multicolumn{1}{c|}{\cellcolor[HTML]{CCCCCC}{\color[HTML]{08090C} \textbf{\begin{tabular}[c]{@{}c@{}}Pre-training \\ Dataset\end{tabular}}}} & \multicolumn{1}{c|}{\cellcolor[HTML]{CCCCCC}{\color[HTML]{08090C} \textbf{\begin{tabular}[c]{@{}c@{}}Downstream \\ Dataset\end{tabular}}}} & {\color[HTML]{08090C} \textbf{Methods}} & \multicolumn{1}{c}{\cellcolor[HTML]{CCCCCC}{\color[HTML]{08090C} \textbf{ACC$\uparrow$}}} & {\color[HTML]{08090C} \textbf{ASR$\downarrow$}} & \multicolumn{1}{l}{\cellcolor[HTML]{CCCCCC}{\color[HTML]{08090C} \textbf{ACC$\uparrow$}}} & \multicolumn{1}{l}{\cellcolor[HTML]{CCCCCC}{\color[HTML]{08090C} \textbf{ASR$\downarrow$}}} \\ \midrule
    \multicolumn{1}{c|}{} & \multicolumn{1}{c|}{} & \multicolumn{1}{c|}{} & No Defense & \multicolumn{1}{c|}{76.58} & 98.51 & \multicolumn{1}{c|}{76.79} & 100.00 \\ \cline{4-8} 
    \multicolumn{1}{c|}{} & \multicolumn{1}{c|}{} & \multicolumn{1}{c|}{\multirow{-2}{*}{STL-10}} & \cellcolor[HTML]{EFEFEF}Ours & \cellcolor[HTML]{EFEFEF}55.23 & \cellcolor[HTML]{EFEFEF}4.29 & \cellcolor[HTML]{EFEFEF}66.24 & \cellcolor[HTML]{EFEFEF}1.40 \\ \cline{3-8} 
    \multicolumn{1}{c|}{} & \multicolumn{1}{c|}{} & \multicolumn{1}{c|}{} & No Defense & 80.77 & 99.63 & 78.45 & 99.97 \\ \cline{4-8} 
    \multicolumn{1}{c|}{} & \multicolumn{1}{c|}{} & \multicolumn{1}{c|}{\multirow{-2}{*}{GTSRB}} & \cellcolor[HTML]{EFEFEF}Ours & \cellcolor[HTML]{EFEFEF}90.86 & \cellcolor[HTML]{EFEFEF}3.90 & \cellcolor[HTML]{EFEFEF}91.92 & \cellcolor[HTML]{EFEFEF}0.01 \\ \cline{3-8} 
    \multicolumn{1}{c|}{} & \multicolumn{1}{c|}{} & \multicolumn{1}{c|}{} & No Defense & 65.35 & 97.56 & 67.93 & 99.44 \\ \cline{4-8} 
    \multicolumn{1}{c|}{} & \multicolumn{1}{c|}{\multirow{-6}{*}{CIFAR-10}} & \multicolumn{1}{c|}{\multirow{-2}{*}{SVHN}} & \cellcolor[HTML]{EFEFEF}Ours & \cellcolor[HTML]{EFEFEF}85.93 & \cellcolor[HTML]{EFEFEF}3.76 & \cellcolor[HTML]{EFEFEF}92.52 & \cellcolor[HTML]{EFEFEF}0.65 \\ \cline{2-8} 
    \multicolumn{1}{c|}{} & \multicolumn{1}{c|}{} & \multicolumn{1}{c|}{} & No Defense & 70.57 & 98.93 & 69.66 & 99.96 \\ \cline{4-8} 
    \multicolumn{1}{c|}{} & \multicolumn{1}{c|}{} & \multicolumn{1}{c|}{\multirow{-2}{*}{CIFAR-10}} & \cellcolor[HTML]{EFEFEF}Ours & \cellcolor[HTML]{EFEFEF}60.65 & \cellcolor[HTML]{EFEFEF}5.22 & \cellcolor[HTML]{EFEFEF}62.90 & \cellcolor[HTML]{EFEFEF}6.80 \\ \cline{3-8} 
    \multicolumn{1}{c|}{} & \multicolumn{1}{c|}{} & \multicolumn{1}{c|}{} & No Defense & 70.83 & 98.99 & 66.67 & 99.83 \\ \cline{4-8} 
    \multicolumn{1}{c|}{} & \multicolumn{1}{c|}{} & \multicolumn{1}{c|}{\multirow{-2}{*}{GTSRB}} & \cellcolor[HTML]{EFEFEF}Ours & \cellcolor[HTML]{EFEFEF}87.08 & \cellcolor[HTML]{EFEFEF}4.93 & \cellcolor[HTML]{EFEFEF}90.43 & \cellcolor[HTML]{EFEFEF}0.76 \\ \cline{3-8} 
    \multicolumn{1}{c|}{} & \multicolumn{1}{c|}{} & \multicolumn{1}{c|}{} & No Defense & 64.89 & 98.98 & 63.55 & 99.57 \\ \cline{4-8} 
    \multicolumn{1}{c|}{\multirow{-12}{*}{\textbf{BadEncoder}}} & \multicolumn{1}{c|}{\multirow{-6}{*}{STL-10}} & \multicolumn{1}{c|}{\multirow{-2}{*}{SVHN}} & \cellcolor[HTML]{EFEFEF}Ours & \cellcolor[HTML]{EFEFEF}86.76 & \cellcolor[HTML]{EFEFEF}6.09 & \cellcolor[HTML]{EFEFEF}87.34 & \cellcolor[HTML]{EFEFEF}0.54 \\ \midrule
    \multicolumn{1}{c|}{} & \multicolumn{1}{c|}{} & \multicolumn{1}{c|}{} & No Defense & 71.85 & 97.72 & 72.39 & 99.94 \\ \cline{4-8} 
    \multicolumn{1}{c|}{} & \multicolumn{1}{c|}{} & \multicolumn{1}{c|}{\multirow{-2}{*}{STL-10}} & \cellcolor[HTML]{EFEFEF}Ours & \cellcolor[HTML]{EFEFEF}54.54 & \cellcolor[HTML]{EFEFEF}6.28 & \cellcolor[HTML]{EFEFEF}66.38 & \cellcolor[HTML]{EFEFEF}5.19 \\ \cline{3-8} 
    \multicolumn{1}{c|}{} & \multicolumn{1}{c|}{} & \multicolumn{1}{c|}{} & No Defense & 76.39 & 98.10 & 75.22 & 99.20 \\ \cline{4-8} 
    \multicolumn{1}{c|}{} & \multicolumn{1}{c|}{} & \multicolumn{1}{c|}{\multirow{-2}{*}{GTSRB}} & \cellcolor[HTML]{EFEFEF}Ours & \cellcolor[HTML]{EFEFEF}93.28 & \cellcolor[HTML]{EFEFEF}4.50 & \cellcolor[HTML]{EFEFEF}90.65 & \cellcolor[HTML]{EFEFEF}3.73 \\ \cline{3-8} 
    \multicolumn{1}{c|}{} & \multicolumn{1}{c|}{} & \multicolumn{1}{c|}{} & No Defense & 72.99 & 92.71 & 71.34 & 99.87 \\ \cline{4-8} 
    \multicolumn{1}{c|}{} & \multicolumn{1}{c|}{\multirow{-6}{*}{CIFAR-10}} & \multicolumn{1}{c|}{\multirow{-2}{*}{SVHN}} & \cellcolor[HTML]{EFEFEF}Ours & \cellcolor[HTML]{EFEFEF}87.27 & \cellcolor[HTML]{EFEFEF}6.47 & \cellcolor[HTML]{EFEFEF}89.57 & \cellcolor[HTML]{EFEFEF}3.60 \\ \cline{2-8} 
    \multicolumn{1}{c|}{} & \multicolumn{1}{c|}{} & \multicolumn{1}{c|}{} & No Defense & 71.14 & 80.49 & 71.21 & 99.66 \\ \cline{4-8} 
    \multicolumn{1}{c|}{} & \multicolumn{1}{c|}{} & \multicolumn{1}{c|}{\multirow{-2}{*}{CIFAR-10}} & \cellcolor[HTML]{EFEFEF}Ours & \cellcolor[HTML]{EFEFEF}63.93 & \cellcolor[HTML]{EFEFEF}1.61 & \cellcolor[HTML]{EFEFEF}63.07 & \cellcolor[HTML]{EFEFEF}5.70 \\ \cline{3-8} 
    \multicolumn{1}{c|}{} & \multicolumn{1}{c|}{} & \multicolumn{1}{c|}{} & No Defense & 65.11 & 85.03 & 64.90 & 99.18 \\ \cline{4-8} 
    \multicolumn{1}{c|}{} & \multicolumn{1}{c|}{} & \multicolumn{1}{c|}{\multirow{-2}{*}{GTSRB}} & \cellcolor[HTML]{EFEFEF}Ours & \cellcolor[HTML]{EFEFEF}84.51 & \cellcolor[HTML]{EFEFEF}3.97 & \cellcolor[HTML]{EFEFEF}85.82 & \cellcolor[HTML]{EFEFEF}0.86 \\ \cline{3-8} 
    \multicolumn{1}{c|}{} & \multicolumn{1}{c|}{} & \multicolumn{1}{c|}{} & No Defense & 58.43 & 96.28 & 58.35 & 99.66 \\ \cline{4-8} 
    \multicolumn{1}{c|}{\multirow{-12}{*}{\textbf{DRUPE}}} & \multicolumn{1}{c|}{\multirow{-6}{*}{STL-10}} & \multicolumn{1}{c|}{\multirow{-2}{*}{SVHN}} & \cellcolor[HTML]{EFEFEF}Ours & \cellcolor[HTML]{EFEFEF}87.37 & \cellcolor[HTML]{EFEFEF}5.58 & \cellcolor[HTML]{EFEFEF}83.91 & \cellcolor[HTML]{EFEFEF}0.37 \\ \midrule
    \multicolumn{1}{c|}{} & \multicolumn{1}{c|}{} & \multicolumn{1}{c|}{} & No Defense & 52.15 & 9.88 & 53.08 & 9.81 \\ \cline{4-8} 
    \multicolumn{1}{c|}{} & \multicolumn{1}{c|}{\multirow{-2}{*}{STL-10}} & \multicolumn{1}{c|}{\multirow{-2}{*}{STL-10}} & \cellcolor[HTML]{EFEFEF}Ours & \cellcolor[HTML]{EFEFEF}48.01 & \cellcolor[HTML]{EFEFEF}0.18 & \cellcolor[HTML]{EFEFEF}48.56 & \cellcolor[HTML]{EFEFEF}1.41 \\ \cline{2-8} 
    \multicolumn{1}{c|}{} & \multicolumn{1}{c|}{} & \multicolumn{1}{c|}{} & No Defense & 75.31 & 44.90 & 75.63 & 53.56 \\ \cline{4-8} 
    \multicolumn{1}{c|}{} & \multicolumn{1}{c|}{\multirow{-2}{*}{CIFAR-10}} & \multicolumn{1}{c|}{\multirow{-2}{*}{CIFAR-10}} & \cellcolor[HTML]{EFEFEF}Ours & \cellcolor[HTML]{EFEFEF}56.66 & \cellcolor[HTML]{EFEFEF}3.07 & \cellcolor[HTML]{EFEFEF}59.35 & \cellcolor[HTML]{EFEFEF}3.72 \\ \cline{2-8} 
    \multicolumn{1}{c|}{} & \multicolumn{1}{c|}{} & \multicolumn{1}{c|}{} & No Defense & 66.78 & 6.54 & 64.29 & 26.11 \\ \cline{4-8} 
    \multicolumn{1}{c|}{\multirow{-6}{*}{\textbf{CTRL}}} & \multicolumn{1}{c|}{\multirow{-2}{*}{GTSRB}} & \multicolumn{1}{c|}{\multirow{-2}{*}{GTSRB}} & \cellcolor[HTML]{EFEFEF}Ours & \cellcolor[HTML]{EFEFEF}82.42 & \cellcolor[HTML]{EFEFEF}0.87 & \cellcolor[HTML]{EFEFEF}88.11 & \cellcolor[HTML]{EFEFEF}1.91 \\ \midrule
    \multicolumn{1}{c|}{} & \multicolumn{1}{c|}{} & \multicolumn{1}{c|}{} & No Defense & 82.85 & 36.48 & 83.29 & 87.94 \\ \cline{4-8} 
    \multicolumn{1}{c|}{\multirow{-2}{*}{\textbf{SSLBackdoor}}} & \multicolumn{1}{c|}{\multirow{-2}{*}{ImageNet}} & \multicolumn{1}{c|}{\multirow{-2}{*}{ImageNet-10}} & \cellcolor[HTML]{EFEFEF}Ours & \cellcolor[HTML]{EFEFEF}72.35 & \cellcolor[HTML]{EFEFEF}0.42 & \cellcolor[HTML]{EFEFEF}81.35 & \cellcolor[HTML]{EFEFEF}1.76 \\ \midrule
    \multicolumn{1}{c|}{} & \multicolumn{1}{c|}{} & \multicolumn{1}{c|}{} & No Defense & 82.35 & 58.46 & 82.47 & 92.12 \\ \cline{4-8} 
    \multicolumn{1}{c|}{\multirow{-2}{*}{\textbf{CorruptEncoder}}} & \multicolumn{1}{c|}{\multirow{-2}{*}{ImageNet}} & \multicolumn{1}{c|}{\multirow{-2}{*}{ImageNet-10}} & \cellcolor[HTML]{EFEFEF}Ours & \cellcolor[HTML]{EFEFEF}72.82 & \cellcolor[HTML]{EFEFEF}1.03 & \cellcolor[HTML]{EFEFEF}81.47 & \cellcolor[HTML]{EFEFEF}4.79 \\ 
     \bottomrule
    \end{tabular}
    }
    \label{tab:threat1andthreatthree}
    \vspace{-2mm}
    \end{table}

\subsection{RQ4: Effectiveness of T-Core Bootstrapping}
As we previously discussed, in the defense context of transfer learning, the defender may not know what kind of backdoor risks it is facing in general, so a successful defense should be able to deal with all kinds of backdoor threats.
Thus, we evaluate the end-to-end procedure of our proposed T-Core Bootstrapping framework in defending all four scenarios: the encoder poisoning alone of \threatone, the dataset poisoning alone of \threattwo, the adaptive poisoning with both the encoder and dataset with the same backdoor trigger of \threatthree, as well as the independent encoder poisoning and dataset poisoning with the different backdoor triggers of \threatone~and \threattwo.

Overall, T-Core effectively defends against all considered backdoor threats, as shown in \cref{tab:threat2-end2end,tab:threat1andthreatthree,tab:threat4-end2end}. 
Specifically, the attack success rates for \threatone, \threattwo, and \threatthree~are all below 10\%.
Additionally, T-Core improves accuracy (ACC) in most cases, due to the optimization of the original untrusted channels. 
However, ACC often decreases for STL-10 due to its limited training images.
T-Core halts training at 4,500 samples (90\%), resulting in incomplete training.
For cases where two independent attackers of \threatone~and \threattwo~both exist, the results in \cref{tab:threat4-end2end} indicate that the attack potency of encoder poisoning is consistently greater than that of downstream poisoning.
For downstream poisoning, Adap-Patch, Adap-Blend, and WaNet also produce limited attack potency as they inject different trigger patterns for different poisoned samples.
Nevertheless, T-Core can ensure low ASRs of both encoder poisoning and dataset poisoning, regardless of the attack's original strength.

\subsection{RQ5: Scalability of T-Core Bootstrapping}

\begin{table*}[t]
\centering
    \caption{
        Performance of seed-sifting module TIS and the entire framework of T-Core against \textbf{adaptive attack} on CIFAR-10. 
        We test out the perturbation budget as 4, 8, 12, and 16  of the infinite norm, respectively.
        `Poi Num' denotes the number of poisons are deemed as clean seed data by TIS. }
    \resizebox{0.98\textwidth}{!}{\large
    \begin{tabular}{c|ccc|ccc|ccc|ccc}
    \toprule
    \multirow{2}{*}{\begin{tabular}[c]{@{}c@{}}Perturbation\\ Range\end{tabular}} & \multicolumn{3}{c|}{4} & \multicolumn{3}{c|}{8} & \multicolumn{3}{c|}{12} & \multicolumn{3}{c}{16} \\ \cline{2-13} 
     & ACC & ASR & Poi Num & ACC & ASR & Poi Num & ACC & ASR & Poi Num & ACC & ASR & Poi Num \\ \midrule
    No Defense & 83.37 & 83.71 & - & 83.65 & 96.08 & - & 83.41 & 96.51 & - & 84.04 & 96.75 & - \\ \hline
    Ours & 81.79 & 5.65 & 0 & 82.97 & 13.42 & 1 & 81.67 & 21.49 & 4 & 81.42 & 92.52 & 38 \\ \bottomrule
    \end{tabular}
    }
\label{tab:adaptive_attack}
\vspace{-2mm}
\end{table*}

{\color{black}
\subsubsection{Resilience against Adaptive Attack}
T-Core relies on the initial topological invariance sifting (TIS) of high-credible samples, which are carried out based on the assumptions of both majority rule and consistency rule. 
To launch an effective adaptive backdoor poisoning, we exploit the seed data sifted by TIS by conducting a layerwise adversarial attack to construct backdoor triggers.
Specifically, we generate a universal adversarial perturbation (UAP) $\delta$ on input samples that minimizes the activation distance between perturbed inputs and target class samples across multiple layers as follows:
$$
\min_{\delta} \frac{1}{|\downdata|\times L} \sum_{x_1 \in \downdata} \sum_{l=N-L-1}^{N-1} \|h^l(x_1 + \delta)- \frac{1}{|\mathcal{S}_t|} \sum_{x_2 \in \mathcal{S}_t} h^l(x_2)\|,
$$
where $\mathcal{S}_t$ is the seed data of the target class (sifted from clean data $\downdata_t$ by TIS).
TIS is conducted on the last $L$ layers before the final layer, \ie, from $(N-L-1)$-th layer to $(N-1)$-th layer.
Here we use the mean activation of target class samples screened by TIS as the optimization in each layer.
The optimized adversarial perturbation is then used as the trigger for backdoor injection.
Such an adaptive adversary is provided with total knowledge of our TIS, as well as access to the pre-trained encode and the downstream dataset.

The experiments reveal a clear trade-off between the adversarial budget and attack effectiveness. 
As demonstrated in \cref{tab:adaptive_attack}, even under this adaptive attack, a significant adversarial perturbation budget is required to completely undermine our defense, specifically, an $l_{\infty}$ norm of 16 (where each pixel value is expected to change by 16).
In contrast, TIS remains effective under smaller adversarial budgets of $l_{\infty}$ norm from 4 to 12, only minimum poisoned samples (0 to 4) are mistakenly selected as clean seed data. 
The ASR rises as the adversarial budget increases, though, T-Core does demonstrate resilience against moderate perturbations. 

\subsubsection{Sensitivity to Hyperparameters Variations}
}
\begin{table}[t]
    \centering
    \caption{Ablation on the hyper-parameters $\gamma_1$, $\gamma_2$ of bootstrapping learning on CIFAR-10. 
    $\gamma_1$ is the selection rate from \textit{each class},  $\gamma_2$ is the selection rate from \textit{entire dataset}.
    }
    \resizebox{0.45\textwidth}{!}{\large
    \begin{tabular}{ccccccccc}
        \toprule
        \multicolumn{9}{c}{\cellcolor[HTML]{FFFFFF}SIG} \\ \hline
        \multicolumn{1}{c|}{\cellcolor[HTML]{FFFFFF}} & \multicolumn{4}{c|}{\cellcolor[HTML]{FFFFFF}\textbf{ACC$\uparrow$}} & \multicolumn{4}{c}{\cellcolor[HTML]{FFFFFF}\textbf{ASR$\downarrow$}} \\ \cline{2-9} 
        \multicolumn{1}{c|}{\multirow{-2}{*}{\cellcolor[HTML]{FFFFFF}\textbf{No Defense}}} & \multicolumn{4}{c|}{\cellcolor[HTML]{FFFFFF}\textbf{84.72}} & \multicolumn{4}{c}{\cellcolor[HTML]{FFFFFF}\textbf{89.1}} \\ \midrule
        \multicolumn{1}{c|}{\textbf{$\gamma_2$(\%)}} & \multicolumn{2}{c|}{\textbf{2}} & \multicolumn{2}{c|}{\textbf{5}} & \multicolumn{2}{c|}{\textbf{7}} & \multicolumn{2}{c}{\textbf{10}} \\ \hline
        \multicolumn{1}{c|}{\textbf{$\gamma_1$(\%)}} & \multicolumn{1}{c|}{ACC$\uparrow$} & \multicolumn{1}{c|}{ASR$\downarrow$} & \multicolumn{1}{c|}{ACC$\uparrow$} & \multicolumn{1}{c|}{ASR$\downarrow$} & \multicolumn{1}{c|}{ACC$\uparrow$} & \multicolumn{1}{c|}{ASR$\downarrow$} & \multicolumn{1}{c|}{ACC$\uparrow$} & ASR$\downarrow$ \\ \hline
        \multicolumn{1}{c|}{\textbf{1}} & \multicolumn{1}{c|}{88.24} & \multicolumn{1}{c|}{2.24} & \multicolumn{1}{c|}{88.36} & \multicolumn{1}{c|}{2.34} & \multicolumn{1}{c|}{87.79} & \multicolumn{1}{c|}{4.53} & \multicolumn{1}{c|}{87.22} & 7.92 \\ \hline
        \multicolumn{1}{c|}{\textbf{2}} & \multicolumn{1}{c|}{89.09} & \multicolumn{1}{c|}{2.76} & \multicolumn{1}{c|}{87.31} & \multicolumn{1}{c|}{2.54} & \multicolumn{1}{c|}{86.91} & \multicolumn{1}{c|}{4.27} & \multicolumn{1}{c|}{87.13} & 8.44 \\ \hline
        \multicolumn{1}{c|}{\textbf{5}} & \multicolumn{1}{c|}{87.98} & \multicolumn{1}{c|}{4.49} & \multicolumn{1}{c|}{86.16} & \multicolumn{1}{c|}{9.59} & \multicolumn{1}{c|}{83.55} & \multicolumn{1}{c|}{21.73} & \multicolumn{1}{c|}{81.84} & 40.57 \\ \midrule
        \multicolumn{9}{c}{\cellcolor[HTML]{FFFFFF}Blended} \\ \hline
        \multicolumn{1}{c|}{\cellcolor[HTML]{FFFFFF}} & \multicolumn{4}{c|}{\cellcolor[HTML]{FFFFFF}\textbf{ACC$\uparrow$}} & \multicolumn{4}{c}{\cellcolor[HTML]{FFFFFF}\textbf{ASR$\downarrow$}} \\ \cline{2-9} 
        \multicolumn{1}{c|}{\multirow{-2}{*}{\cellcolor[HTML]{FFFFFF}\textbf{No Defense}}} & \multicolumn{4}{c|}{\cellcolor[HTML]{FFFFFF}\textbf{84.84}} & \multicolumn{4}{c}{\cellcolor[HTML]{FFFFFF}\textbf{89.12}} \\ \midrule
        \multicolumn{1}{c|}{\textbf{$\gamma_2$(\%)}} & \multicolumn{2}{c|}{\textbf{2}} & \multicolumn{2}{c|}{\textbf{5}} & \multicolumn{2}{c|}{\textbf{7}} & \multicolumn{2}{c}{\textbf{10}} \\ \hline
        \multicolumn{1}{c|}{\textbf{$\gamma_1$(\%)}} & \multicolumn{1}{c|}{ACC$\uparrow$} & \multicolumn{1}{c|}{ASR$\downarrow$} & \multicolumn{1}{c|}{ACC$\uparrow$} & \multicolumn{1}{c|}{ASR$\downarrow$} & \multicolumn{1}{c|}{ACC$\uparrow$} & \multicolumn{1}{c|}{ASR$\downarrow$} & \multicolumn{1}{c|}{ACC$\uparrow$} & ASR$\downarrow$ \\ \hline
        \multicolumn{1}{c|}{\textbf{1}} & \multicolumn{1}{c|}{88.64} & \multicolumn{1}{c|}{4.48} & \multicolumn{1}{c|}{87.38} & \multicolumn{1}{c|}{4.93} & \multicolumn{1}{c|}{88.01} & \multicolumn{1}{c|}{6.37} & \multicolumn{1}{c|}{86.93} & 13.75 \\ \hline
        \multicolumn{1}{c|}{\textbf{2}} & \multicolumn{1}{c|}{88.09} & \multicolumn{1}{c|}{4.43} & \multicolumn{1}{c|}{87.35} & \multicolumn{1}{c|}{5.9} & \multicolumn{1}{c|}{87.14} & \multicolumn{1}{c|}{7.37} & \multicolumn{1}{c|}{86.26} & 16.07 \\ \hline
        \multicolumn{1}{c|}{\textbf{5}} & \multicolumn{1}{c|}{86.63} & \multicolumn{1}{c|}{8.46} & \multicolumn{1}{c|}{85.31} & \multicolumn{1}{c|}{12.74} & \multicolumn{1}{c|}{83.21} & \multicolumn{1}{c|}{26.07} & \multicolumn{1}{c|}{82.55} & 44.15 \\ \bottomrule
    \end{tabular}
    }
    \label{tab:gamma1_gamma2}
    \vspace{-2mm}
\end{table}

\begin{table}[t]
    \centering
    \caption{Ablation on bootstrapping rate $\rho$ in Bootstrapping Learning, \textit{i.e.}, when we choose to halt the bootstrapping learning, on CIFAR-10, 
    `Poi Num' denotes the final number of poison samples that are included.
    }
    \resizebox{0.45\textwidth}{!}{\large
    \begin{tabular}{c|ccc|ccc}
        \toprule
        \rowcolor[HTML]{FFFFFF} 
        \cellcolor[HTML]{FFFFFF} & \multicolumn{3}{c|}{\cellcolor[HTML]{FFFFFF}SIG} & \multicolumn{3}{c}{\cellcolor[HTML]{FFFFFF}Blended} \\ \cline{2-7} 
        \rowcolor[HTML]{FFFFFF} 
        \multirow{-2}{*}{\cellcolor[HTML]{FFFFFF}\begin{tabular}[c]{@{}c@{}}Bootstrapping\\ Rate(\%)\end{tabular}} & \multicolumn{1}{c|}{\cellcolor[HTML]{FFFFFF}ACC$\uparrow$} & \multicolumn{1}{c|}{\cellcolor[HTML]{FFFFFF}ASR$\downarrow$} & Poi Num & \multicolumn{1}{c|}{\cellcolor[HTML]{FFFFFF}ACC$\uparrow$} & \multicolumn{1}{c|}{\cellcolor[HTML]{FFFFFF}ASR$\downarrow$} & Poi Num \\ \hline
        \rowcolor[HTML]{FFFFFF} 
        No Defense & \multicolumn{1}{c|}{\cellcolor[HTML]{FFFFFF}84.72} & \multicolumn{1}{c|}{\cellcolor[HTML]{FFFFFF}89.10} & - & \multicolumn{1}{c|}{\cellcolor[HTML]{FFFFFF}84.84} & \multicolumn{1}{c|}{\cellcolor[HTML]{FFFFFF}89.12} & - \\ \midrule
        97 & \multicolumn{1}{c|}{85.13} & \multicolumn{1}{c|}{12.48} & 65 & \multicolumn{1}{c|}{85.54} & \multicolumn{1}{c|}{71.54} & 568 \\ \hline
        95 & \multicolumn{1}{c|}{87.62} & \multicolumn{1}{c|}{3.11} & 9 & \multicolumn{1}{c|}{86.62} & \multicolumn{1}{c|}{79.38} & 88 \\ \hline
        90 & \multicolumn{1}{c|}{87.31} & \multicolumn{1}{c|}{2.54} & 0 & \multicolumn{1}{c|}{87.35} & \multicolumn{1}{c|}{5.90} & 13 \\ \hline
        85 & \multicolumn{1}{c|}{81.09} & \multicolumn{1}{c|}{1.87} & 0 & \multicolumn{1}{c|}{82.26} & \multicolumn{1}{c|}{3.71} & 5 \\ \hline
        80 & \multicolumn{1}{c|}{79.41} & \multicolumn{1}{c|}{2.06} & 0 & \multicolumn{1}{c|}{80.63} & \multicolumn{1}{c|}{2.31} & 0 \\ \hline
        70 & \multicolumn{1}{c|}{76.22} & \multicolumn{1}{c|}{1.63} & 0 & \multicolumn{1}{c|}{77.10} & \multicolumn{1}{c|}{1.56} & 0 \\ \bottomrule
    \end{tabular}
    }
    \label{tab:rho}
    \vspace{-2mm}
\end{table}

{\color{black}
We conducted ablations on the selection rate for each class $\gamma_1$, the selection rate for the entire dataset $\gamma_2$ (\cref{tab:gamma1_gamma2}) and the stopping criterion $\rho$  for Clean Bootstrapping (\cref{tab:rho}) using two standard dataset poisoning attacks, SIG and Blended.
Results in \cref{tab:gamma1_gamma2} show that selecting from each class requires more caution than selecting from the entire dataset.
A smaller selection rate (more cautious approach) leads to a lower ASR.
On the other hand, results in \cref{tab:rho} demonstrate that for the Blended attack, when $\rho$ exceeds $85 \%$, a few poison samples are already included for training, increasing the ASR while the ACC is not high enough. 
For the SIG attack, a $\rho$ of $90 \%$ or lower effectively eliminates poison samples, achieving a low ASR while maintaining a high ACC. 
A $\rho$ above $90 \%$ has made the ASR significantly high. 
Thus, the choice of $\rho$ should be carefully tuned, considering different dataset poisoning.

\subsubsection{Adaptivity to Vision Transformer}

}
\begin{table*}[ht]
\caption{Performance of T-Core on Vision Transformer against different types of backdoor threats of \threatone~(Dataset Poisoning), \threattwo~(Encoder Poisoning), and \threatthree~(Adaptive Poisoning). 
}
\centering
\renewcommand{\arraystretch}{1.2}
\setlength{\tabcolsep}{4pt} 
\resizebox{\textwidth}{!}{%
\begin{tabular}{@{}lcccccc cccccc cccccc@{}}
\toprule
\multirow{2}{*}{Threat Type} & 
\multicolumn{6}{c}{\textbf{\threattwo} } & 
\multicolumn{4}{c}{\textbf{\threatone}  } & 
\multicolumn{4}{c}{\textbf{\threatthree} } \\ 
\cmidrule(lr){2-7} \cmidrule(lr){8-11} \cmidrule(l){12-15} 
& \multicolumn{2}{c}{BadNets} & 
\multicolumn{2}{c}{Blended} & 
\multicolumn{2}{c}{SIG} & 
\multicolumn{2}{c}{BadEncoder} & 
\multicolumn{2}{c}{DRUPE} & 
\multicolumn{2}{c}{BadEncoder} & 
\multicolumn{2}{c}{DRUPE} \\ 
\cmidrule(lr){2-3} \cmidrule(lr){4-5} \cmidrule(lr){6-7} 
\cmidrule(lr){8-9} \cmidrule(lr){10-11} 
\cmidrule(l){12-13} \cmidrule(l){14-15} \cmidrule(l){16-17} 
\cmidrule(l){18-19} 
& ACC$\uparrow$ & ASR$\downarrow$ & 
ACC$\uparrow$ & ASR$\downarrow$ & 
ACC$\uparrow$ & ASR$\downarrow$ & 
ACC$\uparrow$ & ASR$\downarrow$ & 
ACC$\uparrow$ & ASR$\downarrow$ & 
ACC$\uparrow$ & ASR$\downarrow$ & 
ACC$\uparrow$ & ASR$\downarrow$ \\ 
\midrule
No Defense & 72.17 & 87.57 & 72.09 & 82.53 & 72.03 & 62.99 & 60.33 & 98.79 & 66.22 & 99.83 & 58.02 & 99.99 & 66.20 & 99.99 \\ 
Ours & 70.94 & 4.35 & 70.91 & 3.55 & 70.94 & 3.15 & 56.46 & 0.26 & 64.95 & 4.50 & 55.55 & 2.50 & 65.25 & 5.57 \\ 
\bottomrule
\end{tabular}%
}
\label{tab:vit_performance}
\vspace{-3mm}
\end{table*}


{\color{black}
The concept of channels in Encoder Channel Filtering, originally designed for Convolutional Neural Networks (CNNs), does not directly translate to Vision Transformers (ViTs). However, we can adapt it by reinterpreting each linear layer in the transformer as having multiple channels. For example, a linear layer with dimensions $1536 \times 512$ can be viewed as having 1536 channels, where each channel generates a 512-dimensional feature vector (see the code implementation in \cref{fig:masked_linear}). This enables the application of a learnable mask to the channels of the linear layer, mirroring the approach used in CNNs.
Unlike CNNs, which often rely on batch normalization, ViTs utilize layer normalization. Thus, we leverage layer normalization for the unlearning process. 
Results in \cref{tab:vit_performance} demonstrate that T-Core effectively defends against Threat-1, Threat-2, and Threat-3, achieving low ASRs while maintaining accuracy close to the original undefended models.

\subsubsection{Computational Efficiency Analysis}

}

{\color{black}
We present a detailed comparison of the time expenses of various seed-sifting methods in \cref{tab:seed_sifting_time}. Our Topological Invariance Sifting (TIS) module significantly outperforms state-of-the-art data-sifting approaches, such as MetaSift and SPECTRE, in both efficiency and effectiveness, requiring substantially less computation time. In contrast, faster alternatives like IBD-PSC, SCALE-UP, and Spectral exhibit markedly inferior performance in data sifting.
In \cref{tab:tcore_time_memory}, we report the time expenses and memory usage for each module of T-Core. The total runtime is 1102.3 seconds (under 20 minutes), which we consider highly practical for a defense mechanism that delivers a clean, backdoor-free classifier. Additionally, memory efficiency is achieved by targeting a subset of parameters, with peak GPU memory usage remaining below 5 GB. These demonstrate the scalability of T-Core, making it a viable solution for real-world applications where computational resources are limited.

\begin{table}[t]
    \centering
    \caption{\textbf{Time expenses} (seconds) \textbf{in comparison} of different seed-sifting methods, on the CIFAR-10 dataset at varying dataset sizes per class. 
    According to \cref{tab:clean_data_sifting}, the faster IBD-PSC, SCALE-UP, and Spectral generally perform poorly. 
    }
    \resizebox{0.48\textwidth}{!}{
    \begin{tabular}{c|c|c|c|c|c}
        \toprule
        \rowcolor[HTML]{FFFFFF}
        \textbf{\begin{tabular}[c]{@{}c@{}}Data Size\\ per Class\end{tabular}} & \textbf{100} & \textbf{500} & \textbf{1000} & \textbf{2500} & \textbf{5000} \\ \midrule
        \textbf{IBD-PSC} & 3.83 & 14.94 & 26.81 & 64.81 & 139.69 \\ \hline
        \textbf{SCALE-UP} & 0.69 & 3.02 & 6.00 & 14.81 & 29.51 \\ \hline
        \textbf{Spectral} & \textbf{0.32} & \textbf{1.85} & \textbf{3.07} & \textbf{7.01} & \textbf{14.13} \\ \hline
        \textbf{SPECTRE} & 14.53 & 74.23 & 117.95 & 231.71 & 456.57 \\ \hline
        \textbf{META-SIFT} & 47.99 & 83.68 & 128.34 & 267.09 & 489.26 \\ \hline
        \textbf{Ours} & 7.01 & 26.05 & 46.68 & 129.29 & 294.38 \\ \bottomrule
    \end{tabular}
    }
    \label{tab:seed_sifting_time}
    \vspace{-2mm}
\end{table}

\begin{table}[t]
    \centering
    \caption{\textbf{Time expenses and memory usage} of T-Core. 
    We report the total time and memory expenses for T-Core, along with the breakdown for each module: Topological Invariance Sifting (TIS), Seed Expansion (SE), Encoder Channel Filtering (ECF), and Bootstrapping Learning (BL). Experiments are conducted on the CIFAR-10 dataset (50,000 samples) using a ResNet18 encoder with a batch size of 128. 
    }
    \resizebox{0.48\textwidth}{!}{
    \begin{tabular}{c|c|c|cc|c|c}
        \toprule
        Stages & TIS & SE & \multicolumn{2}{c|}{ECF} & BL & Total Expenses \\ \midrule
        Time (s) & 294.4 & 60.1 & \multicolumn{2}{c|}{105.6} & 642.2 & 1102.3 \\ \midrule
        VRAM (MB) & 5312 & 3338 & \multicolumn{2}{c|}{3484} & 4794 & 5312 \\ \bottomrule
    \end{tabular}
    }
    \label{tab:tcore_time_memory}
    \vspace{-2mm}
\end{table}

\section{Limitation and Future Work}

Our T-Core framework is designed to defend against unknown backdoor threats in blind transfer learning scenarios, where neither the attack pattern nor the integrity of the pre-trained model or data is known. By adopting a proactive defense strategy—identifying and amplifying clean elements—we mitigate potential backdoors without prior knowledge of the threat.

However, this approach relies on sifting a number of clean samples to bootstrap a reliable trust core. 
A key limitation arises when the training data is scarce, making it difficult to isolate enough clean elements for robust initialization. 
For instance, STL-10 (with only 5,000 samples) exhibits a more noticeable accuracy drop compared to larger datasets.
Future work will explore better trade-offs between security and utility in low-data regimes.

Additionally, our current focus is on vision-based backdoors in scenarios where users fine-tune pre-trained encoders on image datasets. We have not yet investigated language-domain backdoors or multimodal zero-shot settings (e.g., CLIP-like models), where threats could emerge from both image and text encoders. Extending T-Core to these domains remains an open challenge for future work.
 }

\section{Conclusion}

In this study, we address the critical challenge of securing transfer learning models from backdoor attacks, where the security risk is amplified by the employment of untrusted pre-trained encoders and potentially poisoned datasets. Our exhaustive analysis shows that traditional defenses, which often depend on reactive approaches, 
are inadequate to the unknown threats and diverse attack vectors within transfer learning.
To overcome this limitation, we thus advocate for a proactive mindset focused on identifying and expanding trustworthy elements and introduce the Trusted Core (T-Core) Bootstrapping framework.
T-Core effectively neutralizes backdoor risks by initializing with meticulously vetted, high-credibility samples and progressively expanding the trusted dataset and model elements. 
Our comprehensive empirical evaluation, spanning a wide array of encoder and dataset poisoning, demonstrates the superiority of the T-Core. 
On the big picture, our work underscores the importance of adopting a proactive mindset in developing backdoor defenses, particularly in transfer learning, where an expanded attack surface and unknown threats heighten the security risk.


{\footnotesize
\bibliographystyle{ieeetr}  
\bibliography{usenix}
}

\appendices \label{Appendix}

\section{Implementation Details of T-Core}


In \cref{sec:T-Core}, we present the overall algorithmic details of our Trusted Core Bootstrapping framework, here we illustrate the implementation details of T-Core.

\noindent\textbf{Details of Topological Invariance Sifting.}
The algorithm  \cref{algo:sifting} in \cref{sec:seedsifting} extracts high-credibility seed data from a poisoned dataset by utilizing topological invariance across various layers of a well-trained poisoned model.
For clustering, we employ activations from the output of the encoder and the first two layers of the classifier, and implement density-based clustering using DBSCAN \cite{ester1996density}.
We set the number of nearest neighbors \( m \) to 50 and employ Euclidean distance for neighbor determination. The seed selection ratio \( \alpha \) is set to 1\%.

\noindent\textbf{Details of Seed Expansion}
\cref{algo:seedexpansion} in \cref{sec:data_expansion} expand the seed data obtained from \cref{algo:sifting}.
We set the select ratio $r_\textrm{expand}$ for expansion as 5\%, and the expansion ratio \( |\subdata| / |\downdata|\) as 10\%.

\noindent\textbf{Details of Encoder Channel Filtering}
The algorithm in \cref{algo:pruning} from \cref{sec:encoder_filtering} filters trusted and untrusted channels within the pre-trained encoder. The unlearning process stops once the training accuracy drops below 20\%. The recovery process is conducted using the Adam optimizer with a learning rate of 0.01. Training is performed for 120 epochs. The filtering threshold is set to cut off 10\% of channels as untrusted.

\noindent\textbf{Details of Clean Bootstrapping Training}
The clean bootstrapping described in \cref{algo:bootstrapping_learning} iteratively expands the clean subset and refines the model parameters. We set both \( Iter1 \) and \( Iter2 \) to 10, with both \( \gamma_1\% \) and \( \gamma_2\% \) set to 2\%. Additionally, \( \gamma_3\% \) is set to 5\%. This process continues until the clean subset reaches \( \rho = 90\% \) of the entire dataset.

\section{Experimental Setup Explanation} \label{sec:implementation_details}

\subsection{Datasets Explanation} \label{sec:dataset_explanation}
We detail the datasets utilized and how we process them as pre-training or downstream datasets as follows:
\begin{itemize}[leftmargin=6pt]

\item {\bf STL-10~\cite{coates2011analysis}:} 
This dataset comprises 5,000 labeled training images, 8,000 labeled testing images, and an additional 100,000 unlabeled images, each with dimensions of 96 $\times$ 96 $\times$ 3, and all parts of the image set can be evenly divided into ten classes, where each image belongs to a class.
We use the unlabeled images as a pre-training dataset and the labeled part for the downstream training and testing.

\item {\bf CIFAR-10~\cite{krizhevsky2009learning}:} This dataset is also a balanced dataset containing 50,000 labeled training images and 10,000 testing images, both of which are evenly divided into 10 classes. Each image has a size of 32$\times$32$\times$3. We use this dataset for the downstream dataset and remove the labels of the training images when pre-training the encoder.

\item {\bf GTSRB~\cite{stallkamp2012man}:} This dataset contains 43 classes of traffic signs, split into 39,209 training images and 12,630 test images. Each image has a size of 32$\times$32$\times$3. Furthermore, this dataset has an uneven distribution of the number of images belonging to each class, ranging from 210 to 2250, which presents a real-world challenge, especially when under poisoning. Thus, we use it for the downstream dataset to evaluate the effectiveness of defenses.

\item {\bf SVHN~\cite{netzer2011reading}:} In this dataset, every image depicts a digit from the house numbers collected from Google Street View with the size of 32$\times$32$\times$3. Additionally, the images fall into one of the ten possible digit categories. 
There are 73,257 training images and 26,032 testing images in this dataset. Moreover, extraneous digits are added at the edges of the primary digit in focus, making it even more challenging to separate poisoned and clean samples under SVNH. Thus, we also use it as a downstream dataset to test the robustness of defenses.

\item {\bf ImageNet~\cite{deng2009imagenet}:} This dataset is built for large-scale object classiﬁcation and contains 1,281,167 training samples and 50,000 testing samples in 1000 classes. The input size is 224$\times$224$\times$3.
We use a 100-class subset for the pre-training dataset and a 10-class subset for the downstream dataset.
\end{itemize}

\subsection{Pre-trained Encoder Explanation}\label{subsec:EncoderExplanation}

In this section, we provide the detailed implementation of each encoder poisoning attack, specifically how we obtain both clean and poisoned encoders, and how these encoders are utilized in the context of transfer learning.

\vspace{1mm}
\noindent \textbf{BadEncoder} and \textbf{DRUPE}: Both of these methods inject backdoors by fine-tuning a clean pre-trained encoder, assigning a target class to a specific concept within a downstream dataset. Our experiments utilize STL-10 as the pre-training dataset while targeting backdoors for CIFAR-10, GTSRB, and SVHN. Conversely, we also use CIFAR-10 as the pre-training dataset, with backdoors aimed at STL-10, GTSRB, and SVHN.

\vspace{1mm}
\noindent \textbf{CTRL}: In contrast to the backdoor injection methods used by BadEncoder and DRUPE, CTRL contaminates the unlabeled images in the pre-training dataset, specifically targeting a concept within the label set of that dataset. As a result, its backdoor functionality is limited to downstream datasets that share the same label set or a subset of it. Accordingly, we adhere to the methodology outlined in \cite{CTRL} and utilize CIFAR-10, STL-10, and GTSRB as both pre-training and downstream datasets.

\vspace{1mm}
\noindent \textbf{SSLBackdoor} and \textbf{CorruptEncoder}: These methods also employ the same 100-class subset of ImageNet as the pre-training dataset. They inject backdoors in a manner similar to the CTRL method by directly poisoning the pre-training data. For our experiments, we select a 10-class subset of ImageNet as the downstream dataset, ensuring that the target classes for both SSLBackdoor and CorruptEncoder are included in this subset. 

Additionally, for the clean encoders of STL-10 and CIFAR-10, we follow the approach in \cite{BadEncoder} to train a ResNet18 using SimCLR, leveraging the publicly available code from SimCLR\footnote{\href{https://github.com/leftthomas/SimCLR}{https://github.com/leftthomas/SimCLR}}. The clean ImageNet encoder is sourced from \cite{SSL-backdoor}\footnote{\href{https://github.com/UMBCvision/SSL-Backdoor}{https://github.com/UMBCvision/SSL-Backdoor}}. The backdoor encoders are constructed through the respective backdoor poisoning methods. For BadEncoder\footnote{\href{https://github.com/jinyuan-jia/BadEncoder}{https://github.com/jinyuan-jia/BadEncoder}} and DRUPE\footnote{\href{https://github.com/Gwinhen/DRUPE}{https://github.com/Gwinhen/DRUPE}}, we directly utilize their open-source checkpoints trained on STL-10 and CIFAR-10. For the ImageNet backdoor encoder, we employ the checkpoints provided by \cite{zhang2022corruptencoder}\footnote{\href{https://github.com/jzhang538/CorruptEncoder}{https://github.com/jzhang538/CorruptEncoder}} for both SSL-Backdoor and CorruptEncoder.

\subsection{Dataset Poisoning Explanation} \label{subsec:DatasetPoisoningExplanation}
For dataset backdoor poisoning,  we use the  backdoor-toolbox\footnote{\href{https://github.com/vtu81/backdoor-toolbox}  {https://github.com/vtu81/backdoor-toolbox}} benchmark to implement all the attacks on downstream datasets.
Note that this benchmark does not provide experimental setups for certain attack methods on some datasets. For instance, TaCT attack on ImageNet is originally not supported. We conducted experiments by adapting the settings from other datasets.

\vspace{1mm}
\noindent
\textbf{Trigger and Target Class:}
In the context of \threattwo, we employed distinct triggers and target classes to avoid conflicts between Encoder Poisoning and Dataset Poisoning. Specifically, when both \threatone~ and \threattwo~ are present, we maintained the same approach used for \threattwo.
In \threatthree, we utilized the same triggers and target classes as those applied in the pre-trained encoder.
For \threattwo, aside from TaCT, the target classes are as follows: ``Car'' for STL-10, ``Bird'' for CIFAR-10, ``Speed limit (50km/h)'' for GTSRB, ``2'' for SVHN, and ``n01855672 (Goose)'' for ImageNet-10. 
The specific target classes for TaCT and \threatthree~ across various datasets are detailed in \cref{tab:target_class}.
\\
\noindent
\textbf{Poisoning and Cover Ratio.}
By default, the poisoning rate was set to 20\% of the target category. For specific attacks like TaCT, WaNet, Adap-Blend, and Adap-Patch, the cover rate was set to 1/4 of the poisoning rate.
\begin{table}[ht]
    \caption{Target Class of Various  Dataset Poisoning Methods on Downstream Dataset}
    \adjustbox{center}{
        \resizebox{0.49\textwidth}{!}{\Huge
        \setlength{\tabcolsep}{0.6pt}
        \begin{tabular}{c|ccc|ccc}
        \toprule
        \rowcolor[HTML]{CCCCCC} 
        \textbf{Threat Type} & \multicolumn{3}{c|}{\cellcolor[HTML]{CCCCCC}\textbf{Threat 2}} & \multicolumn{3}{c}{\cellcolor[HTML]{CCCCCC}\textbf{Threat 3}} \\ \midrule
        \rowcolor[HTML]{D9D9D9} 
        \cellcolor[HTML]{D9D9D9} & \multicolumn{3}{c|}{\cellcolor[HTML]{D9D9D9}TaCT} & \multicolumn{1}{c|}{\cellcolor[HTML]{D9D9D9}BadEncoder} & \multicolumn{1}{c|}{\cellcolor[HTML]{D9D9D9}DRUPE} & CTRL \\ \cline{2-7} 
        \rowcolor[HTML]{D9D9D9} 
        \multirow{-2}{*}{\cellcolor[HTML]{D9D9D9}Dataset} & \multicolumn{1}{c|}{\cellcolor[HTML]{D9D9D9}Target Class} & \multicolumn{1}{c|}{\cellcolor[HTML]{D9D9D9}Source Class} & Cover Class & \multicolumn{1}{c|}{\cellcolor[HTML]{D9D9D9}Target Class} & \multicolumn{1}{c|}{\cellcolor[HTML]{D9D9D9}Target Class} & Target Class \\ \midrule
        STL-10 & \multicolumn{1}{c|}{Car} & \multicolumn{1}{c|}{Truck} & Dog Horse Monkey & \multicolumn{1}{c|}{Trunk} & \multicolumn{1}{c|}{Trunk} & Airplane \\ \midrule
        CIFAR-10 & \multicolumn{1}{c|}{Bird} & \multicolumn{1}{c|}{Truck} & Dog Forg Horse & \multicolumn{1}{c|}{Airplane} & \multicolumn{1}{c|}{Airplane} & Airplane \\ \midrule
        GTSRB & \multicolumn{1}{c|}{Speed limit (60km/h)} & \multicolumn{1}{c|}{Speed limit (50km/h)} & \begin{tabular}[c]{@{}c@{}}Speed limit (50km/h)\\ End of speed limit(80km/h)\\ Speed limit(100km/h)\end{tabular} & \multicolumn{1}{c|}{Priority Road} & \multicolumn{1}{c|}{Priority Road} & Speed limit(20km/h) \\ \midrule
        SVHN & \multicolumn{1}{c|}{2} & \multicolumn{1}{c|}{1} & 5 6 7 & \multicolumn{1}{c|}{0} & \multicolumn{1}{c|}{0} & None \\ \midrule
        \rowcolor[HTML]{D9D9D9} 
        \cellcolor[HTML]{D9D9D9} & \multicolumn{3}{c|}{\cellcolor[HTML]{D9D9D9}TaCT} & \multicolumn{1}{c|}{\cellcolor[HTML]{D9D9D9}SSLBackdoor} & \multicolumn{1}{c|}{\cellcolor[HTML]{D9D9D9}CorruptEncoder} & - \\ \cline{2-7} 
        \rowcolor[HTML]{D9D9D9} 
        \multirow{-2}{*}{\cellcolor[HTML]{D9D9D9}Dataset} & \multicolumn{1}{c|}{\cellcolor[HTML]{D9D9D9}Target Class} & \multicolumn{1}{c|}{\cellcolor[HTML]{D9D9D9}Source Class} & Cover Class & \multicolumn{1}{c|}{\cellcolor[HTML]{D9D9D9}Target Class} & \multicolumn{1}{c|}{\cellcolor[HTML]{D9D9D9}Target Class} & - \\ \midrule
        ImageNet-10 & \multicolumn{1}{c|}{\begin{tabular}[c]{@{}c@{}}n01855672\\ (Goose)\end{tabular}} & \multicolumn{1}{c|}{\begin{tabular}[c]{@{}c@{}}n01775062\\ (Wolfspider)\end{tabular}} & \begin{tabular}[c]{@{}c@{}}n02120079(Whitefox)\\ n02447366(Badger)\\ n02483362(Gibbon)\end{tabular} & \multicolumn{1}{c|}{\begin{tabular}[c]{@{}c@{}}n02116738\\ (Hunting Dog)\end{tabular}} & \multicolumn{1}{c|}{\begin{tabular}[c]{@{}c@{}}n02116738\\ (Hunting Dog)\end{tabular}} & \begin{tabular}[c]{@{}c@{}}-\end{tabular} \\ \bottomrule
        \end{tabular}
        }
    }

    \label{tab:target_class}
    \end{table}

\subsection{Baseline Defense Explanation}
\noindent

For baseline defense, we tailor all the end-to-end training from scratch to transfer learning using a clean or poisoned encoder based on the specific threat.
We utilize the publicly available code of \cite{qi_sec_2023}\footnote{\url{https://github.com/Unispac/Fight-Poison-With-Poison}} to implement STRIP, AC, Spectral, SPECTRE, and CT. 
We utilize the toolbox\footnote{\url{https://github.com/THUYimingLi/BackdoorBox}} to implement IBD-PSC and SCALE-UP.
We implement ASSET\footnote{\url{https://github.com/reds-lab/ASSET}}, 
ABL\footnote{\url{https://github.com/bboylyg/ABL}}, CBD\footnote{\url{https://github.com/zaixizhang/CBD}}, 
CLP\footnote{\url{https://github.com/rkteddy/channel-Lipschitzness-based-pruning}},
SSLBackdoorMitigation\footnote{\url{https://github.com/wssun/SSLBackdoorMitigation}},
META-SIFT\footnote{\url{https://github.com/ruoxi-jia-group/Meta-Sift}}, 
FT-SAM\footnote{\url{https://github.com/SCLBD/BackdoorBench}},
and I-BAU\footnote{\url{https://github.com/YiZeng623/I-BAU}} using their respective official code repositories.

\clearpage

\begin{table*}[t]
    \caption{Performance of the state-of-the-art \textbf{inference-time defenses} SCALE-UP and IBD-PSC under different threat scenarios, using the CIFAR-10 dataset. 
    }
    \resizebox{\textwidth}{!}{
    \begin{tabular}{c|c|cccc|cccc}
    \toprule
    \rowcolor[HTML]{FFFFFF} 
    \cellcolor[HTML]{FFFFFF} & \cellcolor[HTML]{FFFFFF} & \multicolumn{4}{c|}{\cellcolor[HTML]{FFFFFF}\textbf{SCALE-UP}} & \multicolumn{4}{c}{\cellcolor[HTML]{FFFFFF}\textbf{IBD-PSC}} \\ \cline{3-10} 
    \rowcolor[HTML]{FFFFFF} 
    \multirow{-2}{*}{\cellcolor[HTML]{FFFFFF}\textbf{Threat Type}} & \multirow{-2}{*}{\cellcolor[HTML]{FFFFFF}\begin{tabular}[c]{@{}c@{}}\textbf{Poisoning}\\ \textbf{Method}\end{tabular}} & \multicolumn{1}{c|}{\cellcolor[HTML]{FFFFFF}\textbf{TPR(\%)$\uparrow$}} & \multicolumn{1}{c|}{\cellcolor[HTML]{FFFFFF}\textbf{FPR(\%)$\downarrow$}} & \multicolumn{1}{c|}{\cellcolor[HTML]{FFFFFF}\textbf{AUC$\uparrow$}} & \textbf{F1$\uparrow$}& \multicolumn{1}{c|}{\cellcolor[HTML]{FFFFFF}\textbf{TPR(\%)$\uparrow$}} & \multicolumn{1}{c|}{\cellcolor[HTML]{FFFFFF}\textbf{FPR(\%)$\downarrow$}} & \multicolumn{1}{c|}{\cellcolor[HTML]{FFFFFF}\textbf{AUC$\uparrow$}} & \textbf{F1$\uparrow$} \\ \midrule
     & \textbf{BadNets} & \multicolumn{1}{c|}{92.80} & \multicolumn{1}{c|}{33.98} & \multicolumn{1}{c|}{0.79} & 0.12 & \multicolumn{1}{c|}{0.00} & \multicolumn{1}{c|}{3.38} & \multicolumn{1}{c|}{0.48} & 0.00 \\ \cline{2-10} 
     & \textbf{Blended} & \multicolumn{1}{c|}{20.24} & \multicolumn{1}{c|}{38.22} & \multicolumn{1}{c|}{0.41} & 0.03 & \multicolumn{1}{c|}{0.24} & \multicolumn{1}{c|}{3.27} & \multicolumn{1}{c|}{0.48} & 0.00 \\ \cline{2-10} 
     & \textbf{SIG} & \multicolumn{1}{c|}{14.80} & \multicolumn{1}{c|}{37.66} & \multicolumn{1}{c|}{0.39} & 0.02 & \multicolumn{1}{c|}{0.80} & \multicolumn{1}{c|}{3.13} & \multicolumn{1}{c|}{0.49} & 0.01 \\ \cline{2-10} 
     & \textbf{WaNet} & \multicolumn{1}{c|}{28.56} & \multicolumn{1}{c|}{40.07} & \multicolumn{1}{c|}{0.44} & 0.03 & \multicolumn{1}{c|}{4.80} & \multicolumn{1}{c|}{0.42} & \multicolumn{1}{c|}{0.52} & 0.08 \\ \cline{2-10} 
     & \textbf{TaCT} & \multicolumn{1}{c|}{56.88} & \multicolumn{1}{c|}{35.98} & \multicolumn{1}{c|}{0.60} & 0.07 & \multicolumn{1}{c|}{0.00} & \multicolumn{1}{c|}{5.29} & \multicolumn{1}{c|}{0.47} & 0.00 \\ \cline{2-10} 
     &\textbf{ Adap-Blend} & \multicolumn{1}{c|}{29.60} & \multicolumn{1}{c|}{35.54} & \multicolumn{1}{c|}{0.47} & 0.04 & \multicolumn{1}{c|}{0.80} & \multicolumn{1}{c|}{4.14} & \multicolumn{1}{c|}{0.48} & 0.01 \\ \cline{2-10} 
    \multirow{-7}{*}{\threattwo} & \textbf{Adap-Patch} & \multicolumn{1}{c|}{32.48} & \multicolumn{1}{c|}{34.58} & \multicolumn{1}{c|}{0.49} & 0.04 & \multicolumn{1}{c|}{1.92} & \multicolumn{1}{c|}{2.28} & \multicolumn{1}{c|}{0.50} & 0.02 \\ \midrule
    \rowcolor[HTML]{FFFFFF} 
    \textbf{Threat Type} & \begin{tabular}[c]{@{}c@{}}\textbf{Poisoning}\\ \textbf{Method}\end{tabular} & \multicolumn{1}{c|}{\cellcolor[HTML]{FFFFFF}\textbf{TPR(\%)$\uparrow$}} & \multicolumn{1}{c|}{\cellcolor[HTML]{FFFFFF}\textbf{FPR(\%)$\downarrow$}} & \multicolumn{1}{c|}{\cellcolor[HTML]{FFFFFF}\textbf{AUC$\uparrow$}} & \textbf{F1$\uparrow$} & \multicolumn{1}{c|}{\cellcolor[HTML]{FFFFFF}\textbf{TPR(\%)$\uparrow$}} & \multicolumn{1}{c|}{\cellcolor[HTML]{FFFFFF}\textbf{FPR(\%)$\downarrow$}} & \multicolumn{1}{c|}{\cellcolor[HTML]{FFFFFF}\textbf{AUC$\uparrow$}} & \textbf{F1$\uparrow$} \\ \midrule
     & \textbf{BadEncoder} & \multicolumn{1}{c|}{33.60} & \multicolumn{1}{c|}{37.00} & \multicolumn{1}{c|}{0.48} & 0.04 & \multicolumn{1}{c|}{3.76} & \multicolumn{1}{c|}{4.59} & \multicolumn{1}{c|}{0.50} & 0.03 \\ \cline{2-10} 
    \multirow{-2}{*}{\threatone} & \textbf{DRUPE} & \multicolumn{1}{c|}{33.84} & \multicolumn{1}{c|}{36.90} & \multicolumn{1}{c|}{0.48} & 0.04 & \multicolumn{1}{c|}{0.24} & \multicolumn{1}{c|}{3.93} & \multicolumn{1}{c|}{0.48} & 0.00 \\ \hline
     & \textbf{BadEncoder} & \multicolumn{1}{c|}{53.04} & \multicolumn{1}{c|}{38.60} & \multicolumn{1}{c|}{0.57} & 0.06 & \multicolumn{1}{c|}{50.64} & \multicolumn{1}{c|}{3.96} & \multicolumn{1}{c|}{0.73} & 0.33 \\ \cline{2-10} 
    \multirow{-2}{*}{\threatthree} & \textbf{DRUPE} & \multicolumn{1}{c|}{51.76} & \multicolumn{1}{c|}{34.18} & \multicolumn{1}{c|}{0.59} & 0.07 & \multicolumn{1}{c|}{3.76} & \multicolumn{1}{c|}{6.98} & \multicolumn{1}{c|}{0.48} & 0.02 \\ \bottomrule
    \end{tabular}
    }
    \label{tab:inference-time-defense}
\end{table*}

\begin{figure*}[ht]
\begin{center}
\begin{tcolorbox}[colback=lightgray!20, colframe=black, boxrule=0.5pt, arc=0pt, outer arc=0pt, boxsep=0pt, left=2pt, right=2pt, top=2pt, bottom=2pt]
\begin{lstlisting}
class MaskedLinear(nn.Module):
    def __init__(self, original_weight, mask, bias=None):
        super().__init__()
        self.original_weight = original_weight
        self.mask = mask  # Trainable mask initialized as all ones
        self.bias = bias

    def forward(self, x):
        # Clamp mask between 0 and 1
        self.mask.data.clamp_(0,1)
        masked_weight = self.original_weight * self.mask.unsqueeze(1)
        return torch.nn.functional.linear(x, masked_weight, self.bias)
\end{lstlisting}
\end{tcolorbox}
\caption{Implementation of the \texttt{MaskedLinear} class, which applies a trainable mask to the weights of a linear layer. We use this code to transform a linear layer of \textbf{ViT} into a masked layer. The mask is clamped between 0 and 1 to ensure valid weight adjustments.
}
\label{fig:masked_linear}
\end{center}
\end{figure*}

\end{document}